\documentclass[letterpaper]{article} 
\usepackage[draft]{aaai2026}
\usepackage{times}  
\usepackage{helvet}  
\usepackage{courier}  
\usepackage[hyphens]{url}  
\usepackage{graphicx} 
\usepackage{natbib}  
\usepackage{caption} 
\usepackage{algorithm}
\usepackage{algorithmic}

\usepackage{multirow}
\usepackage[table,xcdraw]{xcolor}
\usepackage{subfig}

\usepackage{amsmath}
\usepackage{amssymb}
\usepackage{booktabs}

\usepackage{enumitem}
\usepackage{newfloat}
\usepackage{listings}
\DeclareCaptionStyle{ruled}{labelfont=normalfont,labelsep=colon,strut=off} 
\floatstyle{ruled}
\newfloat{listing}{tb}{lst}{}
\floatname{listing}{Listing}
\title{StreamSTGS: Streaming Spatial and Temporal Gaussian Grids for Real-Time Free-Viewpoint Video}
\author{
    Zhihui Ke, 
    Yuyang Liu,
    Xiaobo Zhou\thanks{Corresponding author.},
    Tie Qiu
}
\affiliations{
    School of Computer Science and Technology, Tianjin University, China\\

    kezhihui@tju.edu.cn, yvyang\_liu@tju.edu.cn, xiaobo.zhou@tju.edu.cn, qiutie@tju.edu.cn
}

\begin{document}

\maketitle

\begin{abstract}
Streaming free-viewpoint video~(FVV) in real-time still faces significant challenges, particularly in training, rendering, and transmission efficiency. Harnessing superior performance of 3D Gaussian Splatting~(3DGS), recent 3DGS-based FVV methods have achieved notable breakthroughs in both training and rendering. However, the storage requirements of these methods can reach up to $10$MB per frame, making stream FVV in real-time impossible. To address this problem, we propose a novel FVV representation, dubbed StreamSTGS, designed for real-time streaming. StreamSTGS represents a dynamic scene using canonical 3D Gaussians, temporal features, and a deformation field. For high compression efficiency, we encode canonical Gaussian attributes as 2D images and temporal features as a video. This design not only enables real-time streaming, but also inherently supports adaptive bitrate control based on network condition without any extra training. Moreover, we propose a sliding window scheme to aggregate adjacent temporal features to learn local motions, and then introduce a transformer-guided auxiliary training module to learn global motions. On diverse FVV benchmarks, StreamSTGS demonstrates competitive performance on all metrics compared to state-of-the-art methods. Notably, StreamSTGS increases the PSNR by an average of $1$dB while reducing the average frame size to just $170$KB.
\end{abstract}

\section{Introduction}
\label{sec:intro}

Free-viewpoint video~(FVV) is a crucial application in Virtual Reality~(VR) with significant potential in education, industry, and entertainment, as it provides immersive user experiences. However, the 4D representation of FVV requires substantial storage, creating challenges for real-time transmission and hindering its practical application.

Recent advances in Neural Radiance Field (NeRF)~\cite{mildenhall2021nerf} have significantly facilitated the development of FVV. These NeRF-based works~\cite{li2022streaming,wang2023neural,zheng2024hpc} reconstruct a NeRF model for each timestamp and store the residuals of these model, thereby allowing the streaming of these residuals. Though enabling real-time transmission, the volume rendering used in NeRF prevents real-time rendering. The most recent 3D Gaussian Splatting~(3DGS)~\cite{kerbl20233d} has achieved real-time rendering by directly projecting 3D Gaussians onto the 2D image plane and has thus attracted much researches.

Leveraging the high performance of 3DGS, many efforts have been made to extend 3DGS to dynamic scene reconstruction. Dynamic 3DGS methods~\cite{yang2023gs4d,duan20244d} incorporate a time dimension directly into Gaussian attributes, while deformable 3DGS methods~\cite{yang2024deformable,wu20244d,li2024spacetime} utilize canonical 3D Gaussians to represent the geometry in canonical space and model motion through deformation fields~(e.g., MLPs, HexPlanes, and Hash Grid). In these works, Gaussian attributes and temporal features are tightly coupled, resulting in them unsuitable for streaming. Consequently, some studies~\cite{sun20243dgstream,girish2024queen,gao2024hicom} focus on reconstructing streamable dynamic 3DGS for FVV. These methods typically use the first frame to reconstruct a 3D Gaussian representation and then predict attribute offsets for subsequent frames using a neural network, employing a per-frame training strategy to model the scene changes. However, the frame-by-frame training method brings significant cumulative errors and hard to dual with new objects. Furthermore, these methods require training multiple Level-of-detail~(LoD) models to accommodate varying network conditions. Importantly, attribute offsets prediction inevitably becomes a part of decoding latency. If a user wants to view frame $i$, they have to wait for all previous $i-1$ frames to be inferred to obtain accumulated Gaussian attribute offsets. 

In this paper, we propose the Streamable Spatial and Temporal Gaussian Grids~(StreamSTGS) representation, designed to achieve real-time FVV while adapting to dynamic network conditions. We decouple dynamic 3DGS into canonical 3D Gaussians, a series of temporal features, and a deformation field. Then, we use temporal features and the deformation field to predict the deformation of canonical 3D Gaussian. Moreover, we apply a sliding window to aggregate adjacent temporal features to learn local object motions. To reduce model size, inspired by~\cite{morgenstern2025compact}, we represent canonical 3D Gaussians and temporal features as spatial grids and temporal grids, respectively. Ultimately, spatial grids are stored as images while temporal grids are saved as a video. Notably, image and video formats naturally support adaptive streaming without any extra training. However, utilizing temporal features rather than HexPlanes or Hash Grid makes StreamSTGS hard to learn global motions, leading to blurring of dynamic areas. To address this, we propose a transformer-guided auxiliary training strategy, which employs a transformer to learn global motions and distill the learned features into StreamSTGS. This allow us to discard transformer during rendering to maintain high FPS. Furthermore, different from the frame-by-frame training strategy used in existing GS-based FVV frameworks, we train our StreamSTGS using a Group of Pictures~(GOP), enabling it to capture large motions and sudden changes. As a result, StreamSTGS achieves superior performance in reconstruction quality, storage size, rendering and training speed.

The contributions are summarized as follows:
\begin{itemize}
   \item We propose StreamSTGS, a streamable spatial and temporal Gaussian grids representation for real-time FVV, where a sliding window is applied to temporal Gaussian girds to capture adjacent temporal motions.
   \item We introduce spatial and temporal smoothness loss to regularize spatial and temporal Gaussian grids, enabling their compression into images and videos, which significantly reduces model size and inherently adapts to dynamic network conditions.
   \item We develop a transformer-guided auxiliary training strategy, which helps StreamSTGS learn global motions without impact the rendering speed.
\end{itemize}

\section{Related Work}
\label{sec:related}

\subsection{Dynamic Scene Reconstruction}
Given the success of NeRFs~\cite{mildenhall2021nerf,yu2021plenoctrees,muller2022instant,chen2022tensorf}, there are many methods have extended NeRF to 4D space-time radiance fields to reconstruct dynamic scenes from multi-view or monocular videos~\cite{xian2021space,gao2021dynamic,li2021neural, park2021nerfies, fang2022fast, park2021hypernerf, du2021neural, li2022nv3d,yan2023nerf,liu2023robust,tian2023mononerf,zhan2024kfd}. However, NeRF-based dynamic scene reconstruction methods are constrained by volume rendering, resulting in slow rendering speed. Consequently, recent works have focused on accelerating NeRF-based methods from different perspectives. Specifically, Mixvoxels~\cite{wang2022mixed}, and \cite{guo2023forward} use voxels to represent the geometry and appearance of 3D space. K-Planes~\cite{fridovich2023k}, HexPlane~\cite{cao2023hexplane}, Tensor4D~\cite{shao2023tensor4d}, and DMiT~\cite{yang2025dmit} decompose 4D space into multiple planes to enhance computational and storage efficiency. MSTH~\cite{wang2024masked}, $F^2$NeRF~\cite{wang2023f2}, and \cite{park2023temporal} utilize hash tables~\cite{muller2022instant} to store features in voxels or planes. NeRFPlayer~\cite{song2023nerfplayer} and HyperReel~\cite{attal2023hyperreel} adopt decomposed vector-matrix~\cite{chen2022tensorf}, while DaReNeRF~\cite{lou2024darenerf} propose a discrete wavelet transform (DWT)-based representation to model the 4D space. Though these methods significantly accelerate the rendering speed, a considerable gap remains in achieving real-time performance. 

Recently, following the revolutionary 3DGS~\cite{kerbl20233d}, dynamic Gaussian Splatting has quickly gained attention for dynamic scene reconstruction. For instance, deformable 3DGS methods~\cite{luiten2024dynamic, bae2025per, lu20243d, shaw2024swings, yang2024deformable, huang2024sc, zhao2024gaussianprediction, wansuperpoint, wu20244d, lu2024dn, duisterhof2023md, li2024spacetime, katsumata2025compact, lin2024gaussian, xu2024grid4d,yan20244d, zhu2024motiongs,kwak2025modec,lei2025mosca,fan2025spectromotion} construct canonical 3D Gaussians and utilize a deformation field to deform each Gaussians to specific timestamps. Other works~\cite{yang2023gs4d,duan20244d} incorporate a time dimension into 3D Gaussian attributes to form 4D Gaussian primitives. Meanwhile, some works directly learn 3D Gaussian motions through spline function~\cite{lee2024fully,park2025splinegs}. Kim et al.~\cite{kim20244d} introduce uncertainty-aware regularization to improve reconstruction performance. These methods train a dynamic scene as a single model, which limits their streamability.

\subsection{Free-viewpoint Video Reconstruction}
The emergence of implicit neural representations offers a new paradigm for capturing FVV from multi-view videos. StreamRF~\cite{li2022streaming}, ReRF~\cite{wang2023neural}, HPC~\cite{zheng2024hpc},\cite{zhang2024rate},FSVFG~\cite{yin2024fsvfg}, and VRVVC~\cite{hu2025vrvvc} employ a per-frame training strategy, where a NeRF model is initially trained on the first frames and then incrementally refined with field residuals to adapt to the current timestamp. This approach allows streaming only the field residuals rather than the entire NeRF model. Similarly, 3DGStream~\cite{sun20243dgstream}, HiCoM~\cite{gao2024hicom}, QUEEN~\cite{girish2024queen}, 4DGC~\cite{hu20254dgc} apply per-frame training strategy to 3DGS for FVV reconstruction. 3DGStream~\cite{sun20243dgstream} utilizes InstantNGP~\cite{muller2022instant} to predict Gaussian attribute offsets and streams InstantNGP directly, which have a large model size. HiCoM~\cite{gao2024hicom} prefer to store and stream Gaussian attribute offsets, but it still exhibits high spatial and temporal redundancy. QUEEN~\cite{girish2024queen}, VideoGS~\cite{wang2024v}, GIFStream~\cite{li2025gifstream}, and 4DGC~\cite{hu20254dgc} further employ entropy encoding to compress Gaussian attribute offsets but suffer from high decoding latency. Moreover, frame-by-frame training strategy brings significant cumulative errors. VideoRF~\cite{wang2024videorf} and TeTriRF~\cite{wu2024tetrirf} use 2D grids to store temporal 3D voxel features and compress these grids as videos. Nevertheless, their NeRF-based representation and aggressive quantization limit rendering speed and reconstruction quality. 

\section{Preliminary}

\begin{figure*}[!t]
   \centering
   \includegraphics[width=1\textwidth]{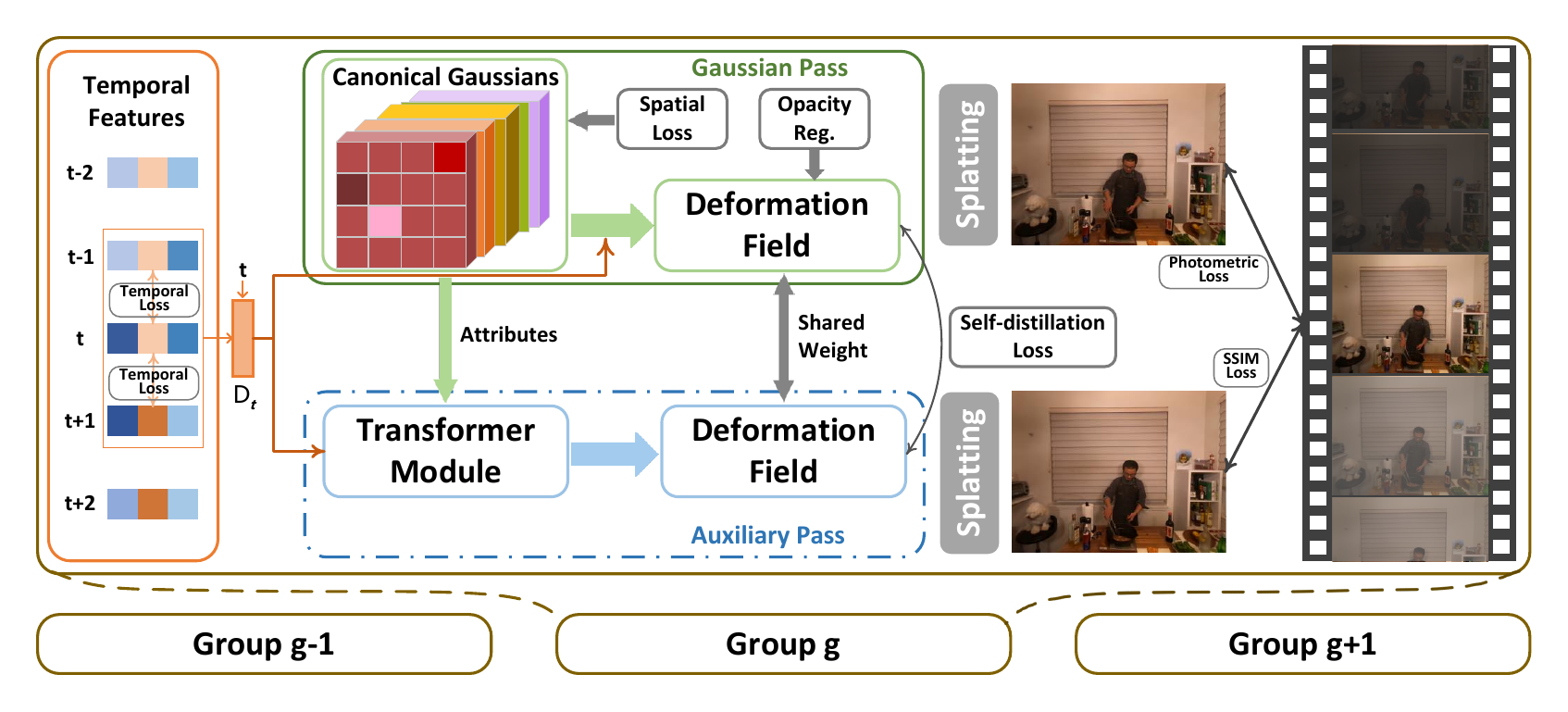}
   \caption{Overview of the StreamSTGS framework. First, the long video sequence is split into multiple groups. Within each group, a sliding window is employed to extract multiple temporal features, which serve as inputs to the deformation field for predicting the deformation of canonical 3D Gaussians. For real-time streaming, the canonical Gaussians are represented as images, while the temporal features are encoded as a video. To improve global motion learning, we introduce a Transformer-guided auxiliary training strategy, which can be removed during inference to achieve higher FPS.}
   \label{fig:framework}
\end{figure*}

3D Gaussian Splatting~(3DGS)~\cite{kerbl20233d} explicitly reconstructs a scene using anisotropic 3D Gaussians, which have five attributes: 1) Position $X \in \mathbb{R}^3$~(i.e., the mean of the Gaussian function), 2) Scale $S \in \mathbb{R}^3$, represented as a diagonal matrix, 3) Rotation matrix $R$, parameterized by a quaternion vector $Q \in \mathbb{R}^4$, 4) Opacity $O \in [0, 1]$, and 5) Color $C$, represented by spherical harmonic coefficients. The covariance matrix is derived from scale and rotation matrices as $\Sigma=RSS^{T}R^{T}$. For rendering, these 3D Gaussians are then projected into 2D Gaussians according to a given camera intrinsic matrix $K$ and a viewing transformation matrix $W$ through differential splatting~\cite{zwicker2001surface}:
\begin{equation}
   X^{\prime} = K((WX) / (WX)_z), \Sigma^{\prime} = JW \Sigma W^TJ^T
\end{equation}
where $J$ denotes the Jacobian matrix. Following depth sorting, the color of a pixel $x$ is computed as:
\begin{equation}
   color(x) = \sum_{m \in N} c_m \alpha_m \prod_{j=1}^{m-1} (1 -\alpha_j)
\end{equation}
where $c_m$ and $\alpha_m$ represent the color and opacity of Gaussian $m$.

\section{Method}
\label{sec:method}
To realize real-time dynamic 3DGS streaming, our proposed StreamSTGS framework incorporates four key innovations, as illustrated in Fig.~\ref{fig:framework}. First, we introduce a stream-friendly dynamic 3DGS representation, StreamSTGS, which facilitates real-time streaming through traditional video codecs. Second, we develop Gaussian attribute compression, which significantly reduces the model size of our StreamSTGS representation. Additionally, we propose a Transformer-guided auxiliary training strategy to help the global motion learning in StreamSTGS without compromising FPS performance. Finally, we introduce dynamic-aware density and Gaussian relocate schemes to improve the reconstruction quality under limited Gaussian quantity.

\subsection{StreamSTGS Representation}
\label{sec:streamgs}
We adopt the group of pictures (GOP) structure widely used in video compression, where each GOP comprises $G$ frames, denoted as $\mathcal{T} = \{ t_0, t_1, \dots, t_G \}$. The dynamic scene within each GOP is independently reconstructed using our proposed StreamSTGS representation. As shown in Fig.~\ref{fig:framework}, StreamSTGS within a GOP consists of a canonical 3D Gaussians set $\mathcal{G}$, temporal features, and a deformation field. Unlike existing deformable 3DGS methods that typically rely on HexPlanes~\cite{cao2023hexplane} to model motion in dynamic scenes, we employ temporal features $\mathcal{E} = \{e_0, e_1, \dots, e_E \}$ to directly learn dynamic features, enabling the streaming of temporal features in a manner akin to traditional video. 

Considering the temporal correlation of motion in real-world dynamic scenarios, we employ a sliding window of length $W$ to capture the motion relationship between adjacent frames, extracting $W$ temporal features, as depicted in Fig.~\ref{fig:framework}. For example, when $W = 3$, we concatenate the temporal features $e_{i-1}$, $e_i$, and $e_{i+1}$ at timestamp $t_i$. For the next timestamp $t_{i+1}$, we concatenate $e_i$, $e_{i+1}$, and $e_{i+2}$. These concatenated features are then processed by a temporal MLP $D_t$ to predict temporal feature $f_i$:
\begin{align}
\label{eq:temporal-feat}
fe_i &= concate(e_{i-1}, e_{i}, e_{i+1}) \\
f_i &= D_t(fe_i, \gamma(t_i)), \\
\gamma(t_i) &= (sin(2^l \pi t_i), cos(2^l \pi t_i))_{l=0}^{L-1},
\end{align}
where position encoding~\cite{mildenhall2021nerf} is applied to transform timestamp $t_i$ to a high frequency representation of dimension $L$. This approach not only captures the motion relationships between adjacent frames but also reduces the number of temporal features to $E = G + W - 1$.

The original 3DGS uses spherical harmonic to represent view-dependent color, which requires large storage and is not well-suited for real-time streaming. Therefore, we use a three-channel tensor $C$ to learn view-independent color and employ a color decoder $D_c$ and an opacity decoder $D_o$ to model view-dependent and time-varying color as follows:
\begin{equation}
\Delta O = tanh(D_o(f_i, view)),\ \Delta C = D_c(f_i, view),
\end{equation}
where $view$ denotes the camera direction. Since dynamic objects may appear or disappear over time, we apply the tanh activation function to the output of the opacity decoder to model its variations.
Moreover, we use a velocity decoder $D_v$ and a covariance decoder $D_{cov}$ to predict the deformation of position $\Delta X = D_v(f_i)$ as well as scale and rotation changes $[\Delta S, \Delta Q] = D_{cov}(f_i)$. The final deformed 3D Gaussians $\mathcal{G}^i$ at timestamp $t_i$ are given by:
\begin{align}
      \mathcal{G}^i &= (X^i, S^i, R^i, O^i, C^i) \\
                  &= (X + \Delta X, S + \Delta S, Q + \Delta Q, O + \Delta O, C^{\prime} + \Delta C), \nonumber
\end{align}
where $C^{\prime} = relu(C)$ ensures the canonical color remains non-negative. Finally, we employ the splatting algorithm~\cite{zwicker2001surface} to render the target image $\mathcal{I}_i$.

\subsection{Gaussian Attribute Compression}
\label{sec:compression}
\begin{figure}[!t]
   \centering
   \includegraphics[width=1.0\columnwidth]{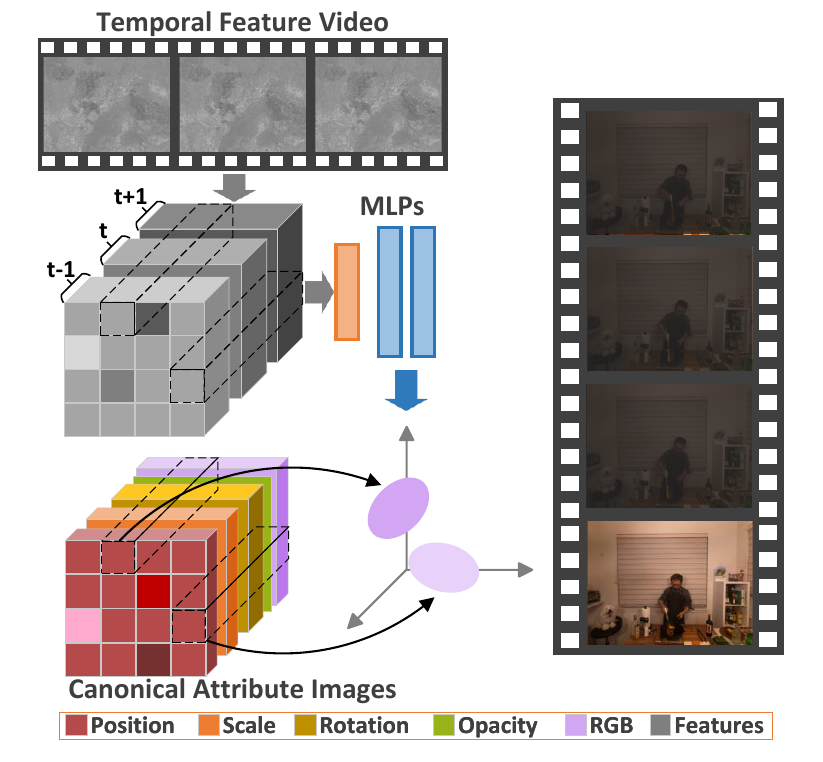}
   \caption{The 2D grid representation of our StreamSTGS. Canonical Gaussian attributes are compressed as 2D images and temporal features are encoded as a video for real-time streaming.}
   \label{fig:representation}

\end{figure}

In achieve a compact representation for real-time streaming, we draw inspiration from work~\cite{morgenstern2025compact} that sorts 3D Gaussians into 2D grids through the PLAS algorithm. We organize position, scale, rotation, opacity, and color into five attribute images and further organize $E$ temporal features into $E$ feature images. These feature images are then compressed into a feature video using traditional video codecs~(e.g., H.264 and HEVC), thereby significantly reducing redundancy among temporal features, as shown in Fig.~\ref{fig:representation}. Furthermore, our proposed representation inherently supports adaptive bit-rate transmission without requiring extra training, as feature video can be encoded and decoded using different bit-rates according to network conditions.

\textbf{Consistency regularization} For high compression efficiency, we adopt a method~\cite{morgenstern2025compact} that applies a 2D Gaussian filter to the sorted 2D attribute images to smooth adjacent Gaussian attributes. The spatial smooth regularization, $\mathcal{L}_{spatial}$, is defined as the difference between the smoothed 2D images and the original 2D images. As for feature images, we introduce temporal consistency regularization. Specifically, we compute the difference between $e_{i-1}$ and $e_i$, as well as between $e_i$ and $e_{i+1}$. Then, we calculate the mean of these two differences as the temporal consistency regularization $\mathcal{L}_{temporal}$:
\begin{align}
&L_i = huber(e_{i-1} - e_i), \\
&L_{i+1} = huber(e_i - e_{i+1}), \\
&\mathcal{L}_{temp} = mean(L_i, L_{i+1}),
\end{align}
where $huber$ is the Huber loss, which applies MSE to deviations when the delta below a threshold and MAE to deviations when the delta exceeds the threshold. This loss is well-suited for temporal consistency regularization. Since static Gaussians exhibit similarity across different temporal features, MSE is used to preserve these feature values consistency to reduce the data size of feature video. In contrast, dynamic Gaussians show significant differences across different temporal features, and MAE is used to smooth these values while tolerating outliers. Applying MSE to such outliers would result in over-smoothing and degrade dynamic motion reconstruction.

\subsection{Transformer-Guided Auxiliary Training}
\label{sec:transformer}
Though we design a sliding window scheme to learn local motion, learning global motions remains challenging due to the limited length of the sliding window. Inspired by TimeFormer~\cite{jiang2024timeformer}, which treats all timestamps as a sequence and utilizes Batch Attention~\cite{hou2022batchformer} to learn robust global motions. However, they concatenate the position of canonical 3D Gaussians $X$ and different timestamps $t_i$ as input, which primarily enhances the learning of canonical 3D Gaussians rather than temporal features. In this paper, we use the output of the temporal MLP $f_i$ as the input:
\begin{equation}
f^{\prime}_i = \mathcal{F}(f_i, \gamma(t_i), \gamma(X)),
\end{equation}
where $\mathcal{F}$ is the Transformer module. Note that $\gamma(t_i)$ and $\gamma(X)$ provide position encoding for the Transformer. $\gamma(t_i)$ encodes the position of the sliding window in temporal features, allowing the model to establish relationships among different sliding windows for global motion learning. $\gamma(X)$ provides the position of 3D Gaussians and helps transformer jointly learn the spatial-temporal features, especially in our fully decoupled spatial and temporal representation. 

To tolerate with the loss of feature encoding, we add a small noise $\varepsilon$ to the sliding window during training:
\begin{equation}
   fe_i = concate(e_{i-1} + \varepsilon_1, e_{i} + \varepsilon_2, e_{i+1} + \varepsilon_3),
\end{equation}
where $\varepsilon_1$, $\varepsilon_2$, and $\varepsilon_3$ are random sampled from a uniform distribution $\lambda \cdot \mathcal{N}(-0.5, 0.5)$. 

Integrating the Transformer module into the StreamSTGS results in low FPS. Therefore, we employ a two-pass design, as illustrated in Fig.~\ref{fig:framework}, where the deformation field is shared between the Gaussian Pass and the Auxiliary Pass. This design allows the global motion knowledge learned by Transformer to be transferred to our StreamSTGS representation. During inference, we can remove the Transformer module to maintain a high FPS. Additionally, we introduce a self-distillation loss $\mathcal{L}_{sd}$ to facilitate knowledge transfer:
\begin{equation}
\mathcal{L}_{sd} = \left\| f_i - f^{\prime}_i \right\|_1.
\end{equation}
Moreover, we find that the SSIM loss helps our StreamSTGS representation learn finer details. Thus, we use SSIM loss to supervise the prediction of the Auxiliary Pass $\mathcal{I}^{\prime}_i$:
\begin{equation}
   \mathcal{L}_{t} = SSIM(\mathcal{I}^{\prime}_i - \mathcal{I}^{gt}_i).
\end{equation}

\subsection{Optimization}
\label{sec:optimization}
\textbf{dynamic-aware density} Existing Gaussian-based dynamic scene reconstruction methods typically use views from the first timestamp to generate a SFM point cloud via COLMAP~\cite{schoenberger2016sfm}. However, subsequent timestamps lack enough prior points in dynamic areas. This limitation causes some 3D Gaussians in static areas attempt to clone and split to fit dynamic objects, resulting in an uneven distribution of 3D Gaussians overly dense in static areas and insufficient in dynamic areas. Consequently, the reconstruction quality in dynamic areas is poor. Thus, we propose a simple yet efficient method, namely dynamic-aware density, which applies an L1 loss to the entire predicted image but restricts the SSIM loss to dynamic pixels:
\begin{equation}
\begin{aligned}
\mathcal{L}_c &= (1-\beta) \cdot \left\| \mathcal{I}_i - \mathcal{I}^{gt}_i \right\|_1 \\
              &+ \beta \cdot SSIM(\mathcal{I}_i \cdot \mathcal{I}^{mask} - \mathcal{I}^{gt}_i \cdot \mathcal{I}^{mask}),
\end{aligned}
\end{equation}
where $I^{mask}$ is the dynamic mask. We calculate the standard deviation across $30$ frames per camera, identifying pixels as dynamic if their stand deviation exceeds a predefined threshold $\theta$. Alternatively, dynamic masks can be obtained using SAM~\cite{kirillov2023segment} or other models.

\noindent \textbf{Pruning} To limit the number of 3D Gaussians, we collect the predicted opacity $O^i$ during training and compute its average value. If the average value falls below a given threshold, we prune these 3D Gaussians. This pruning operation is safe because we have collected the opacity of Gaussians across all timestamps within a GOP. To encourage low opacity for pruning, we introduce opacity regularization $\mathcal{L}_{o}$:
\begin{equation}
   \mathcal{L}_{o} = \left\| O^i \right\|_1.
\end{equation}

In summary, the total loss is:
\begin{equation}
\begin{aligned}
   \mathcal{L} &= \mathcal{L}_c + 0.2 \cdot \mathcal{L}_{t} + \mathcal{L}_{spatial} + \alpha_{temp} \cdot \mathcal{L}_{temp} \\
               &+ \alpha_o \cdot \mathcal{L}_{o} + \alpha_{sd} \cdot \mathcal{L}_{sd}.
\end{aligned}
\end{equation}

\noindent \textbf{Gaussian relocate} We limit the number of 3D Gaussians to approximately $150k$ to reduce data size. However, this upper bound may be reached during training. In such cases, all density operations, including prune, clone, and split, have to be halted, which inevitably leads to suboptimal model performance. Therefore, we introduce a relocate operation inspired by work~\cite{kheradmand20243d}, which does not prune unnecessary Gaussians but instead moves them to more optimal position. This mechanism is particularly well-suited to our method, as it allows continuous optimization of the 3D Gaussians distribution without changing the total number of Gaussians.

\section{Experiments and Results}

\subsection{Experimental Setup}
We utilize two real-world multi-view dynamic scene datasets: 1) N3DV~\cite{li2022nv3d} dataset, which comprises six scenes captured by 18 to 21 cameras at a resolution of $2704 \times 2028$ and a frame rate of 30 FPS. Traditionally, we downsample the resolution to $1352 \times 1014$. 2) MeetRoom~\cite{li2022streaming} dataset, which uses 12 cameras to capture three scenes at a resolution of $1280 \times 720$ and a frame rate of 30 FPS. Following the approach of 4DGaussians~\cite{wu20244d}, we use COLMAP~\cite{schoenberger2016sfm} to generate initial point clouds. 

\noindent \textbf{Implementation}.
All experiments are conducted on a RTX A6000 GPU. We first train a coarse 3DGS model using all multi-view frames about $3000$ iterations with a batch size of $2$. Then, a refined model is trained for each GOP with a size of $60$. Each GOP is trained for $12000$ and $7000$ iterations for N3DV and MeetRoom datasets, respectively. The noise parameter $\lambda$ is set to $0.001$, as we observe that the loss of temporal features after compression is about $0.0002$. Additionally, $\alpha_{temp}$, $\alpha_o$, and $\alpha_{sd}$ are set to $1.0$, $0.01$, and $0.005$, respectively. Dynamic-aware density is applied after $5000$ iteration for N3DV dataset and $3000$ iteration for MeetRoom dataset. More implementation details are provided in the \textbf{Appendix}.

\begin{table*}[t!]
   \resizebox{\textwidth}{!}{%
   \begin{tabular}{c|ccc|ccccccc}
   \toprule
   Method    & PSNR $\uparrow$  & SSIM $\uparrow$  & LPIPS $\downarrow$ & \begin{tabular}[c]{@{}c@{}}Storage \\ (KB) $\downarrow$\end{tabular} & \begin{tabular}[c]{@{}c@{}}K.F. Size \\ (MB) $\downarrow$ \end{tabular} & \begin{tabular}[c]{@{}c@{}}Decoding \\ (ms) $\downarrow$ \end{tabular} & \begin{tabular}[c]{@{}c@{}}Render \\ (ms) $\downarrow$ \end{tabular} & FPS $\uparrow$  & \begin{tabular}[c]{@{}c@{}}Train \\ (s)$\downarrow$ \end{tabular} & VBR \\ 
   \midrule
   TeTriRF~\cite{wu2024tetrirf}   & 30.07                         & 0.900                         & 0.299                         & \cellcolor[HTML]{FF9E98}65.89                       & \cellcolor[HTML]{FF9E98}2.03                             & 149                                         & 652            & 1.53                        & \cellcolor[HTML]{FFFC9E}32      & $\checkmark$                    \\ \hline
   3DGStream~\cite{sun20243dgstream} & 30.73 & 0.935 & \cellcolor[HTML]{FFFC9E}0.147 & 8204                                                & 42.22                                                    & 7$\times$n      & 14                         & \cellcolor[HTML]{FFFC9E}72                          & \cellcolor[HTML]{FFCE93}17    & $\times$                      \\
   VideoGS~\cite{wang2024v}   & 27.45                         & 0.871                         & 0.213                         & 932.9                       & \cellcolor[HTML]{FFCE93}3.24                             & 48                                          & \cellcolor[HTML]{FFCE93}7            & 21               & 143        & $\checkmark$                                         \\
   HiCoM~\cite{gao2024hicom}     & \cellcolor[HTML]{FFFC9E}31.32 & \cellcolor[HTML]{FFFC9E}0.939 & 0.147 & 10704                                               & 83.35                                                    & 0      & \cellcolor[HTML]{FF9E98}6          & \cellcolor[HTML]{FF9E98}163 & \cellcolor[HTML]{FF9E98}10         & $\times$                 \\
   4DGC~\cite{hu20254dgc}     & \cellcolor[HTML]{FFCE93}31.52 & \cellcolor[HTML]{FFCE93}0.941 & \cellcolor[HTML]{FF9E98}0.143 & \cellcolor[HTML]{FFFC9E}784                                               & 21.94                                                    & 2.5$\times$n      & 12          & 78.6 & 62            & $\times$              \\
   Ours      & \cellcolor[HTML]{FF9E98}32.30 & \cellcolor[HTML]{FF9E98}0.943 & \cellcolor[HTML]{FFCE93}0.147 & \cellcolor[HTML]{FFCE93}173.6                       & \cellcolor[HTML]{FFFC9E}3.86                             & 8                             & \cellcolor[HTML]{FFFC9E}10  & \cellcolor[HTML]{FFCE93}100 & 67     & $\checkmark$                                             \\ 
   \bottomrule
   \end{tabular}%
   }
   \caption{Quantitative comparison on N3DV dataset. `VBR' indicates variable bitrate transmission.}
   \label{tab:n3dv}

\end{table*}

\begin{table*}[!t]
   \resizebox{\textwidth}{!}{%
   \begin{tabular}{c|ccc|ccccccc}
   \toprule
   Method    & PSNR $\uparrow$  & SSIM $\uparrow$  & LPIPS $\downarrow$ & \begin{tabular}[c]{@{}c@{}}Storage \\ (KB) $\downarrow$\end{tabular} & \begin{tabular}[c]{@{}c@{}}K.F. Size \\ (MB) $\downarrow$ \end{tabular} & \begin{tabular}[c]{@{}c@{}}Decoding \\ (ms) $\downarrow$ \end{tabular} & \begin{tabular}[c]{@{}c@{}}Render \\ (ms) $\downarrow$ \end{tabular} & FPS $\uparrow$  & \begin{tabular}[c]{@{}c@{}}Train \\ (s)$\downarrow$ \end{tabular} & VBR \\ 
   \midrule
   3DGStream~\cite{sun20243dgstream} & 26.41                         & 0.90                         & 0.24                         & 4108                        & 18                               & 3$\times$n             & 8.23                  & 121                         & \cellcolor[HTML]{FFCE93}11      & $\times$                    \\
   HiCoM~\cite{gao2024hicom}     & 26.69 & 0.90 & 0.23 & 5535 & 42  & 0          & \cellcolor[HTML]{FF9E98}3.64                     & \cellcolor[HTML]{FF9E98}275 & \cellcolor[HTML]{FF9E98}6        & $\times$                   \\
   4DGC~\cite{hu20254dgc}     & \cellcolor[HTML]{FFCE93}27.11 & \cellcolor[HTML]{FFCE93}0.91 & \cellcolor[HTML]{FFCE93}0.23 & \cellcolor[HTML]{FFCE93} 1196        & \cellcolor[HTML]{FFCE93} 11        & 2$\times$n          & 9.06                     & 110 & 60     & $\times$                      \\
   Ours      & \cellcolor[HTML]{FF9E98}27.41 & \cellcolor[HTML]{FF9E98}0.92 & \cellcolor[HTML]{FF9E98}0.21 & \cellcolor[HTML]{FF9E98}142                         & \cellcolor[HTML]{FF9E98}2.8                              & 6                         & \cellcolor[HTML]{FFCE93}7.93                              & \cellcolor[HTML]{FFCE93}126 & 29             & $\checkmark$                                     \\ 
   \bottomrule
   \end{tabular}%
   }
   \caption{Quantitative comparison on MeetRoom dataset. `VBR' indicates variable bitrate transmission.}
   \label{tab:meet}
\end{table*}

\begin{figure*}[!t]
   \centering
   \subfloat[TeTriRF]{
      \includegraphics[width=0.192\textwidth]{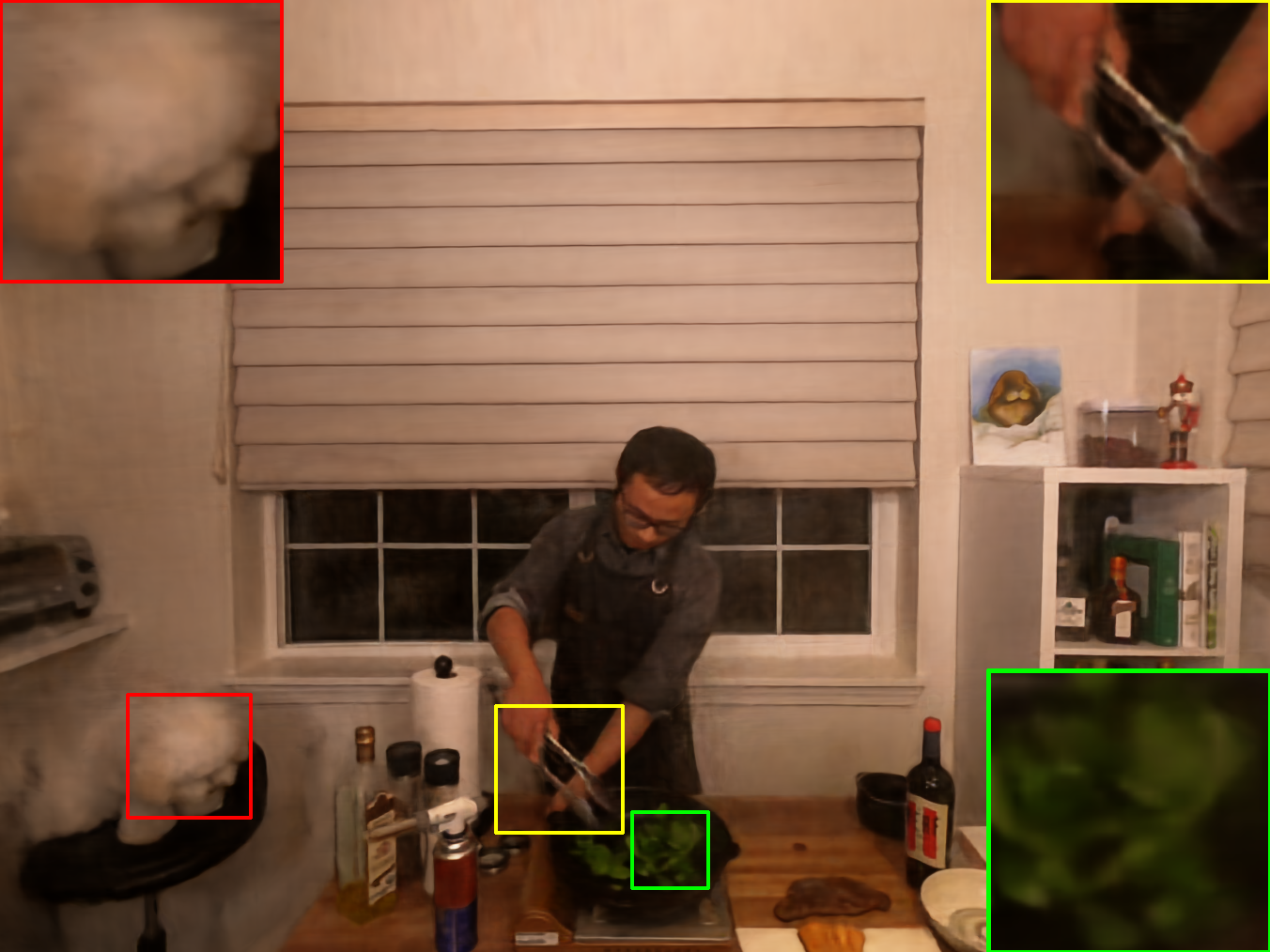}
   }
   \subfloat[HiCoM]{
      \includegraphics[width=0.192\textwidth]{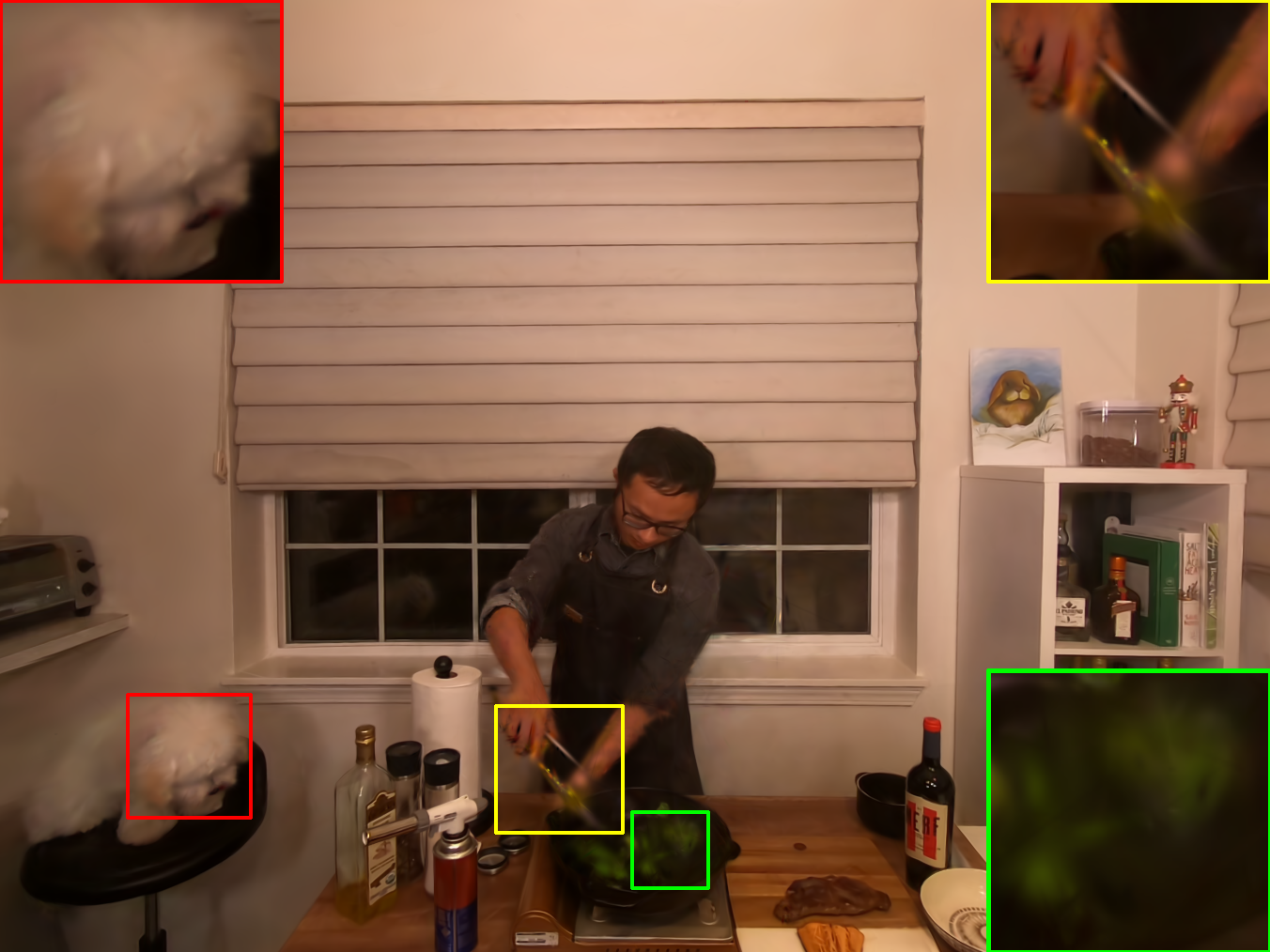}
   }
   \subfloat[4DGC]{
      \includegraphics[width=0.192\textwidth]{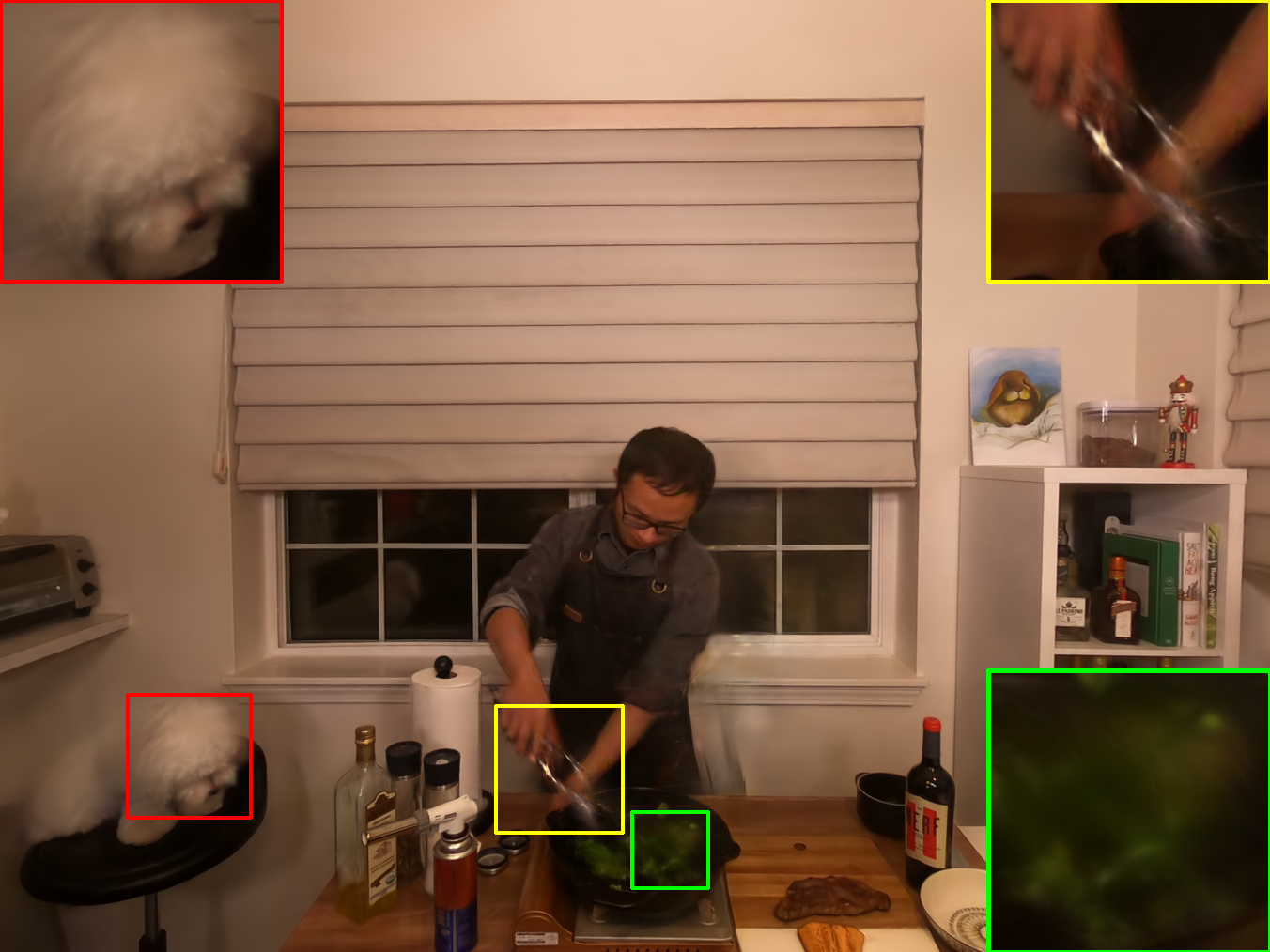}
   }
   \subfloat[Ours]{
      \includegraphics[width=0.192\textwidth]{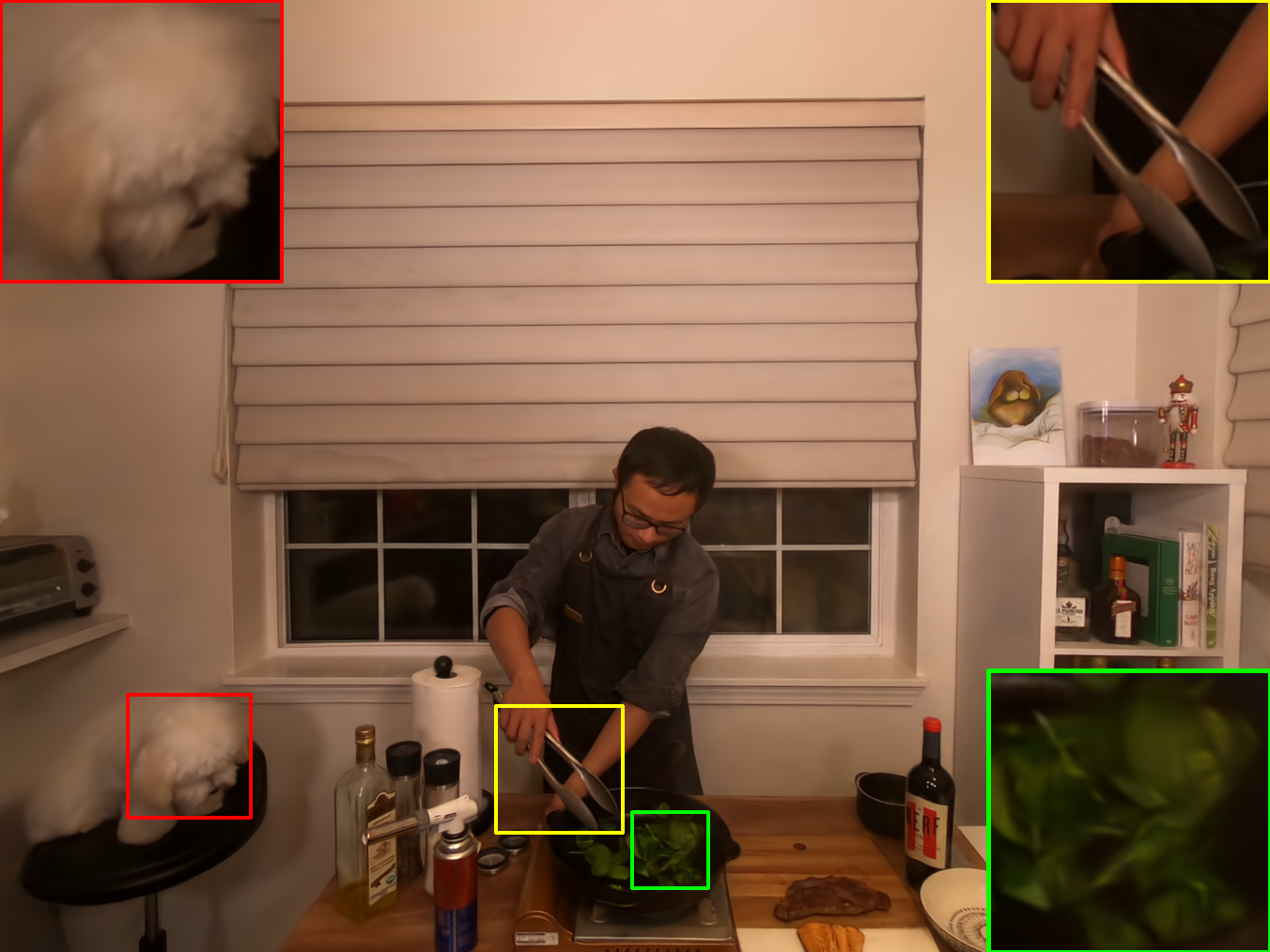}
   }
   \subfloat[GroudTruth]{
      \includegraphics[width=0.192\textwidth]{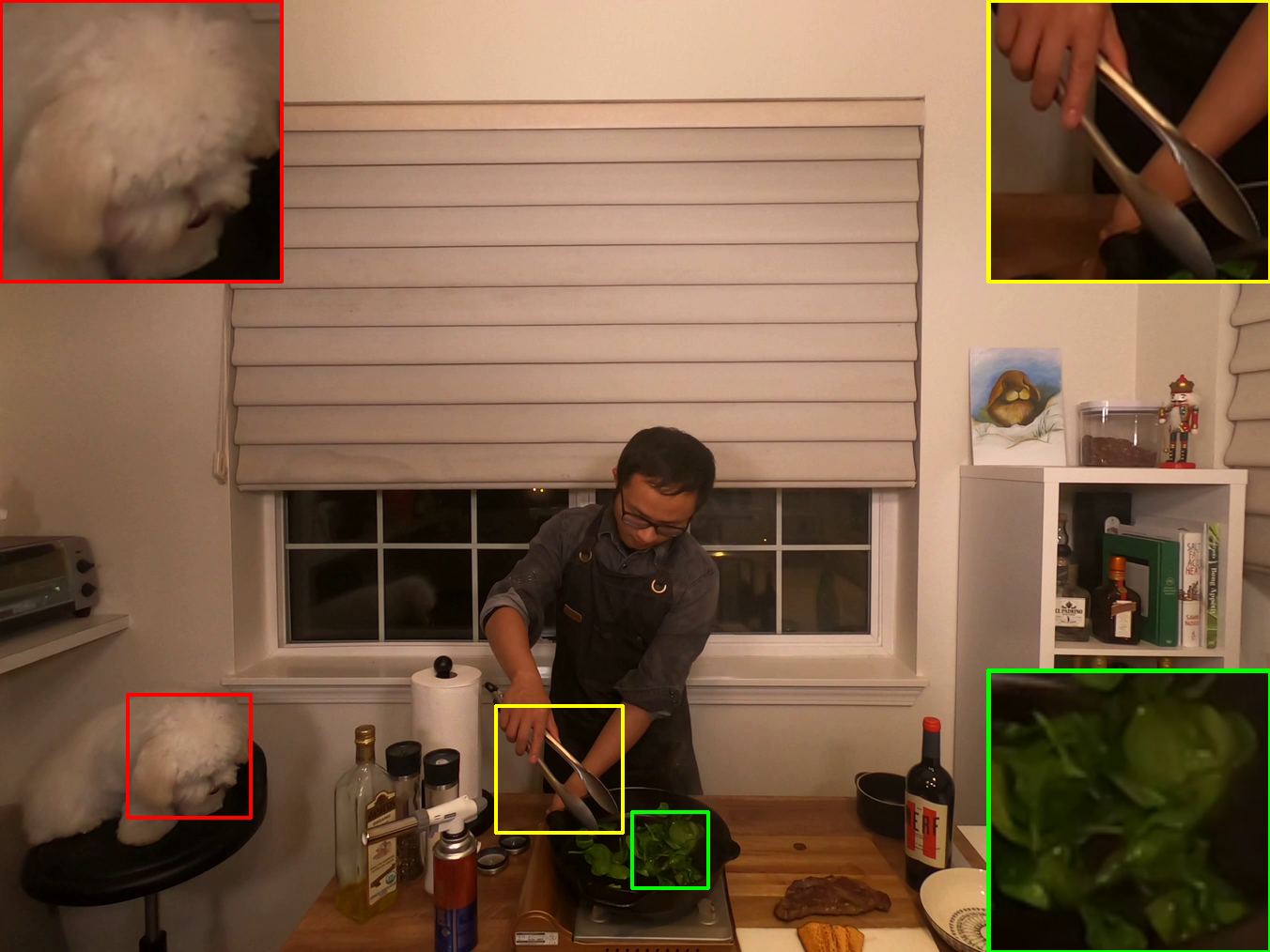}
   }   
   \caption{Qualitative comparison of ours with the benchmark methods on \textit{Cook Spinach} scene of N3DV dataset.}
   \label{fig:n3dv}
\end{figure*}

\begin{figure*}[!t]
   \centering
   \subfloat[3DGStream]{
      \includegraphics[width=0.192\textwidth]{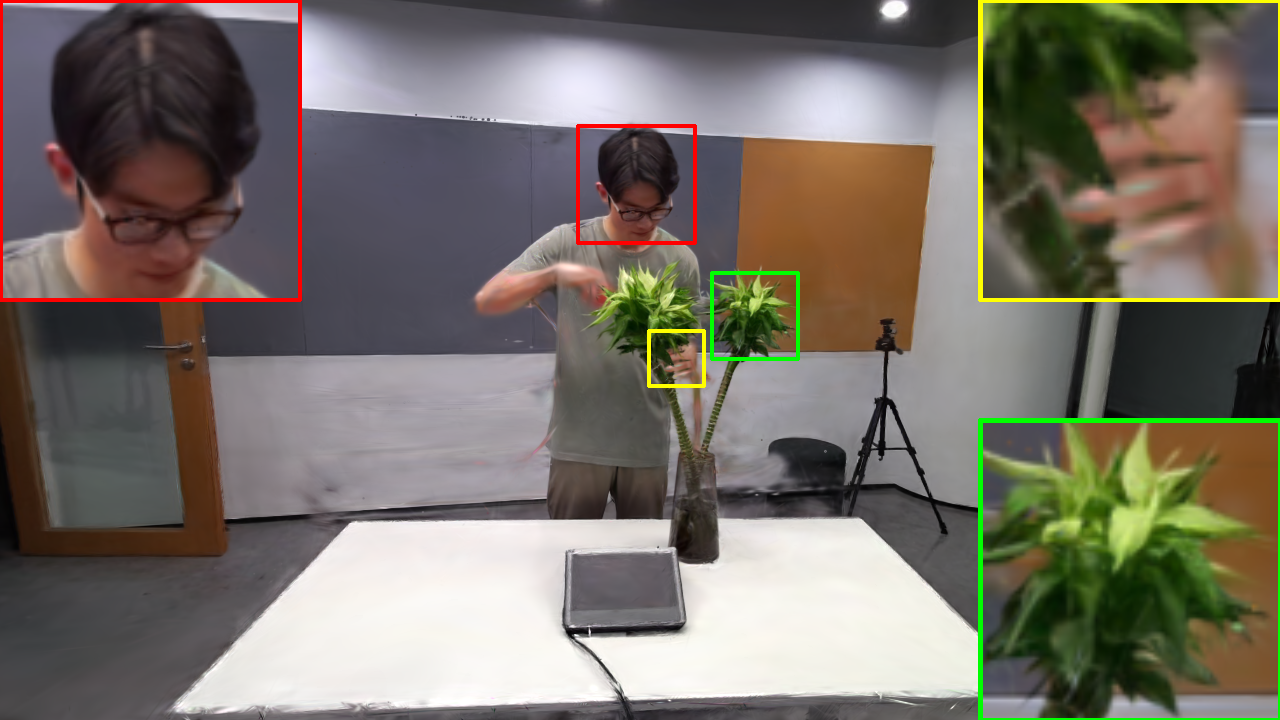}
   }
   \subfloat[HiCoM]{
      \includegraphics[width=0.192\textwidth]{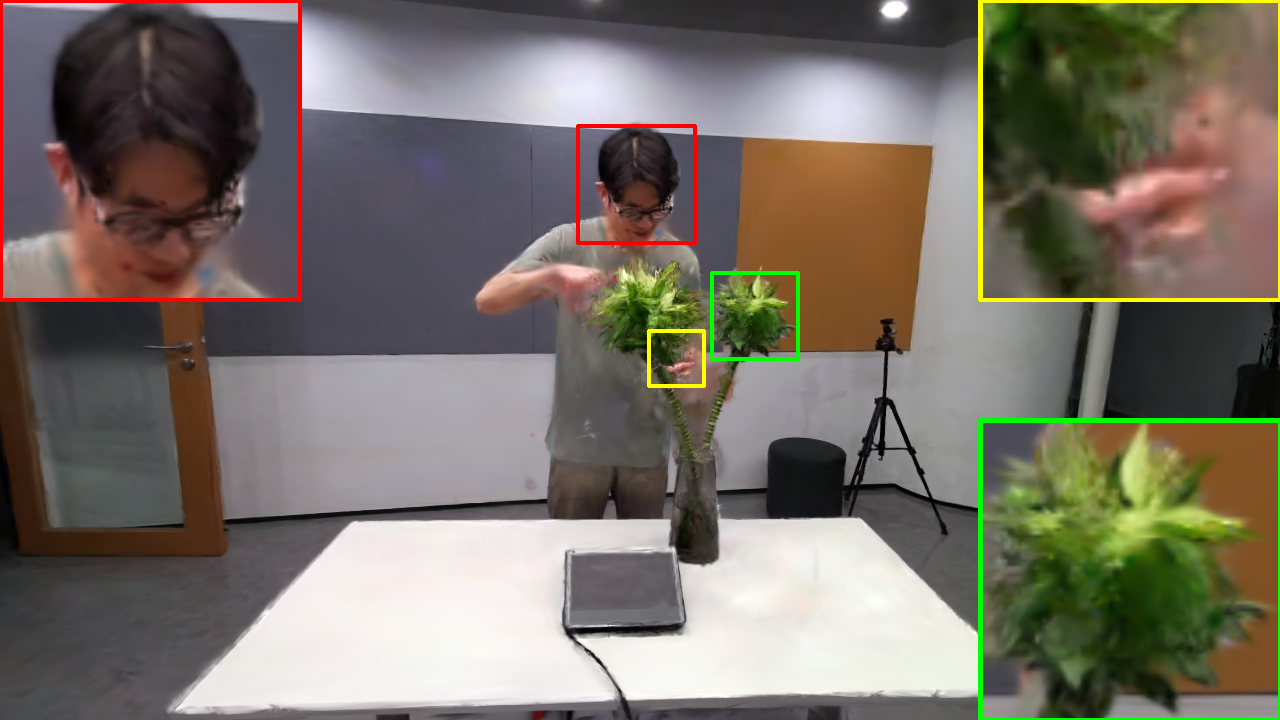}
   }
   \subfloat[4DGC]{
      \includegraphics[width=0.192\textwidth]{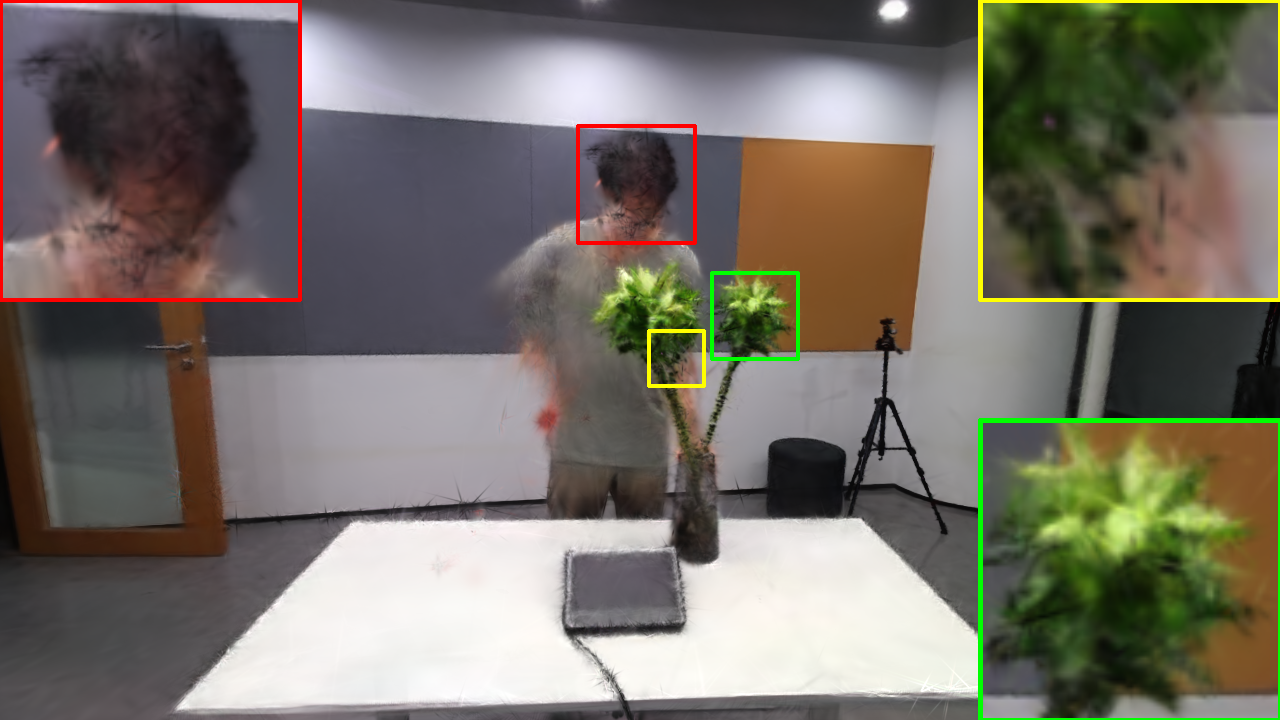}
   }
   \subfloat[Ours]{
      \includegraphics[width=0.192\textwidth]{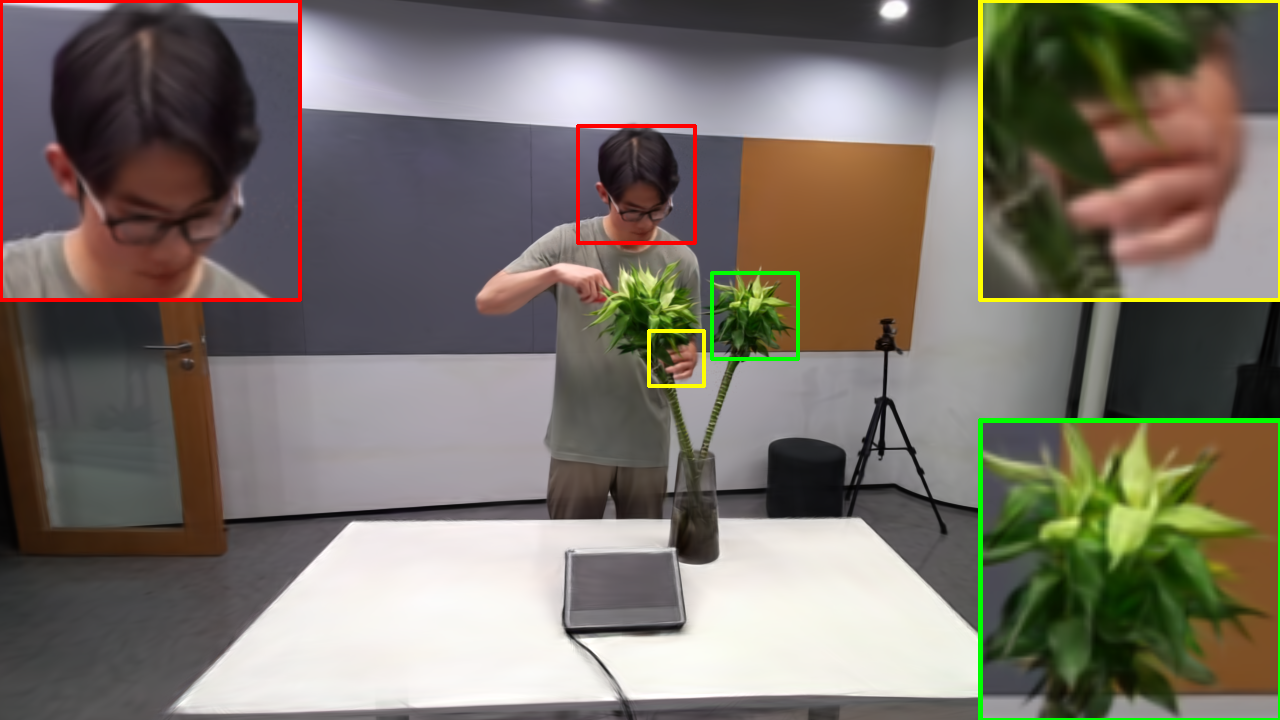}
   }
   \subfloat[GroudTruth]{
      \includegraphics[width=0.192\textwidth]{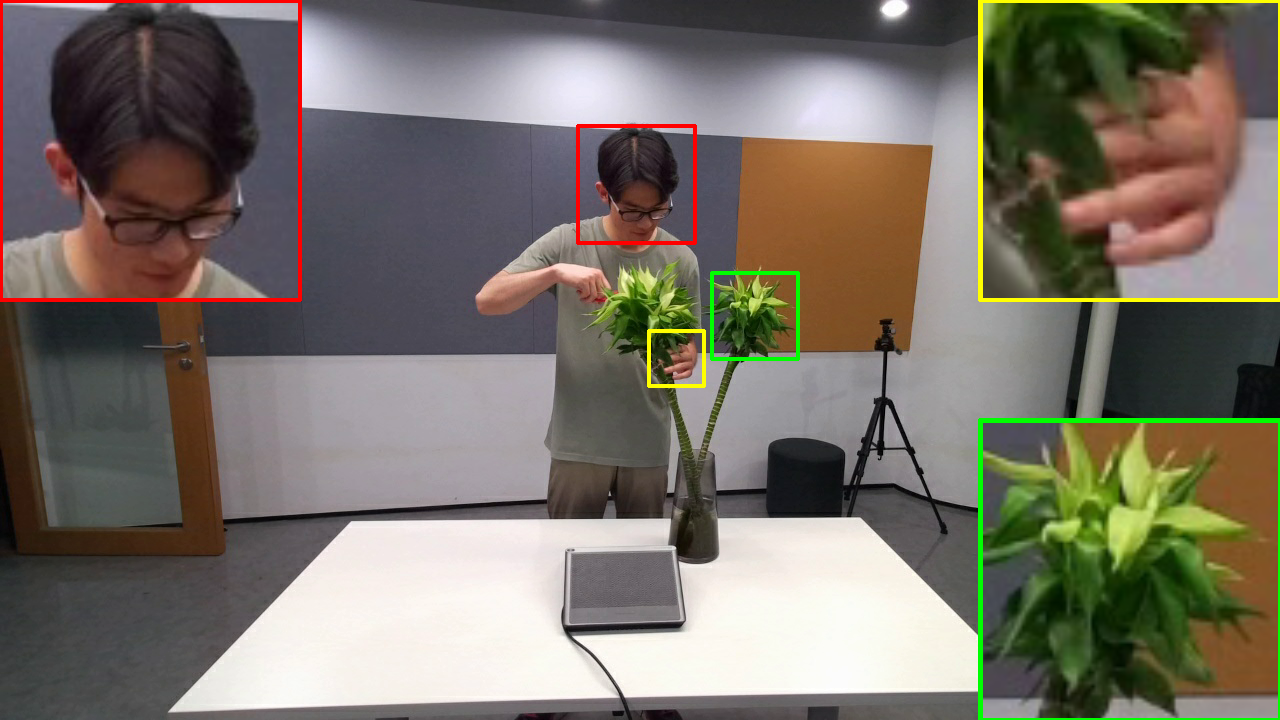}
   }
   \caption{Qualitative comparison of ours with the benchmark methods on \textit{Trimming} scene of MeetRoom dataset.}
   \label{fig:meet}

\end{figure*}

\subsection{Quantitative Comparisons}
We conduct a comprehensive quantitative comparison with state-of-the-art streamable FVV methods, including TeTriRF~\cite{wu2024tetrirf}, 3DGStream~\cite{sun20243dgstream}, VideoGS~\cite{wang2024v}, HiCoM~\cite{gao2024hicom}, and 4DGC~\cite{hu20254dgc}. First, we evaluate reconstruction quality by PSNR, SSIM, and LPIPS(VGG). As presented results in Table~\ref{tab:n3dv} and Table~\ref{tab:meet}, our StreamSTGS achieves the best reconstruction quality on both N3DV and MeetRoom datasets. In the meanwhile, the average frame size of StreamSTGS is only $170$KB, which is $4X$ smaller than that of GS-based methods. Though TeTriRF achieve the smallest frame size, its reconstruction quality and FPS are inferior to StreamSTGS. Furthermore, since users may not start watching FVV from the beginning, they must wait for previous $n$ frames to be decoded or for keyframes with large amounts of data to be transmitted. Therefore, we also compare Key-Frame~(K.F.) size and decoding delay. As shown in Table~\ref{tab:n3dv} and Table~\ref{tab:meet}, our StreamSTGS demonstrates competitive performance in these metrics. More quantitative results are provided in the \textbf{Appendix}.

\subsection{Qualitative Comparisons}

\begin{figure*}[!t]
   \centering

   \subfloat[Full model]{
      \includegraphics[width=0.192\textwidth]{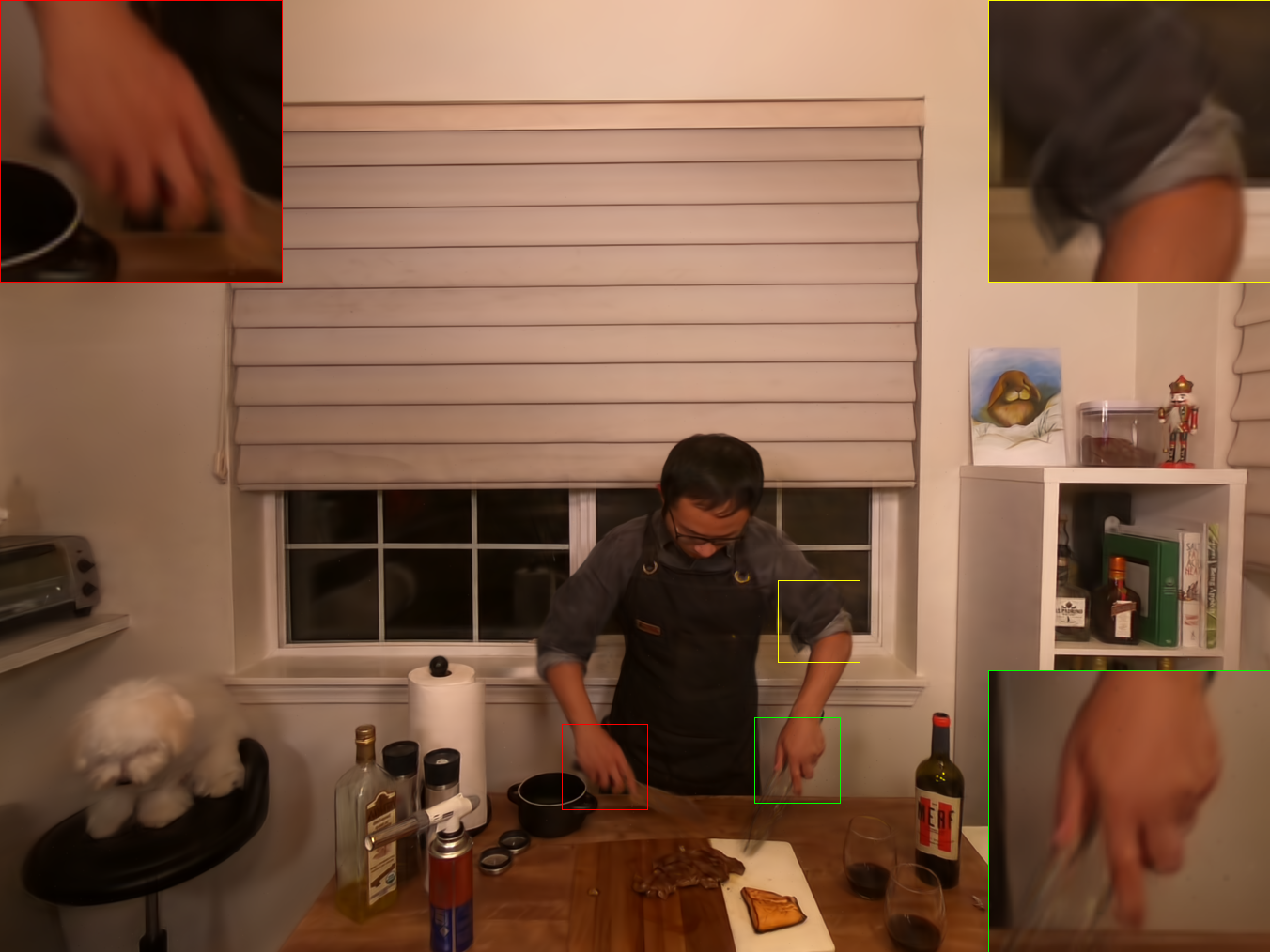}
   }
   \subfloat[w/o Aux. tra.]{
      \includegraphics[width=0.192\textwidth]{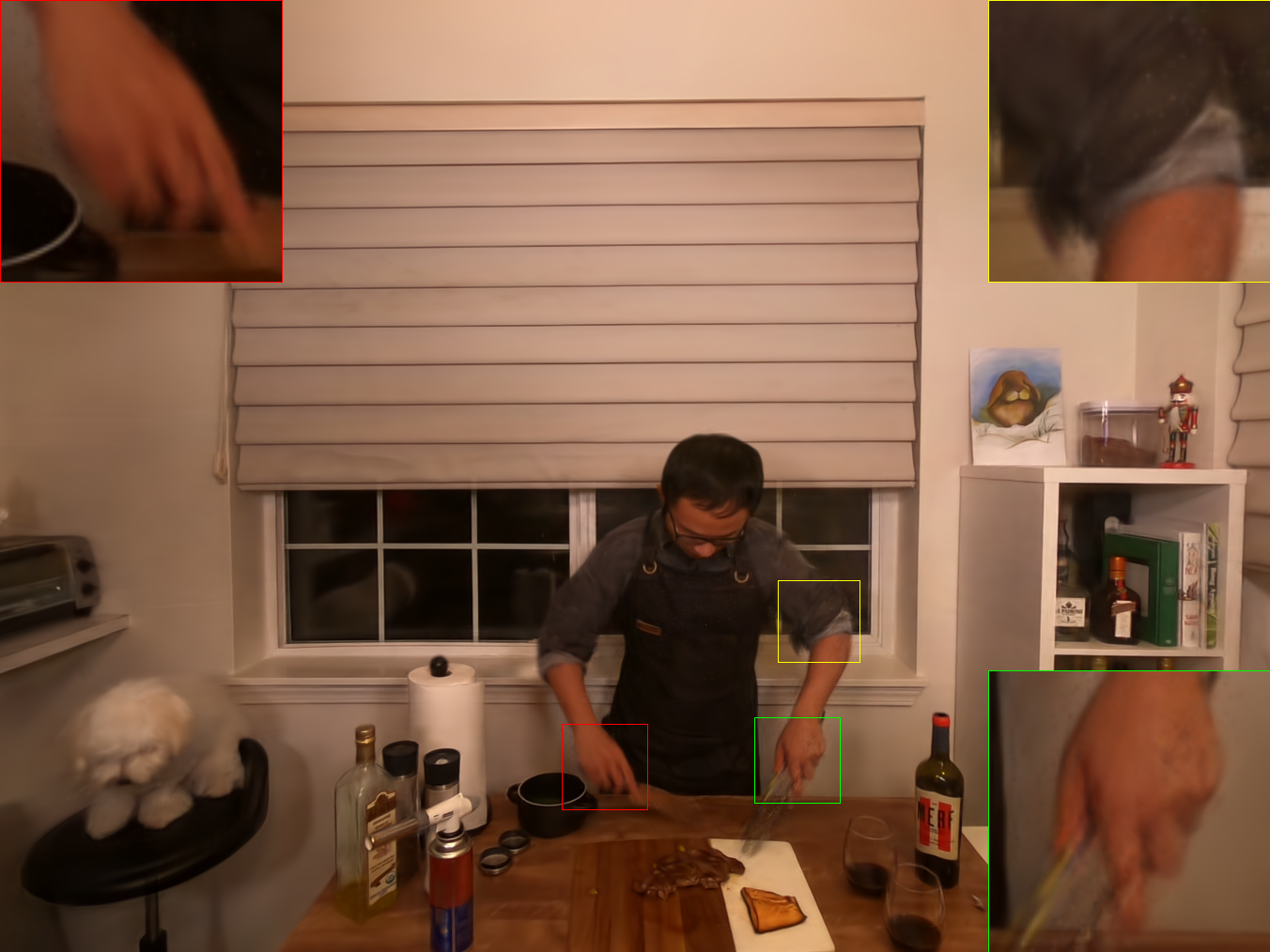}
   }
   \subfloat[w/o Dyn. den.]{
      \includegraphics[width=0.192\textwidth]{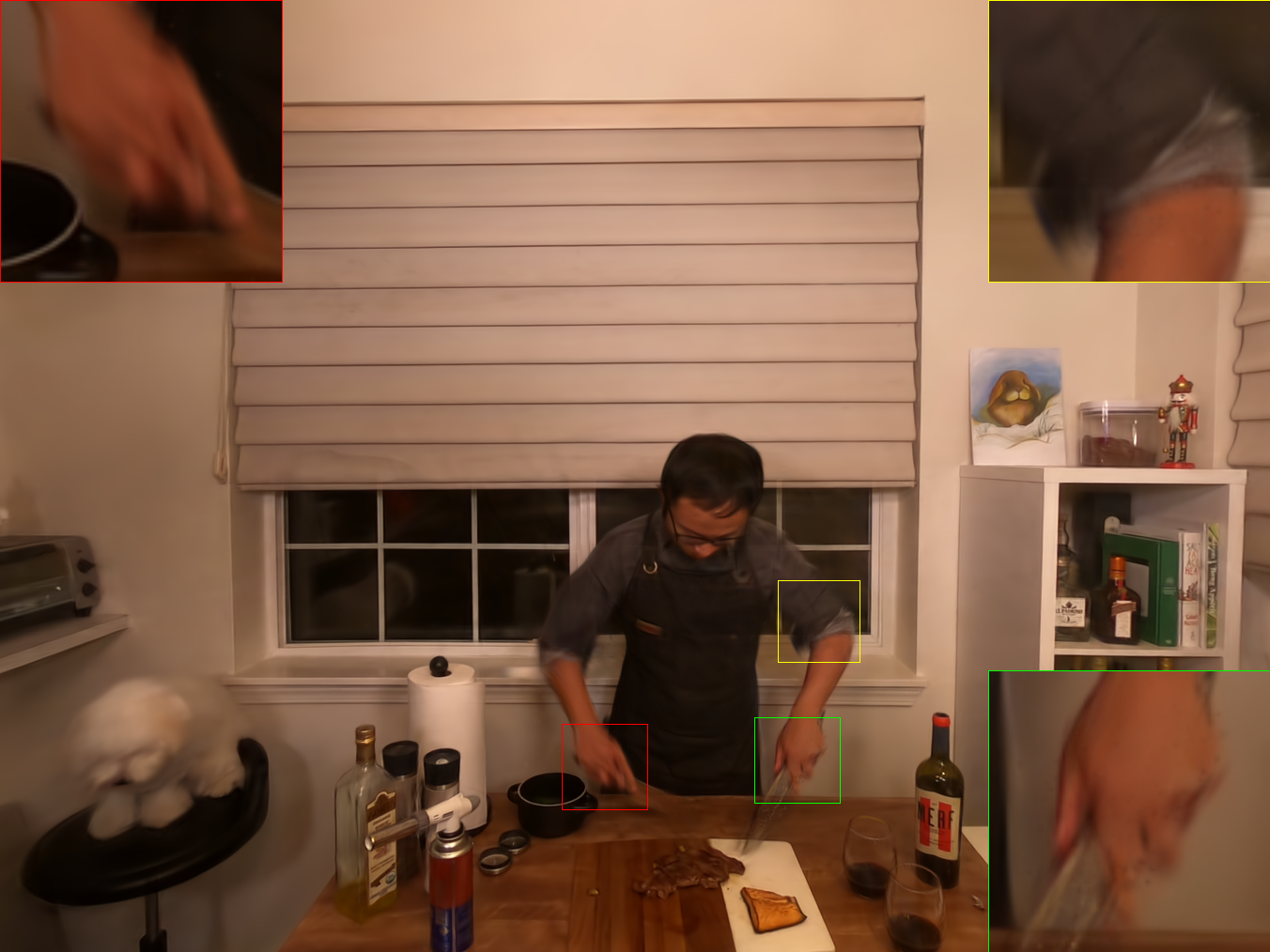}
   }
   \subfloat[w/o Temp. reg.]{
      \includegraphics[width=0.192\textwidth]{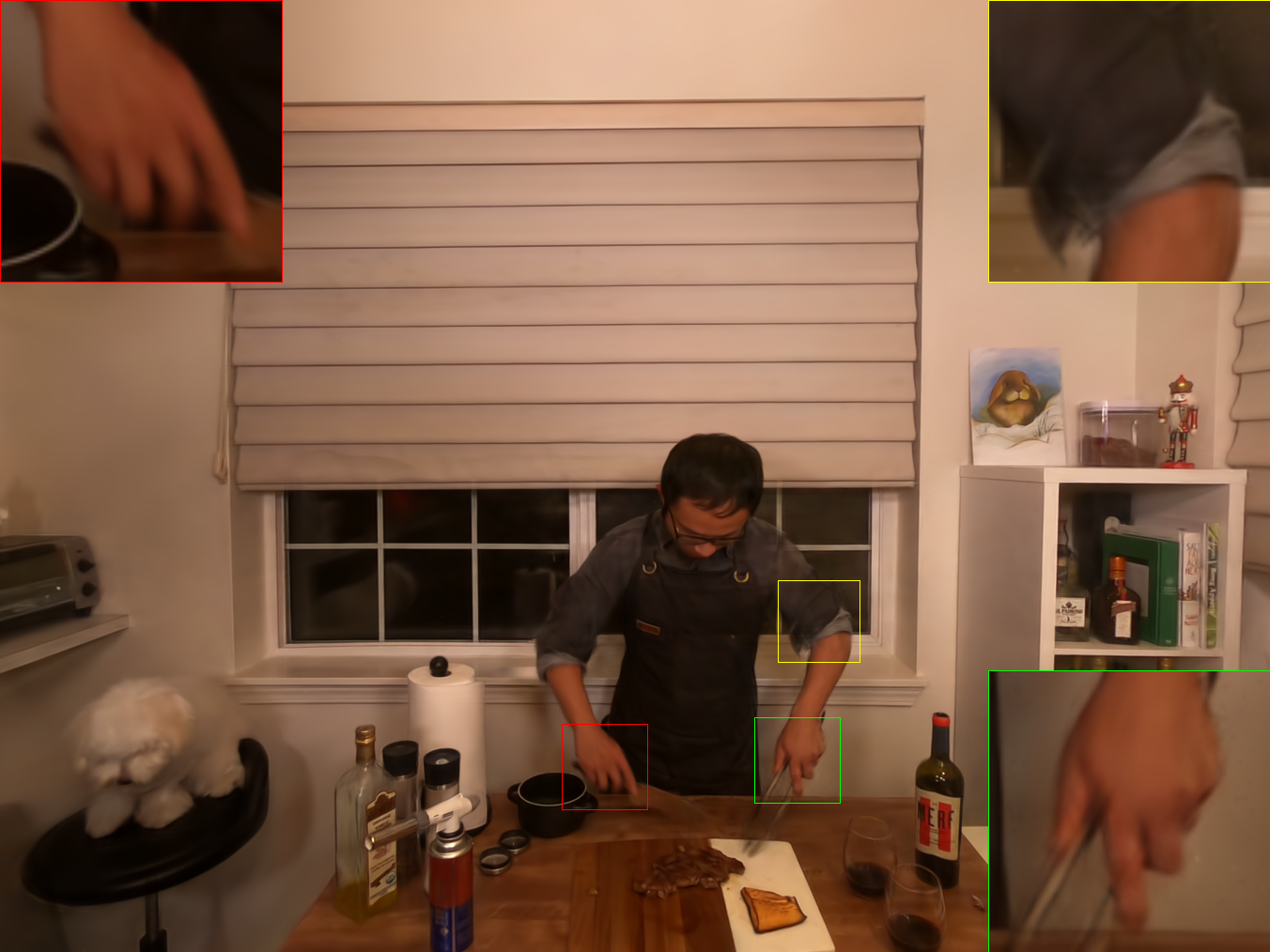}
   } 
    \subfloat[w/o Relocate]{
      \includegraphics[width=0.192\textwidth]{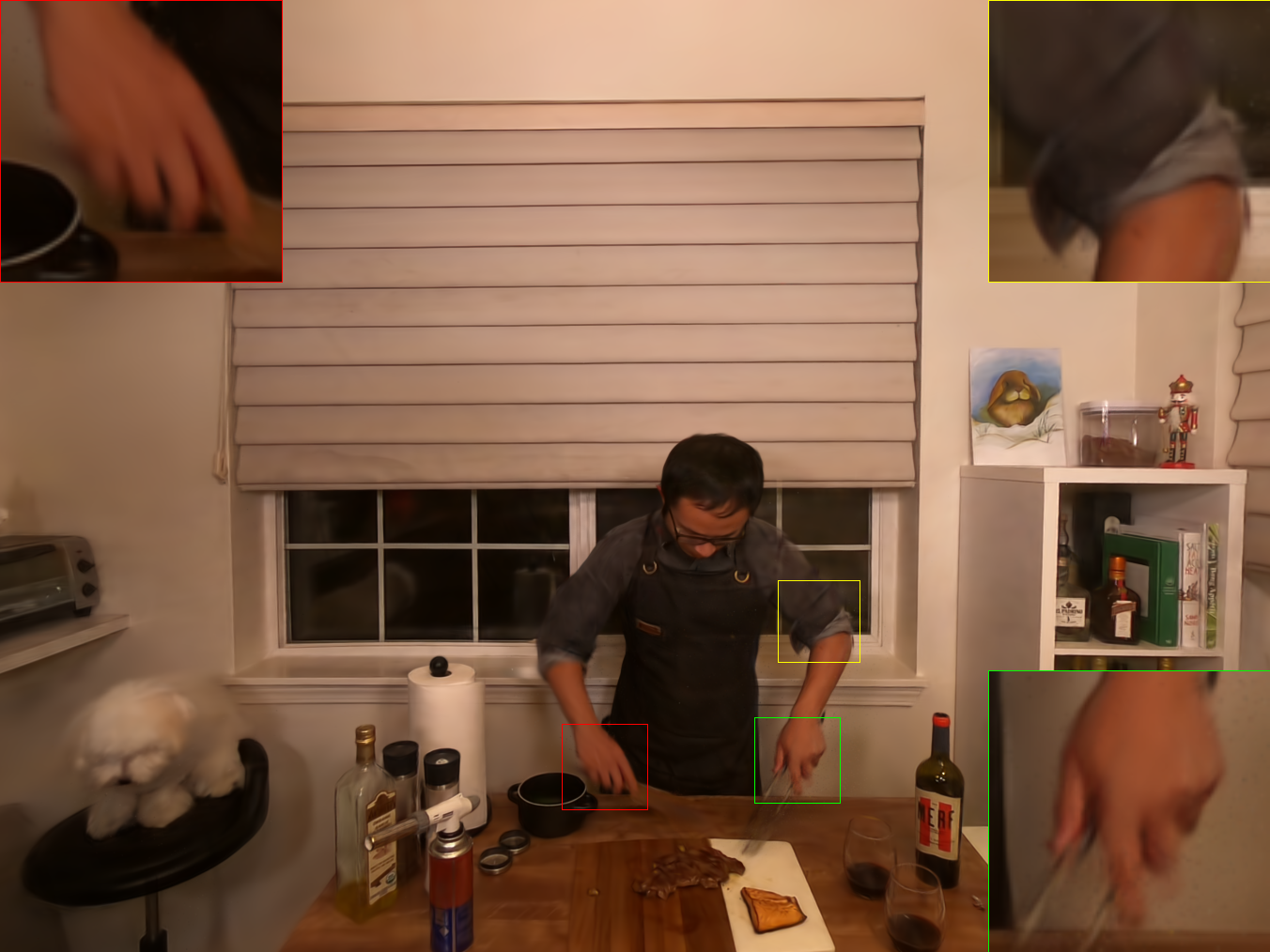}
   } 
   \caption{Ablation experiments on \textit{Cut Roasted Beef} scene of N3DV dataset.}
   \label{fig:ablations}

\end{figure*}

Fig.~\ref{fig:n3dv} and Fig.~\ref{fig:meet} present the qualitative results of our StreamSTGS compared to benchmark methods. It is evident that StreamSTGS achieves significantly improvements in dynamic areas, owing to the proposed Transformer-guided auxiliary module that enhances the learning of motion. Notably, in the area of complex human interaction, such as \textit{yellow} region in both Fig.~\ref{fig:n3dv} and Fig.~\ref{fig:meet}, our StreamSTGS accurately reconstructs hands and objects, whereas comparative methods yield blurred results. This improvements can be attributed to our dynamic-aware density strategy, which directs a higher concentration of 3D Gaussians toward dynamic areas. These Gaussians are subsequently optimized to model motion patterns more effectively through Transformer module. Besides, our StreamSTGS successfully reconstructs high-light effect, demonstrating that our simplified color model not only reduces storage but also preserves reconstruction quality. 

\subsection{Ablations}

\textbf{Compression QP}. We evaluate the performance of our StreamSTGS under different compression QP parameters using libx265, with results presented in Table.~\ref{tab:qp}. It can be observed that setting QP to $20$ achieves an optimal tradeoff between reconstruction quality and storage. Even when the QP is set to 28 or 32, our StreamSTGS still outperforms benchmark methods in reconstruction quality while significantly reducing storage size.

\begin{table}[!t]
   \centering
   \resizebox{\columnwidth}{!}{%
   \begin{tabular}{ccccc}
      \toprule
      \multirow{2}{*}{QP} & \multicolumn{2}{c}{\textbf{N3DV}} & \multicolumn{2}{c}{\textbf{MeetRoom}} \\ \cmidrule(lr){2-3} \cmidrule(lr){4-5}
                          & PSNR$\uparrow$  & Storage$\downarrow$  & PSNR$\uparrow$    & Storage$\downarrow$    \\ 
      \midrule
      16                  & 32.36     & 247.52       & 27.46       & 208.71         \\
      20                  & 32.30     & 173.59       & 27.41       & 142.53         \\
      24                  & 32.15     & 121.97       & 27.32       & 97.06          \\
      28                  & 31.68     & 87.95        & 27.02       & 68.25          \\
      32                  & 30.76     & 68.35        & 26.46       & 52.34          \\ 
      \bottomrule
      \end{tabular}%
   }
   \caption{Performance under different compression QP.}
   \label{tab:qp}
\end{table}

\begin{table}[!t]
   \resizebox{\columnwidth}{!}{%
   \begin{tabular}{cccccc}
      \toprule
      \begin{tabular}[c]{@{}c@{}}Sliding Window\\ Size \end{tabular} & PSNR  & SSIM  & Storage & K.F. Size \\ 
      \midrule
      W = 1                                                          & 32.01 & 0.941 & 298.06  & 3.92      \\
      W = 3                                                          & 32.30 & 0.944 & 173.59  & 3.86      \\
      W = 5                                                          & 32.26 & 0.944 & 176.05  & 3.86      \\ 
      \bottomrule
      \end{tabular}%
   }
   \caption{Performance under different sizes of sliding window on N3DV dataset.}
   \label{tab:sws}
\end{table}

\noindent \textbf{Sliding Window}. As shown in Table~\ref{tab:sws}, we evaluate the performance of our StreamSTGS under different sliding window sizes $W$. When the sliding window scheme is removed~(i.e., $W=1$), both the reconstruction quality and the storage efficiency degrade, as the relationships between adjacent temporal features cannot be captured and optimized during training. Note that increasing the size of sliding window to $5$ does not provide significant benefits, as learning motions across five frames is challenging.

\begin{table}[!t]
\centering
\small
\resizebox{\columnwidth}{!}{%
\begin{tabular}{c|cccc}
\toprule
          & PSNR  & SSIM  & \multicolumn{1}{l}{Storage} & Train \\ \midrule
GOP-30    & 32.32 & 0.944                     & 228.7                       & 133   \\
GOP-60(Ours)    & 32.30 & 0.943                     & 173.6                       & 67    \\
GOP-100   & 32.06 & 0.942                     & 161.9                       & 40    \\ \bottomrule
\end{tabular}%
}
\caption{Ablation of GOP length.}
\label{tab:gop}
\vspace{-2mm}
\end{table}

\noindent \textbf{GOP Length}. We also conduct ablation experiments on the GOP length. We set the GOP length as 30, 60, and 100, and the results are shown in the Table~\ref{tab:gop}. When GOP is 60, a good tradeoff is achieved in terms of quality, storage size, and training time. Since our method does not require training multiple models for different bitrates, sacrificing training time for higher performance is worthwhile. Existing methods exceed $60$s to train 3-6 models for variable bitrates.

\noindent \textbf{Key Components}.
We ablate four key components of our StreamSTGS as shown in Table.~\ref{tab:ablation}. In the second row, we do not apply Transformer-guided auxiliary training, which significantly reduced training costs. However, this leads to decreased reconstruction quality and motion blur, as illustrated in Fig.~\ref{fig:ablations}(b), due to the difficulty in learning global motions. In the third row, we disable the dynamic-aware density strategy, which causes blurry motions in dynamic areas as shown in Fig.~\ref{fig:ablations}(c). Correspondingly, the reduced storage in this case is because static Gaussians exhibit more temporal consistency, which improves compression efficiency. In the fourth row, we omit temporal regularization, which substantially increase storage size without any improvement in reconstruction quality as shown in Fig.~\ref{fig:ablations}(d). In the fifth row, we remove the Gaussian relocate strategy as demonstrated in Fig.~\ref{fig:ablations}(e), which prevents unnecessary Gaussians from being repositioned effectively, resulting in local optima.

\begin{table}[!t]
   \resizebox{\columnwidth}{!}{%
   \begin{tabular}{cccc}
      \toprule
      Components             & PSNR  & Storage & Train  \\ 
      \midrule
      Full model             & 32.30 & 173.59  & 67     \\
      w/o Auxiliary training & 31.99 & 174.51  & 29     \\
      w/o Dynamic density    & 32.07 & 114.15  & 69     \\
      w/o Temporal reg.      & 32.23 & 319.48  & 64     \\
      w/o Gaussian relocate  & 32.11 & 169.70  & 68     \\
      \bottomrule
      \end{tabular}%
   }
   \caption{Effect of various components ablated on N3DV.}
   \label{tab:ablation}
   \vspace{-0.2cm}
\end{table}

\section{Conclusion and Limitation}

This paper introduces a novel streamable FVV representation, StreamSTGS, which represents canonical 3D Gaussians as 2D images and temporal features as a video, thereby resulting in a compact enough representation capable of meeting real-time requirements. Moreover, a Transformer-guided auxiliary training strategy is proposed to improve the learning of global motions. Additionally, several key improvements, such as sliding window-based temporal feature aggregation, dynamic-aware density, and Gaussian relocate strategy, collectively contribute to the superior performance of StreamSTGS. However, each 3D Gaussians in StreamSTGS needs temporal features, but some 3D Gaussians maintain static within a GOP. Therefore, by categorizing 3D Gaussians into static and dynamic sets and assigning temporal features exclusively to dynamic Gaussians, we can further reduces storage and improve FPS.

\bigskip

\bibliography{aaai2026}

@article{mildenhall2021nerf,
  title={Nerf: Representing scenes as neural radiance fields for view synthesis},
  author={Mildenhall, Ben and Srinivasan, Pratul P and Tancik, Matthew and Barron, Jonathan T and Ramamoorthi, Ravi and Ng, Ren},
  journal={Communications of the ACM},
  volume={65},
  number={1},
  pages={99--106},
  year={2021},
  publisher={ACM New York, NY, USA}
}

@article{kerbl20233d,
  title={3D Gaussian Splatting for Real-Time Radiance Field Rendering.},
  author={Kerbl, Bernhard and Kopanas, Georgios and Leimk{\"u}hler, Thomas and Drettakis, George},
  journal={ACM transactions on graphics (TOG)},
  volume={42},
  number={4},
  pages={139--1},
  year={2023}
}

@inproceedings{yu2021plenoctrees,
  title={Plenoctrees for real-time rendering of neural radiance fields},
  author={Yu, Alex and Li, Ruilong and Tancik, Matthew and Li, Hao and Ng, Ren and Kanazawa, Angjoo},
  booktitle={Proceedings of the IEEE/CVF International Conference on Computer Vision},
  pages={5752--5761},
  year={2021}
}

@article{muller2022instant,
  title={Instant neural graphics primitives with a multiresolution hash encoding},
  author={M{\"u}ller, Thomas and Evans, Alex and Schied, Christoph and Keller, Alexander},
  journal={ACM transactions on graphics (TOG)},
  volume={41},
  number={4},
  pages={1--15},
  year={2022},
  publisher={ACM New York, NY, USA}
}

@inproceedings{park2021nerfies,
  title={Nerfies: Deformable neural radiance fields},
  author={Park, Keunhong and Sinha, Utkarsh and Barron, Jonathan T and Bouaziz, Sofien and Goldman, Dan B and Seitz, Steven M and Martin-Brualla, Ricardo},
  booktitle={Proceedings of the IEEE/CVF International Conference on Computer Vision},
  pages={5865--5874},
  year={2021}
}

@inproceedings{fang2022fast,
  title={Fast dynamic radiance fields with time-aware neural voxels},
  author={Fang, Jiemin and Yi, Taoran and Wang, Xinggang and Xie, Lingxi and Zhang, Xiaopeng and Liu, Wenyu and Nie{\ss}ner, Matthias and Tian, Qi},
  booktitle={SIGGRAPH Asia 2022 Conference Papers},
  pages={1--9},
  year={2022}
}

@article{park2021hypernerf,
  author = {Park, Keunhong and Sinha, Utkarsh and Hedman, Peter and Barron, Jonathan T. and Bouaziz, Sofien and Goldman, Dan B and Martin-Brualla, Ricardo and Seitz, Steven M.},
  title = {HyperNeRF: A Higher-Dimensional Representation for Topologically Varying Neural Radiance Fields},
  journal = {ACM transactions on graphics (TOG)},
  issue_date = {December 2021},
  publisher = {ACM},
  volume = {40},
  number = {6},
  month = {dec},
  year = {2021},
  articleno = {238},
}

@inproceedings{chen2022tensorf,
  title={Tensorf: Tensorial radiance fields},
  author={Chen, Anpei and Xu, Zexiang and Geiger, Andreas and Yu, Jingyi and Su, Hao},
  booktitle={European conference on computer vision},
  pages={333--350},
  year={2022},
  organization={Springer}
}

@inproceedings{du2021neural,
  title={Neural radiance flow for 4d view synthesis and video processing},
  author={Du, Yilun and Zhang, Yinan and Yu, Hong-Xing and Tenenbaum, Joshua B and Wu, Jiajun},
  booktitle={2021 IEEE/CVF International Conference on Computer Vision (ICCV)},
  pages={14304--14314},
  year={2021},
  organization={IEEE Computer Society}
}

@inproceedings{xian2021space,
  title={Space-time neural irradiance fields for free-viewpoint video},
  author={Xian, Wenqi and Huang, Jia-Bin and Kopf, Johannes and Kim, Changil},
  booktitle={Proceedings of the IEEE/CVF conference on computer vision and pattern recognition},
  pages={9421--9431},
  year={2021}
}

@inproceedings{li2022nv3d,
  title={Neural 3d video synthesis from multi-view video},
  author={Li, Tianye and Slavcheva, Mira and Zollhoefer, Michael and Green, Simon and Lassner, Christoph and Kim, Changil and Schmidt, Tanner and Lovegrove, Steven and Goesele, Michael and Newcombe, Richard and others},
  booktitle={Proceedings of the IEEE/CVF Conference on Computer Vision and Pattern Recognition},
  pages={5521--5531},
  year={2022}
}

@article{li2022streaming,
  title={Streaming radiance fields for 3d video synthesis},
  author={Li, Lingzhi and Shen, Zhen and Wang, Zhongshu and Shen, Li and Tan, Ping},
  journal={Advances in Neural Information Processing Systems},
  volume={35},
  pages={13485--13498},
  year={2022}
}

@inproceedings{wang2022mixed,
  title={Mixed neural voxels for fast multi-view video synthesis},
  author={Wang, Feng and Tan, Sinan and Li, Xinghang and Tian, Zeyue and Song, Yafei and Liu, Huaping},
  booktitle={Proceedings of the IEEE/CVF International Conference on Computer Vision},
  pages={19706--19716},
  year={2023}
}

@inproceedings{fridovich2023k,
  title={K-planes: Explicit radiance fields in space, time, and appearance},
  author={Fridovich-Keil, Sara and Meanti, Giacomo and Warburg, Frederik Rahb{\ae}k and Recht, Benjamin and Kanazawa, Angjoo},
  booktitle={Proceedings of the IEEE/CVF Conference on Computer Vision and Pattern Recognition},
  pages={12479--12488},
  year={2023}
}

@inproceedings{cao2023hexplane,
  title={Hexplane: A fast representation for dynamic scenes},
  author={Cao, Ang and Johnson, Justin},
  booktitle={Proceedings of the IEEE/CVF Conference on Computer Vision and Pattern Recognition},
  pages={130--141},
  year={2023}
}

@inproceedings{wu2024tetrirf,
  title={TeTriRF: Temporal Tri-Plane Radiance Fields for Efficient Free-Viewpoint Video},
  author={Wu, Minye and Wang, Zehao and Kouros, Georgios and Tuytelaars, Tinne},
  booktitle={Proceedings of the IEEE/CVF Conference on Computer Vision and Pattern Recognition},
  pages={6487--6496},
  year={2024}
}

@article{wang2024masked,
  title={Masked space-time hash encoding for efficient dynamic scene reconstruction},
  author={Wang, Feng and Chen, Zilong and Wang, Guokang and Song, Yafei and Liu, Huaping},
  journal={Advances in Neural Information Processing Systems},
  volume={36},
  year={2024}
}

@article{song2023nerfplayer,
  title={Nerfplayer: A streamable dynamic scene representation with decomposed neural radiance fields},
  author={Song, Liangchen and Chen, Anpei and Li, Zhong and Chen, Zhang and Chen, Lele and Yuan, Junsong and Xu, Yi and Geiger, Andreas},
  journal={IEEE Transactions on Visualization and Computer Graphics},
  volume={29},
  number={5},
  pages={2732--2742},
  year={2023},
  publisher={IEEE}
}

@inproceedings{attal2023hyperreel,
  title={Hyperreel: High-fidelity 6-dof video with ray-conditioned sampling},
  author={Attal, Benjamin and Huang, Jia-Bin and Richardt, Christian and Zollhoefer, Michael and Kopf, Johannes and O’Toole, Matthew and Kim, Changil},
  booktitle={Proceedings of the IEEE/CVF Conference on Computer Vision and Pattern Recognition},
  pages={16610--16620},
  year={2023}
}

@inproceedings{lou2024darenerf,
  title={DaReNeRF: Direction-aware Representation for Dynamic Scenes},
  author={Lou, Ange and Planche, Benjamin and Gao, Zhongpai and Li, Yamin and Luan, Tianyu and Ding, Hao and Chen, Terrence and Noble, Jack and Wu, Ziyan},
  booktitle={Proceedings of the IEEE/CVF Conference on Computer Vision and Pattern Recognition},
  pages={5031--5042},
  year={2024}
}

@inproceedings{yang2025dmit,
  title={DMiT: Deformable Mipmapped Tri-Plane Representation for Dynamic Scenes},
  author={Yang, Jing-Wen and Sun, Jia-Mu and Yang, Yong-Liang and Yang, Jie and Shan, Ying and Cao, Yan-Pei and Gao, Lin},
  booktitle={European Conference on Computer Vision},
  pages={436--453},
  year={2025},
  organization={Springer}
}

@inproceedings{zhan2024kfd,
  title={KFD-NeRF: Rethinking Dynamic NeRF with Kalman Filter},
  author={Zhan, Yifan and Li, Zhuoxiao and Niu, Muyao and Zhong, Zhihang and Nobuhara, Shohei and Nishino, Ko and Zheng, Yinqiang},
  booktitle={European Conference on Computer Vision},
  year={2024},
  publisher={Springer}
}

@inproceedings{guo2023forward,
  title={Forward flow for novel view synthesis of dynamic scenes},
  author={Guo, Xiang and Sun, Jiadai and Dai, Yuchao and Chen, Guanying and Ye, Xiaoqing and Tan, Xiao and Ding, Errui and Zhang, Yumeng and Wang, Jingdong},
  booktitle={Proceedings of the IEEE/CVF International Conference on Computer Vision},
  pages={16022--16033},
  year={2023}
}

@inproceedings{yan2023nerf,
  title={Nerf-ds: Neural radiance fields for dynamic specular objects},
  author={Yan, Zhiwen and Li, Chen and Lee, Gim Hee},
  booktitle={Proceedings of the IEEE/CVF Conference on Computer Vision and Pattern Recognition},
  pages={8285--8295},
  year={2023}
}

@inproceedings{wang2023f2,
  title={F2-nerf: Fast neural radiance field training with free camera trajectories},
  author={Wang, Peng and Liu, Yuan and Chen, Zhaoxi and Liu, Lingjie and Liu, Ziwei and Komura, Taku and Theobalt, Christian and Wang, Wenping},
  booktitle={Proceedings of the IEEE/CVF Conference on Computer Vision and Pattern Recognition},
  pages={4150--4159},
  year={2023}
}

@inproceedings{park2023temporal,
  title={Temporal interpolation is all you need for dynamic neural radiance fields},
  author={Park, Sungheon and Son, Minjung and Jang, Seokhwan and Ahn, Young Chun and Kim, Ji-Yeon and Kang, Nahyup},
  booktitle={Proceedings of the IEEE/CVF Conference on Computer Vision and Pattern Recognition},
  pages={4212--4221},
  year={2023}
}

@inproceedings{liu2023robust,
  title={Robust dynamic radiance fields},
  author={Liu, Yu-Lun and Gao, Chen and Meuleman, Andreas and Tseng, Hung-Yu and Saraf, Ayush and Kim, Changil and Chuang, Yung-Yu and Kopf, Johannes and Huang, Jia-Bin},
  booktitle={Proceedings of the IEEE/CVF Conference on Computer Vision and Pattern Recognition},
  pages={13--23},
  year={2023}
}

@inproceedings{tian2023mononerf,
  title={Mononerf: Learning a generalizable dynamic radiance field from monocular videos},
  author={Tian, Fengrui and Du, Shaoyi and Duan, Yueqi},
  booktitle={Proceedings of the IEEE/CVF International Conference on Computer Vision},
  pages={17903--17913},
  year={2023}
}

@inproceedings{shao2023tensor4d,
  title={Tensor4d: Efficient neural 4d decomposition for high-fidelity dynamic reconstruction and rendering},
  author={Shao, Ruizhi and Zheng, Zerong and Tu, Hanzhang and Liu, Boning and Zhang, Hongwen and Liu, Yebin},
  booktitle={Proceedings of the IEEE/CVF Conference on Computer Vision and Pattern Recognition},
  pages={16632--16642},
  year={2023}
}

@inproceedings{li2021neural,
  title={Neural scene flow fields for space-time view synthesis of dynamic scenes},
  author={Li, Zhengqi and Niklaus, Simon and Snavely, Noah and Wang, Oliver},
  booktitle={Proceedings of the IEEE/CVF Conference on Computer Vision and Pattern Recognition},
  pages={6498--6508},
  year={2021}
}

@inproceedings{gao2021dynamic,
  title={Dynamic view synthesis from dynamic monocular video},
  author={Gao, Chen and Saraf, Ayush and Kopf, Johannes and Huang, Jia-Bin},
  booktitle={Proceedings of the IEEE/CVF International Conference on Computer Vision},
  pages={5712--5721},
  year={2021}
}

@inproceedings{luiten2024dynamic,
  title={Dynamic 3d gaussians: Tracking by persistent dynamic view synthesis},
  author={Luiten, Jonathon and Kopanas, Georgios and Leibe, Bastian and Ramanan, Deva},
  booktitle={2024 International Conference on 3D Vision (3DV)},
  pages={800--809},
  year={2024},
  organization={IEEE}
}

@inproceedings{bae2025per,
  title={Per-gaussian embedding-based deformation for deformable 3d gaussian splatting},
  author={Bae, Jeongmin and Kim, Seoha and Yun, Youngsik and Lee, Hahyun and Bang, Gun and Uh, Youngjung},
  booktitle={European Conference on Computer Vision},
  pages={321--335},
  year={2025},
  organization={Springer}
}

@inproceedings{lu20243d,
  title={3d geometry-aware deformable gaussian splatting for dynamic view synthesis},
  author={Lu, Zhicheng and Guo, Xiang and Hui, Le and Chen, Tianrui and Yang, Min and Tang, Xiao and Zhu, Feng and Dai, Yuchao},
  booktitle={Proceedings of the IEEE/CVF Conference on Computer Vision and Pattern Recognition},
  pages={8900--8910},
  year={2024}
}

@inproceedings{shaw2024swings,
  title={Swings: sliding windows for dynamic 3D gaussian splatting},
  author={Shaw, Richard and Nazarczuk, Michal and Song, Jifei and Moreau, Arthur and Catley-Chandar, Sibi and Dhamo, Helisa and P{\'e}rez-Pellitero, Eduardo},
  year={2024},
  booktitle={European Conference on Computer Vision},
  organization={Springer}
}

@inproceedings{yang2024deformable,
  title={Deformable 3d gaussians for high-fidelity monocular dynamic scene reconstruction},
  author={Yang, Ziyi and Gao, Xinyu and Zhou, Wen and Jiao, Shaohui and Zhang, Yuqing and Jin, Xiaogang},
  booktitle={Proceedings of the IEEE/CVF Conference on Computer Vision and Pattern Recognition},
  pages={20331--20341},
  year={2024}
}

@inproceedings{huang2024sc,
  title={SC-GS: Sparse-controlled gaussian splatting for editable dynamic scenes},
  author={Huang, Yi-Hua and Sun, Yang-Tian and Yang, Ziyi and Lyu, Xiaoyang and Cao, Yan-Pei and Qi, Xiaojuan},
  booktitle={Proceedings of the IEEE/CVF Conference on Computer Vision and Pattern Recognition},
  pages={4220--4230},
  year={2024}
}

@inproceedings{zhao2024gaussianprediction,
  title={Gaussianprediction: Dynamic 3d gaussian prediction for motion extrapolation and free view synthesis},
  author={Zhao, Boming and Li, Yuan and Sun, Ziyu and Zeng, Lin and Shen, Yujun and Ma, Rui and Zhang, Yinda and Bao, Hujun and Cui, Zhaopeng},
  booktitle={ACM SIGGRAPH 2024 Conference Papers},
  pages={1--12},
  year={2024}
}

@inproceedings{wansuperpoint,
  title={Superpoint Gaussian Splatting for Real-Time High-Fidelity Dynamic Scene Reconstruction},
  author={Wan, Diwen and Lu, Ruijie and Zeng, Gang},
  booktitle={Forty-first International Conference on Machine Learning},
  year={2024}
}

@inproceedings{wu20244d,
  title={4d gaussian splatting for real-time dynamic scene rendering},
  author={Wu, Guanjun and Yi, Taoran and Fang, Jiemin and Xie, Lingxi and Zhang, Xiaopeng and Wei, Wei and Liu, Wenyu and Tian, Qi and Wang, Xinggang},
  booktitle={Proceedings of the IEEE/CVF Conference on Computer Vision and Pattern Recognition},
  pages={20310--20320},
  year={2024}
}

@article{lu2024dn,
  title={DN-4DGS: Denoised Deformable Network with Temporal-Spatial Aggregation for Dynamic Scene Rendering},
  author={Lu, Jiahao and Deng, Jiacheng and Zhu, Ruijie and Liang, Yanzhe and Yang, Wenfei and Zhang, Tianzhu and Zhou, Xu},
  journal={Advances in Neural Information Processing Systems},
  year={2024}
}

@article{duisterhof2023md,
  title={Md-splatting: Learning metric deformation from 4d gaussians in highly deformable scenes},
  author={Duisterhof, Bardienus P and Mandi, Zhao and Yao, Yunchao and Liu, Jia-Wei and Shou, Mike Zheng and Song, Shuran and Ichnowski, Jeffrey},
  journal={arXiv preprint arXiv:2312.00583},
  year={2023}
}

@inproceedings{li2024spacetime,
  title={Spacetime gaussian feature splatting for real-time dynamic view synthesis},
  author={Li, Zhan and Chen, Zhang and Li, Zhong and Xu, Yi},
  booktitle={Proceedings of the IEEE/CVF Conference on Computer Vision and Pattern Recognition},
  pages={8508--8520},
  year={2024}
}

@inproceedings{katsumata2025compact,
  title={A compact dynamic 3d gaussian representation for real-time dynamic view synthesis},
  author={Katsumata, Kai and Vo, Duc Minh and Nakayama, Hideki},
  booktitle={European Conference on Computer Vision},
  pages={394--412},
  year={2025},
  organization={Springer}
}

@inproceedings{lin2024gaussian,
  title={Gaussian-flow: 4d reconstruction with dynamic 3d gaussian particle},
  author={Lin, Youtian and Dai, Zuozhuo and Zhu, Siyu and Yao, Yao},
  booktitle={Proceedings of the IEEE/CVF Conference on Computer Vision and Pattern Recognition},
  pages={21136--21145},
  year={2024}
}

@article{xu2024grid4d,
  title={Grid4D: 4D Decomposed Hash Encoding for High-fidelity Dynamic Gaussian Splatting},
  author={Xu, Jiawei and Fan, Zexin and Yang, Jian and Xie, Jin},
  journal={Advances in Neural Information Processing Systems},
  year={2024}
}

@inproceedings{yan20244d,
  title={4D Gaussian Splatting with Scale-aware Residual Field and Adaptive Optimization for Real-time rendering of temporally complex dynamic scenes},
  author={Yan, Jinbo and Peng, Rui and Tang, Luyang and Wang, Ronggang},
  booktitle={Proceedings of the 32nd ACM International Conference on Multimedia},
  pages={7871--7880},
  year={2024}
}

@inproceedings{yang2023gs4d,
  title={Real-time Photorealistic Dynamic Scene Representation and Rendering with 4D Gaussian Splatting},
  author={Yang, Zeyu and Yang, Hongye and Pan, Zijie and Zhang, Li},
  booktitle = {International Conference on Learning Representations (ICLR)},
  year={2024}
}

@inproceedings{duan20244d,
  title={4d-rotor gaussian splatting: towards efficient novel view synthesis for dynamic scenes},
  author={Duan, Yuanxing and Wei, Fangyin and Dai, Qiyu and He, Yuhang and Chen, Wenzheng and Chen, Baoquan},
  booktitle={ACM SIGGRAPH 2024 Conference Papers},
  pages={1--11},
  year={2024}
}

@article{lee2024fully,
  title={Fully Explicit Dynamic Gaussian Splatting},
  author={Lee, Junoh and Won, Chang-Yeon and Jung, Hyunjun and Bae, Inhwan and Jeon, Hae-Gon},
  journal={Advances in Neural Information Processing Systems},
  year={2024}
}

@article{kim20244d,
  title={4D Gaussian Splatting in the Wild with Uncertainty-Aware Regularization},
  author={Kim, Mijeong and Lim, Jongwoo and Han, Bohyung},
  journal={Advances in Neural Information Processing Systems},
  year={2024}
}

@article{zhu2024motiongs,
  title={Motiongs: Exploring explicit motion guidance for deformable 3d gaussian splatting},
  author={Zhu, Ruijie and Liang, Yanzhe and Chang, Hanzhi and Deng, Jiacheng and Lu, Jiahao and Yang, Wenfei and Zhang, Tianzhu and Zhang, Yongdong},
  journal={Advances in Neural Information Processing Systems},
  year={2024}
}

@inproceedings{kwak2025modec,
  title={MoDec-GS: Global-to-Local Motion Decomposition and Temporal Interval Adjustment for Compact Dynamic 3D Gaussian Splatting},
  author={Kwak, Sangwoon and Kim, Joonsoo and Jeong, Jun Young and Cheong, Won-Sik and Oh, Jihyong and Kim, Munchurl},
  booktitle={Proceedings of the Computer Vision and Pattern Recognition Conference},
  year={2025}
}

@inproceedings{park2025splinegs,
  title={Splinegs: Robust motion-adaptive spline for real-time dynamic 3d gaussians from monocular video},
  author={Park, Jongmin and Bui, Minh-Quan Viet and Bello, Juan Luis Gonzalez and Moon, Jaeho and Oh, Jihyong and Kim, Munchurl},
  booktitle={Proceedings of the Computer Vision and Pattern Recognition Conference},
  pages={26866--26875},
  year={2025}
}

@inproceedings{lei2025mosca,
  title={Mosca: Dynamic gaussian fusion from casual videos via 4d motion scaffolds},
  author={Lei, Jiahui and Weng, Yijia and Harley, Adam W and Guibas, Leonidas and Daniilidis, Kostas},
  booktitle={Proceedings of the Computer Vision and Pattern Recognition Conference},
  pages={6165--6177},
  year={2025}
}

@inproceedings{fan2025spectromotion,
  title={Spectromotion: Dynamic 3d reconstruction of specular scenes},
  author={Fan, Cheng-De and Chang, Chen-Wei and Liu, Yi-Ruei and Lee, Jie-Ying and Huang, Jiun-Long and Tseng, Yu-Chee and Liu, Yu-Lun},
  booktitle={Proceedings of the Computer Vision and Pattern Recognition Conference},
  pages={21328--21338},
  year={2025}
}

@inproceedings{wang2023neural,
  title={Neural residual radiance fields for streamably free-viewpoint videos},
  author={Wang, Liao and Hu, Qiang and He, Qihan and Wang, Ziyu and Yu, Jingyi and Tuytelaars, Tinne and Xu, Lan and Wu, Minye},
  booktitle={Proceedings of the IEEE/CVF Conference on Computer Vision and Pattern Recognition},
  pages={76--87},
  year={2023}
}

@inproceedings{zheng2024hpc,
  title={HPC: Hierarchical Progressive Coding Framework for Volumetric Video},
  author={Zheng, Zihan and Zhong, Houqiang and Hu, Qiang and Zhang, Xiaoyun and Song, Li and Zhang, Ya and Wang, Yanfeng},
  booktitle={Proceedings of the 32nd ACM International Conference on Multimedia},
  pages={7937--7946},
  year={2024}
}

@inproceedings{zhang2024rate,
  title={Rate-aware Compression for NeRF-based Volumetric Video},
  author={Zhang, Zhiyu and Lu, Guo and Liang, Huanxiong and Cheng, Zhengxue and Tang, Anni and Song, Li},
  booktitle={Proceedings of the 32nd ACM International Conference on Multimedia},
  pages={3974--3983},
  year={2024}
}

@inproceedings{yin2024fsvfg,
  title={FSVFG: Towards Immersive Full-Scene Volumetric Video Streaming with Adaptive Feature Grid},
  author={Yin, Daheng and Shi, Jianxin and Zhang, Miao and Huang, Zhaowu and Liu, Jiangchuan and Dong, Fang},
  booktitle={Proceedings of the 32nd ACM International Conference on Multimedia},
  pages={11089--11098},
  year={2024}
}

@inproceedings{hu2025vrvvc,
  title={VRVVC: Variable-Rate NeRF-Based Volumetric Video Compression},
  author={Hu, Qiang and Zhong, Houqiang and Zheng, Zihan and Zhang, Xiaoyun and Cheng, Zhengxue and Song, Li and Zhai, Guangtao and Wang, Yanfeng},
  booktitle={Proceedings of the AAAI Conference on Artificial Intelligence},
  pages={3563--3571},
  year={2025}
}

@inproceedings{sun20243dgstream,
  title={3dgstream: On-the-fly training of 3d gaussians for efficient streaming of photo-realistic free-viewpoint videos},
  author={Sun, Jiakai and Jiao, Han and Li, Guangyuan and Zhang, Zhanjie and Zhao, Lei and Xing, Wei},
  booktitle={Proceedings of the IEEE/CVF Conference on Computer Vision and Pattern Recognition},
  pages={20675--20685},
  year={2024}
}

@article{gao2024hicom,
  title={HiCoM: Hierarchical Coherent Motion for Streamable Dynamic Scene with 3D Gaussian Splatting},
  author={Gao, Qiankun and Meng, Jiarui and Wen, Chengxiang and Chen, Jie and Zhang, Jian},
  journal={Advances in Neural Information Processing Systems},
  year={2024}
}

@article{girish2024queen,
  title={QUEEN: QUantized Efficient ENcoding of Dynamic Gaussians for Streaming Free-viewpoint Videos},
  author={Girish, Sharath and Li, Tianye and Mazumdar, Amrita and Shrivastava, Abhinav and Luebke, David and De Mello, Shalini},
  journal={Advances in Neural Information Processing Systems},
  year={2024}
}

@article{wang2024v,
  title={V\^{} 3: Viewing Volumetric Videos on Mobiles via Streamable 2D Dynamic Gaussians},
  author={Wang, Penghao and Zhang, Zhirui and Wang, Liao and Yao, Kaixin and Xie, Siyuan and Yu, Jingyi and Wu, Minye and Xu, Lan},
  journal={ACM Transactions on Graphics (TOG)},
  volume={43},
  number={6},
  pages={1--13},
  year={2024},
  publisher={ACM New York, NY, USA}
}

@inproceedings{wang2024videorf,
  title={VideoRF: Rendering Dynamic Radiance Fields as 2D Feature Video Streams},
  author={Wang, Liao and Yao, Kaixin and Guo, Chengcheng and Zhang, Zhirui and Hu, Qiang and Yu, Jingyi and Xu, Lan and Wu, Minye},
  booktitle={Proceedings of the IEEE/CVF Conference on Computer Vision and Pattern Recognition},
  pages={470--481},
  year={2024}
}

@article{hu20254dgc,
  title={4DGC: Rate-Aware 4D Gaussian Compression for Efficient Streamable Free-Viewpoint Video},
  author={Hu, Qiang and Zheng, Zihan and Zhong, Houqiang and Fu, Sihua and Song, Li and Zhai, Guangtao and Wang, Yanfeng and others},
  journal={Proceedings of the IEEE/CVF Conference on Computer Vision and Pattern Recognition},
  year={2025}
}

@inproceedings{li2025gifstream,
  title={GIFStream: 4D Gaussian-based Immersive Video with Feature Stream},
  author={Li, Hao and Li, Sicheng and Gao, Xiang and Batuer, Abudouaihati and Yu, Lu and Liao, Yiyi},
  booktitle={Proceedings of the Computer Vision and Pattern Recognition Conference},
  pages={21761--21770},
  year={2025}
}

@inproceedings{zwicker2001surface,
  title={Surface splatting},
  author={Zwicker, Matthias and Pfister, Hanspeter and Van Baar, Jeroen and Gross, Markus},
  booktitle={Proceedings of the 28th annual conference on Computer graphics and interactive techniques},
  pages={371--378},
  year={2001}
}

@inproceedings{morgenstern2025compact,
  title={Compact 3d scene representation via self-organizing gaussian grids},
  author={Morgenstern, Wieland and Barthel, Florian and Hilsmann, Anna and Eisert, Peter},
  booktitle={European Conference on Computer Vision},
  pages={18--34},
  year={2025},
  organization={Springer}
}

@article{jiang2024timeformer,
  title={TimeFormer: Capturing Temporal Relationships of Deformable 3D Gaussians for Robust Reconstruction},
  author={Jiang, DaDong and Ke, Zhihui and Zhou, Xiaobo and Hou, Zhi and Yang, Xianghui and Hu, Wenbo and Qiu, Tie and Guo, Chunchao},
  journal={arXiv preprint arXiv:2411.11941},
  year={2024}
}

@inproceedings{hou2022batchformer,
  title={Batchformer: Learning to explore sample relationships for robust representation learning},
  author={Hou, Zhi and Yu, Baosheng and Tao, Dacheng},
  booktitle={Proceedings of the IEEE/CVF Conference on Computer Vision and Pattern Recognition},
  pages={7256--7266},
  year={2022}
}

@article{kheradmand20243d,
  title={3D Gaussian Splatting as Markov Chain Monte Carlo},
  author={Kheradmand, Shakiba and Rebain, Daniel and Sharma, Gopal and Sun, Weiwei and Tseng, Jeff and Isack, Hossam and Kar, Abhishek and Tagliasacchi, Andrea and Yi, Kwang Moo},
  journal={arXiv preprint arXiv:2404.09591},
  year={2024}
}

@inproceedings{schoenberger2016sfm,
    author={Sch\"{o}nberger, Johannes Lutz and Frahm, Jan-Michael},
    title={Structure-from-Motion Revisited},
    booktitle={Conference on Computer Vision and Pattern Recognition (CVPR)},
    year={2016},
}

@inproceedings{kirillov2023segment,
  title={Segment anything},
  author={Kirillov, Alexander and Mintun, Eric and Ravi, Nikhila and Mao, Hanzi and Rolland, Chloe and Gustafson, Laura and Xiao, Tete and Whitehead, Spencer and Berg, Alexander C and Lo, Wan-Yen and others},
  booktitle={Proceedings of the IEEE/CVF International Conference on Computer Vision},
  pages={4015--4026},
  year={2023}
}

\clearpage

\section{Appendix}

\section{More Implementation Details}
\noindent \textbf{Compression}.
Gaussian attribute images are stored in lossless JPEG XL format, while temporal feature images are encoded  as a video using libx265 with the default quantization parameter $qp=20$. The channel of temporal features $\mathcal{E}$ is set to 16, allowing the reshaping of its dimensions from $W \times H \times 16$ to $4W \times 4H$ to ensure compatibility with existing video codecs. Rotation $R$ is clipped to the range $[-1, 2]$, opacity $O$ and color $C$ are clipped to the range $[-4, 4]$ and $[0, 4]$, respectively. These attributes, along with position $X$ and temporal features $\mathcal{E}$, are then normalized to the range $[0, 1]$. For smaller image size, rotation is quantized with $q_{r} = 2^{7}$, and scale and opacity are quantized with $q_{o} = q_{s} = 2^{6}$. Since position $X$ and scale $S$ are highly dependent on scene size and temporal features are content-related, no quantization is applied to these attributes.

\begin{figure}[!t]
   \centering
   \includegraphics[width=1.0\columnwidth]{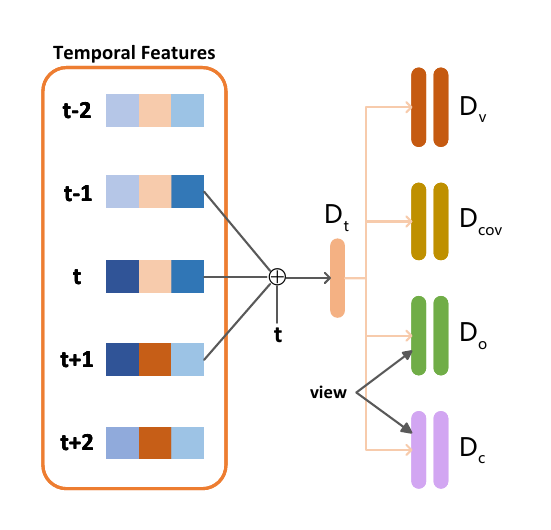}
   \caption{The deformation field of our StreamSTGS representation.}
   \label{fig:deform}
   \vspace{-0.2cm}
\end{figure}

\noindent \textbf{Deformation field.} 
The network architecture of our deformation field is presented as Fig.~\ref{fig:deform}. The temporal MLP comprises a single linear layer with 64 neurons, followed by a \textit{Tanh} activation function. The resulting output serves as the input for four Gaussian attribute prediction MLPs, i.e., $D_v$, $D_{cov}$, $D_o$, and $D_c$. Each of these MLPs consists of two linear layers, each containing 64 neurons. Bias are disabled for all MLPs. The learning rate for the canonical Gaussian attributes is same to that used in the original 3DGS. The temporal features are trained with a learning rate of $0.0025$. Furthermore, the learning rates for the deformation field networks are as follows: $D_t$ from $0.0025$ to $0.000025$, $D_v$ from $0.005$ to $0.00005$, $D_{cov}$ at $0.04$, $D_o$ from $0.002$ to $0.00002$, and $D_c$ from $0.008$ to $0.00005$.

\noindent \textbf{Auxiliary training module.} Our transformer module consists of two \textit{nn.TransformerEncoderLayer} and a single linear layer. Each \textit{nn.TransformerEncoderLayer} have two heads, hidden layers with 64 neurons, and \textit{Tanh} activation function. A \textit{Tanh} activation function is also applied to the output of the linear layer. The learning rate for the transformer is from $0.002$ to $0.00001$. To avoid the impact of timestamp length on positional encoding, we normalize it by the GOP length, e.g., timestamps within $[0, 59]$ or $[60, 119]$ are normalized into $[0,1]$.

\begin{table}[!t]
\resizebox{\columnwidth}{!}{%
    \begin{tabular}{ccccccc}
    \toprule
    Scene            & GOP1  & GOP2  & GOP3  & GOP4  & GOP5  & Average \\ \midrule
    \multicolumn{7}{c}{PSNR}                                           \\
    Coffee Martini   & 28.70 & 29.10 & 28.83 & 28.73 & 28.77 & 28.83   \\
    Cook Spinach     & 33.70 & 33.78 & 33.49 & 33.14 & 33.28 & 33.48   \\
    Cut Roasted Beef & 33.82 & 33.92 & 33.97 & 33.72 & 33.38 & 33.76   \\
    Flame Salmon     & 29.47 & 29.37 & 29.34 & 29.43 & 29.37 & 29.40   \\
    Flame Steak      & 34.14 & 33.95 & 33.97 & 33.69 & 33.55 & 33.86   \\
    Sear Steak       & 34.47 & 34.65 & 34.31 & 34.29 & 34.64 & 34.47   \\ \midrule
    \multicolumn{7}{c}{SSIM}                                           \\
    Coffee Martini   & 0.91  & 0.92  & 0.91  & 0.91  & 0.91  & 0.91    \\
    Cook Spinach     & 0.96  & 0.96  & 0.96  & 0.95  & 0.95  & 0.95    \\
    Cut Roasted Beef & 0.96  & 0.96  & 0.96  & 0.96  & 0.95  & 0.96    \\
    Flame Salmon     & 0.92  & 0.92  & 0.92  & 0.92  & 0.92  & 0.92    \\
    Flame Steak      & 0.96  & 0.96  & 0.96  & 0.96  & 0.96  & 0.96    \\
    Sear Steak       & 0.96  & 0.96  & 0.96  & 0.96  & 0.96  & 0.96    \\ \midrule
    \multicolumn{7}{c}{LPIPS}                                          \\
    Coffee Martini   & 0.160 & 0.160 & 0.162 & 0.162 & 0.160 & 0.161   \\
    Cook Spinach     & 0.145 & 0.146 & 0.148 & 0.153 & 0.149 & 0.148   \\
    Cut Roasted Beef & 0.146 & 0.143 & 0.148 & 0.146 & 0.154 & 0.148   \\
    Flame Salmon     & 0.160 & 0.155 & 0.157 & 0.156 & 0.153 & 0.156   \\
    Flame Steak      & 0.134 & 0.135 & 0.139 & 0.140 & 0.138 & 0.137   \\
    Sear Steak       & 0.134 & 0.133 & 0.133 & 0.135 & 0.134 & 0.134   \\ \bottomrule
    \end{tabular}%
}
\caption{Per-GOP Metrics for the N3DV dataset.}
\label{fig:per-gop}
\vspace{-0.2cm}
\end{table}

\begin{table*}[!t]
\resizebox{\textwidth}{!}{%
    \begin{tabular}{cccccccc}
    \toprule
    Scene            & PSNR $\uparrow$ & SSIM $\uparrow$ & LPIPS $\downarrow$ & \begin{tabular}[c]{@{}c@{}}Storage\\ (MB) $\downarrow$\end{tabular} & FPS $\uparrow$ & \begin{tabular}[c]{@{}c@{}}Training\\ (s) $\downarrow$\end{tabular} & \begin{tabular}[c]{@{}c@{}}Streamable/ \\ Variable bitrate \end{tabular} \\
    \midrule
    HyperReel~\cite{attal2023hyperreel}   & 31.10 & 0.928 & -     & 1.2   & 2.0  & 104                         & $\times$/$\times$          \\
    HexPlanes~\cite{cao2023hexplane}      & 31.70 & -     & -     & 0.8   & 0.21 & 144                         & $\times$/$\times$          \\
    KPlanes~\cite{fridovich2023k}         & 31.63 & -     & -     & 1.0   & 0.15 & 48                          & $\times$/$\times$          \\
    MixVoxels~\cite{wang2022mixed}        & 30.81 & -     & -     & 1.7   & 38   & 16                          & $\times$/$\times$          \\
    NeRFPlayer~\cite{song2023nerfplayer}  & 30.69 & 0.932 & 0.209 & 17.1  & 0.05 & 72                          & $\checkmark$/$\times$      \\
    StreamRF~\cite{li2022streaming}       & 30.68 & -     & -     & 31.4  & 8.3  & 15                          & $\checkmark$/$\times$      \\
    TeTriRF$^{*}$~\cite{wu2024tetrirf}    & 30.07 & 0.90  & 0.299 & 0.06  & 1.5  & 32                          & $\checkmark$/$\checkmark$      \\ \midrule
    4DGS~\cite{yang2023gs4d}              & 32.01 & -     & -     & 29    & 30   & 76                          & $\times$/$\times$          \\
    4DGaussians$^{*}$~\cite{wu20244d}     & 31.11 & 0.938 & 0.141 & 0.11  & 30   & 9                           & $\times$/$\times$          \\
    STGS~\cite{li2024spacetime}           & 31.62 & 0.946 & -     & 0.60  & 140  & 30                          & $\times$/$\times$          \\
    4D-Rotor~\cite{duan20244d}            & 31.62 & 0.94  & -     & -     & 277  & 12                          & $\times$/$\times$          \\
    DN-4DGS~\cite{lu2024dn}               & 32.02 & 0.944 & -     & 0.37  & 15   & 10                          & $\times$/$\times$          \\
    Ex4DGS~\cite{lee2024fully}            & 32.11 & 0.94  & -     & 0.38  & -    & -                           & $\times$/$\times$          \\
    3DGStream$^{*}$~\cite{sun20243dgstream} & 30.73 & 0.935 & 0.147 & 8.0   & 72   & 17                        & $\checkmark$/$\times$      \\
    VideoGS$^{*}$~\cite{wang2024v}        & 27.45 & 0.871 & 0.214 & 0.91  & 21   & 143                         & $\checkmark$/$\checkmark$      \\
    HiCoM$^{*}$~\cite{gao2024hicom}       & 31.32 & 0.939 & 0.147 & 10.5  & 163  & 10                          & $\checkmark$/$\times$      \\
    QUEEN~\cite{girish2024queen}          & 32.19 & 0.946 & -     & 0.75  & -    & -                           & $\checkmark$/$\times$      \\
    4DGC$^{*}$~\cite{hu20254dgc}          & 31.52 & 0.941 & 0.143 & 0.77  & 79   & 62                          & $\checkmark$/$\times$       \\
    Ours        & 32.30 & 0.943 & 0.147 & 0.17  & 100  & 67                                                    & $\checkmark$/$\checkmark$      \\
    \bottomrule
    \end{tabular}%
}
\caption{Quantitative Comparisons on the N3DV Datasets. $*$ refers to our re-implementation with our dataset preparation and sparse point cloud generation based on their official open-sourced codes.}
\label{tab:all}
\vspace{-0.2cm}
\end{table*}

\section{Additional Quantitative Results}
Table~\ref{tab:all} presents a comprehensive comparisons against additional dynamic scene reconstruction baselines on the N3DV dataset, representing a superset of the results reported in Table~1 of the main paper. Methods marked with $*$ in Table~\ref{tab:all} denote our re-implementation, using our dataset preparation and sparse point cloud generation based on their official public codes. Owing to a bit performance variations stem from differences in dataset preparation and sparse cloud point generation methods. For example, image undistortion, adding depth information from pre-trained depth prediction model, or merging of per-frame point clouds into a unified initial point cloud, instead of relying solely on the first frame's point cloud, can lead to performance improvements.

\section{Per-scene Quantitative Results}
In addition to the average quantitative results over the full datasets of N3DV and MeetRoom in Table~\ref{tab:all}, we show results for each scene in both the N3DV and MeetRoom datasets of various metrics, including PSNR, SSIM, LPIPS(VGG), storage(frame size), FPS, decoding time, training time in Table~\ref{tab:n3dvp} and Table~\ref{tab:meetp}.

\begin{table*}[!t]
\resizebox{\textwidth}{!}{%
    \begin{tabular}{cccccccc}
    \toprule
    Scene            & PSNR $\uparrow$ & SSIM $\uparrow$ & LPIPS $\downarrow$ & \begin{tabular}[c]{@{}c@{}}Storage\\ (KB) $\downarrow$\end{tabular} & FPS $\uparrow$ & \begin{tabular}[c]{@{}c@{}}Decoding\\ (ms) $\downarrow$\end{tabular} & \begin{tabular}[c]{@{}c@{}}Training\\ (s) $\downarrow$\end{tabular} \\
    \midrule
    Coffee Martini   & 28.83 & 0.91 & 0.161 & 164        & 105 & 8.7        & 69                                                     \\
    Cook Spinach     & 33.48 & 0.95 & 0.148 & 183        & 98  & 8.7        & 64                                                     \\
    Cut Roasted Beef & 33.76 & 0.96 & 0.148 & 188        & 102 & 9.0        & 65                                                     \\
    Flame Salmon     & 29.4  & 0.92 & 0.156 & 148        & 98  & 8.0        & 68                                                     \\
    Flame Steak      & 33.86 & 0.96 & 0.137 & 179        & 97  & 8.6        & 68                                                     \\
    Sear Steak       & 34.47 & 0.96 & 0.134 & 180        & 99  & 8.9        & 66                                                     \\
    \bottomrule
    \end{tabular}%
    }
\caption{Per-scene metrics for the N3DV dataset.}
\label{tab:n3dvp}
\vspace{-0.2cm}
\end{table*}

\begin{table*}[!t]
\resizebox{\textwidth}{!}{%
    \begin{tabular}{cccccccc}
    \toprule
    Scene            & PSNR $\uparrow$ & SSIM $\uparrow$ & LPIPS $\downarrow$ & \begin{tabular}[c]{@{}c@{}}Storage\\ (KB) $\downarrow$\end{tabular} & FPS $\uparrow$ & \begin{tabular}[c]{@{}c@{}}Decoding\\ (ms) $\downarrow$\end{tabular} & \begin{tabular}[c]{@{}c@{}}Training\\ (s) $\downarrow$\end{tabular} \\
    \midrule
    Discussion & 27.74 & 0.91 & 0.211 & 161                                                    & 125 & 7.2                                                     & 29                                                     \\
    Trimming   & 27.65 & 0.92 & 0.213 & 122                                                    & 139 & 5.8                                                     & 29                                                     \\
    Vrheadset  & 26.83 & 0.91 & 0.223 & 144                                                    & 114 & 6.7                                                     & 30                                                     \\
    \bottomrule
    \end{tabular}%
    }
\caption{Per-scene metrics for the MeetRoom dataset.}
\label{tab:meetp}
\vspace{-0.2cm}
\end{table*}

\begin{figure*}[!t]
   \centering
   \subfloat[TeTriRF~\cite{wu2024tetrirf}]{
      \includegraphics[width=0.24\textwidth]{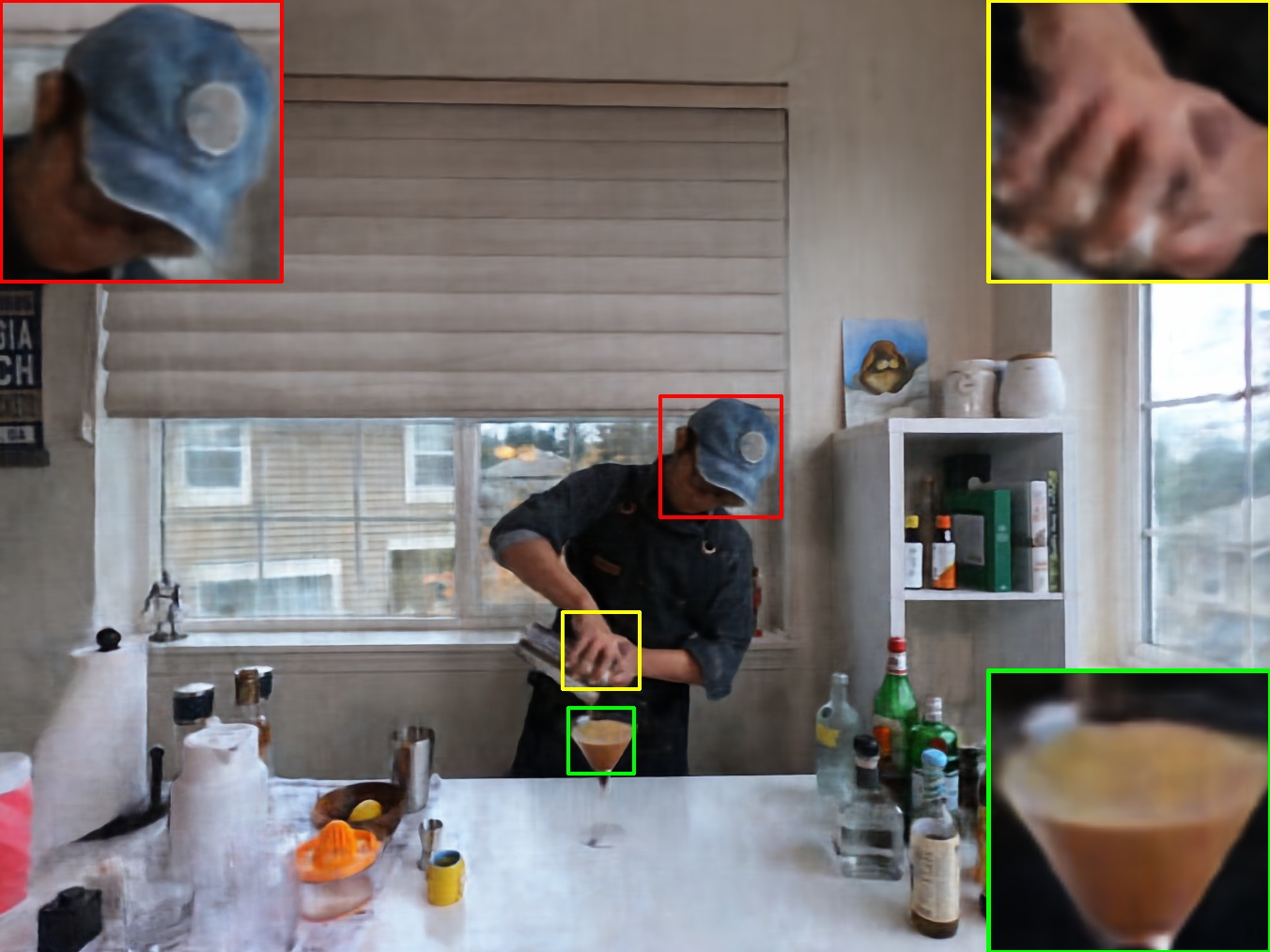}
   }
   \subfloat[3DGStream~\cite{sun20243dgstream}]{
      \includegraphics[width=0.24\textwidth]{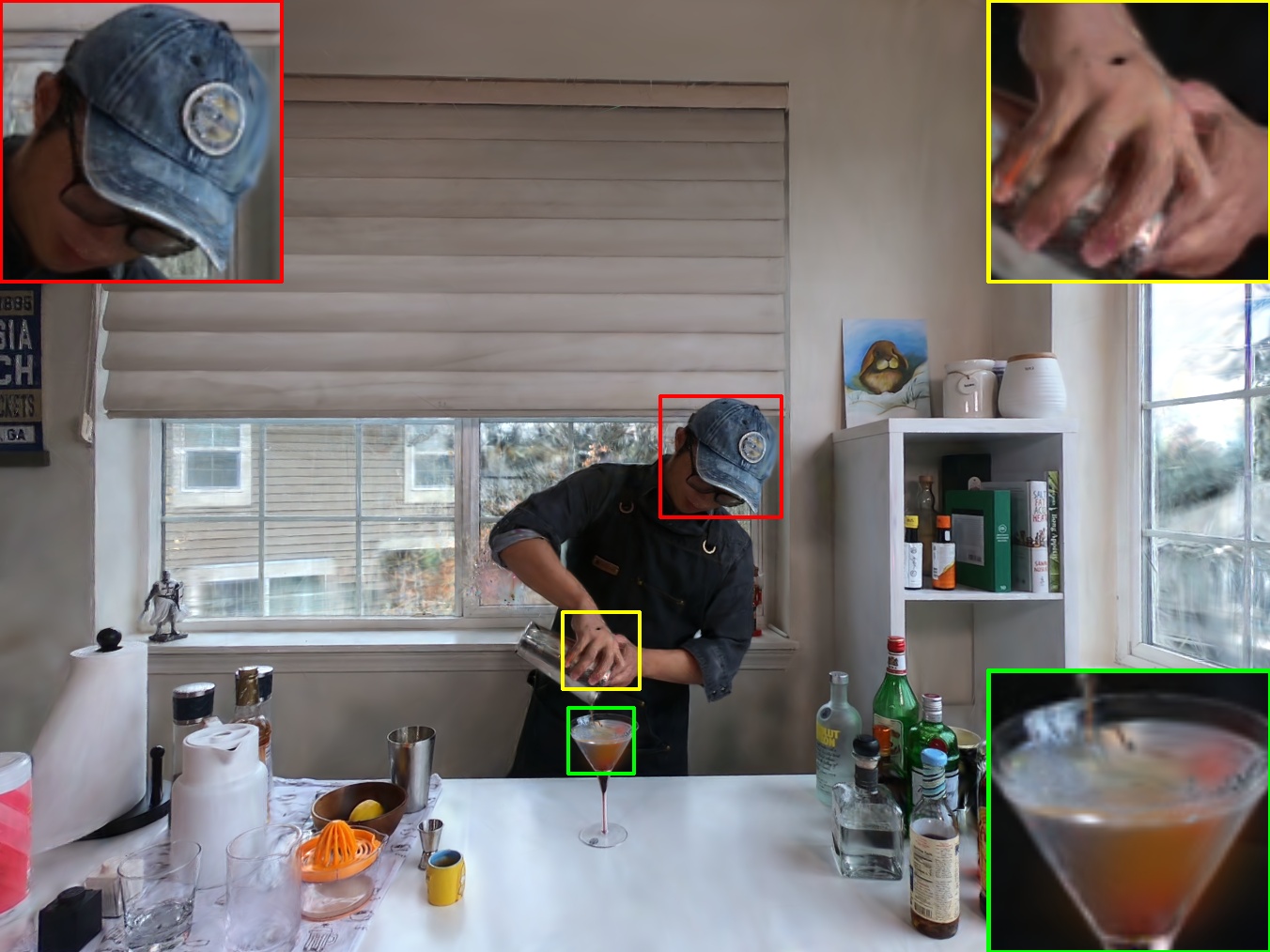}
   }
   \subfloat[HiCoM~\cite{gao2024hicom}]{
      \includegraphics[width=0.24\textwidth]{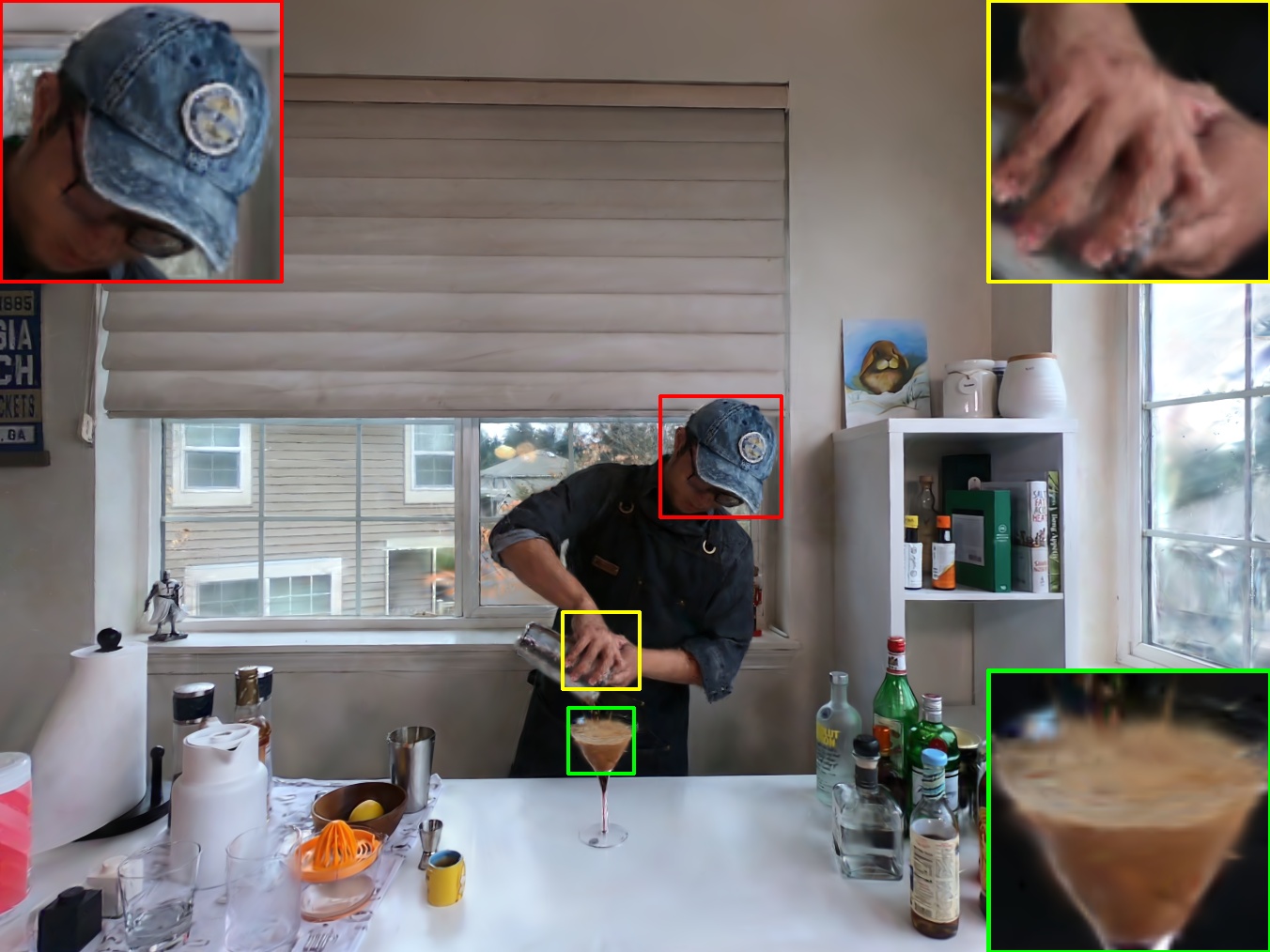}
   }
   \subfloat[VideoGS~\cite{wang2024v}]{
      \includegraphics[width=0.24\textwidth]{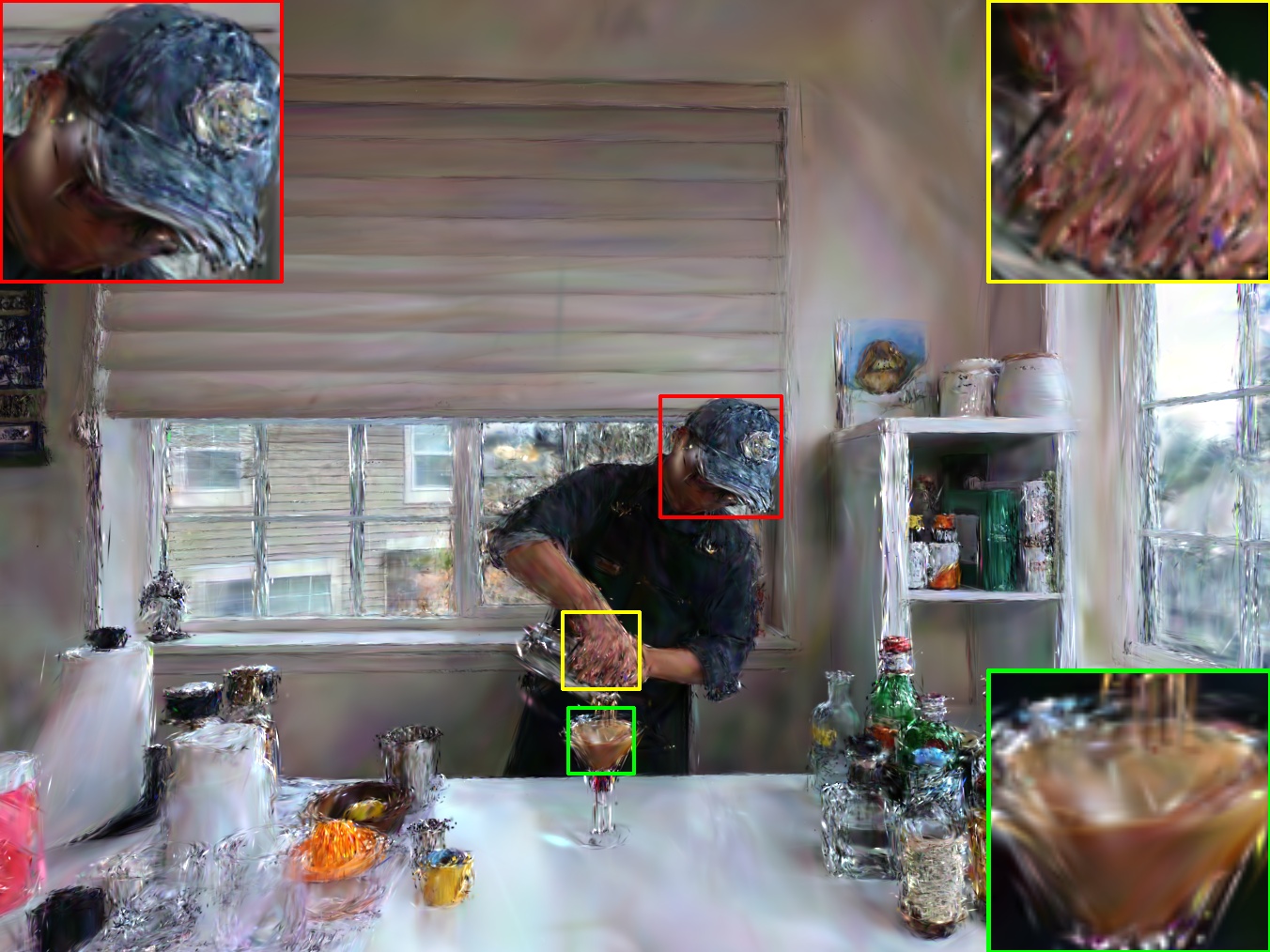}
   }
   \ 
   \subfloat[4DGC~\cite{hu20254dgc}]{
      \includegraphics[width=0.32\textwidth]{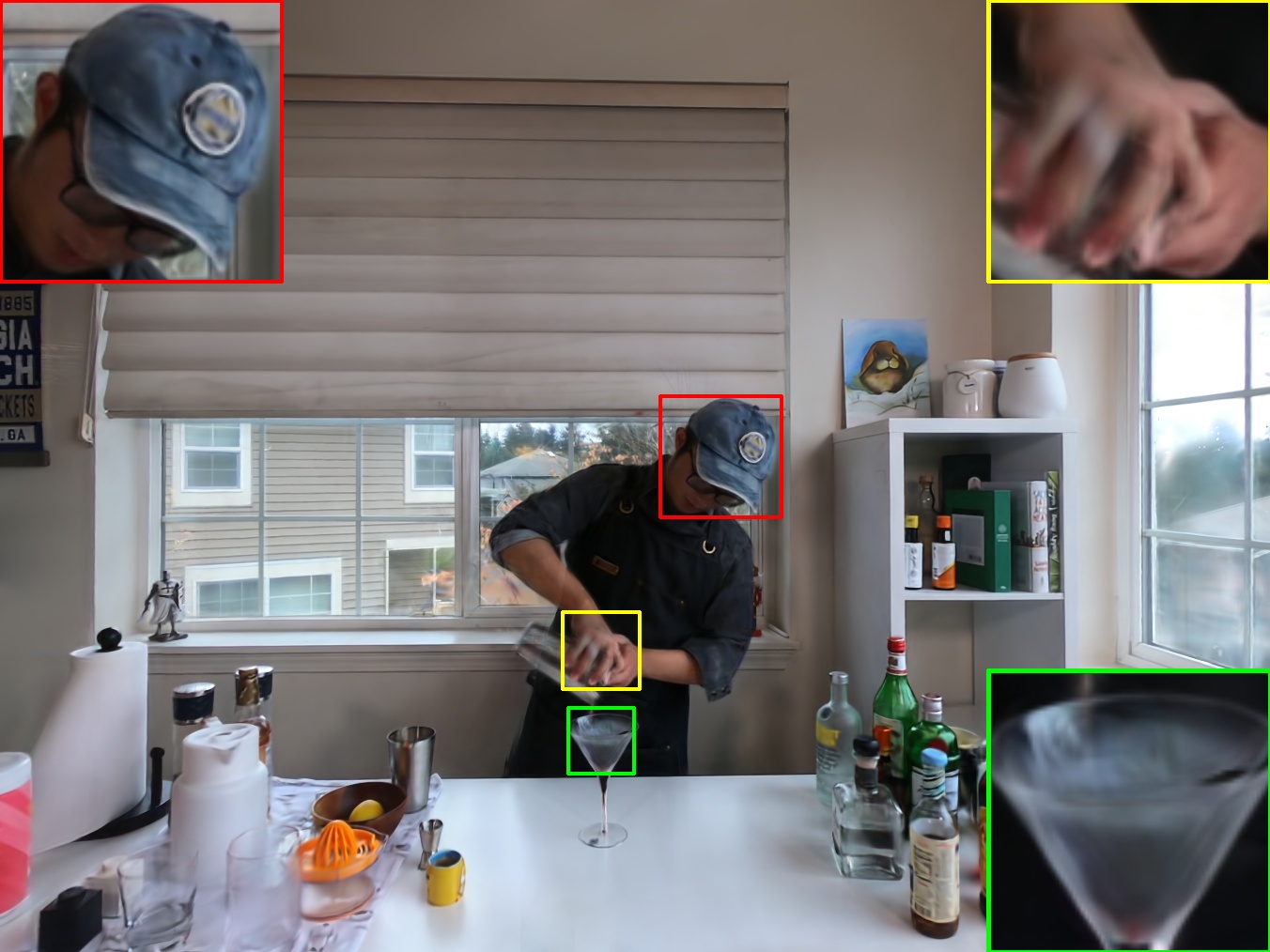}
   }
   \subfloat[Ours]{
      \includegraphics[width=0.32\textwidth]{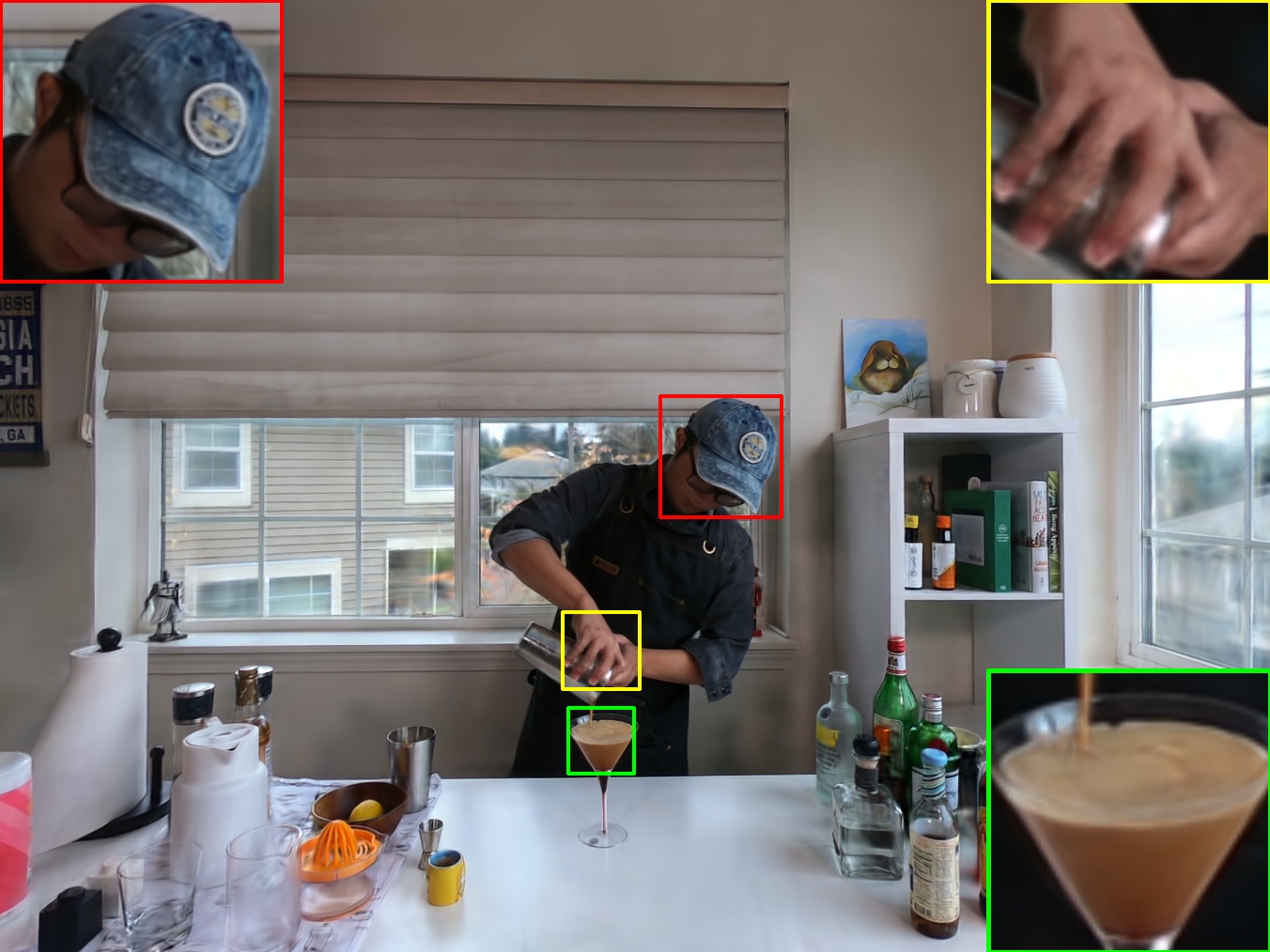}
   }
   \subfloat[GroudTruth]{
      \includegraphics[width=0.32\textwidth]{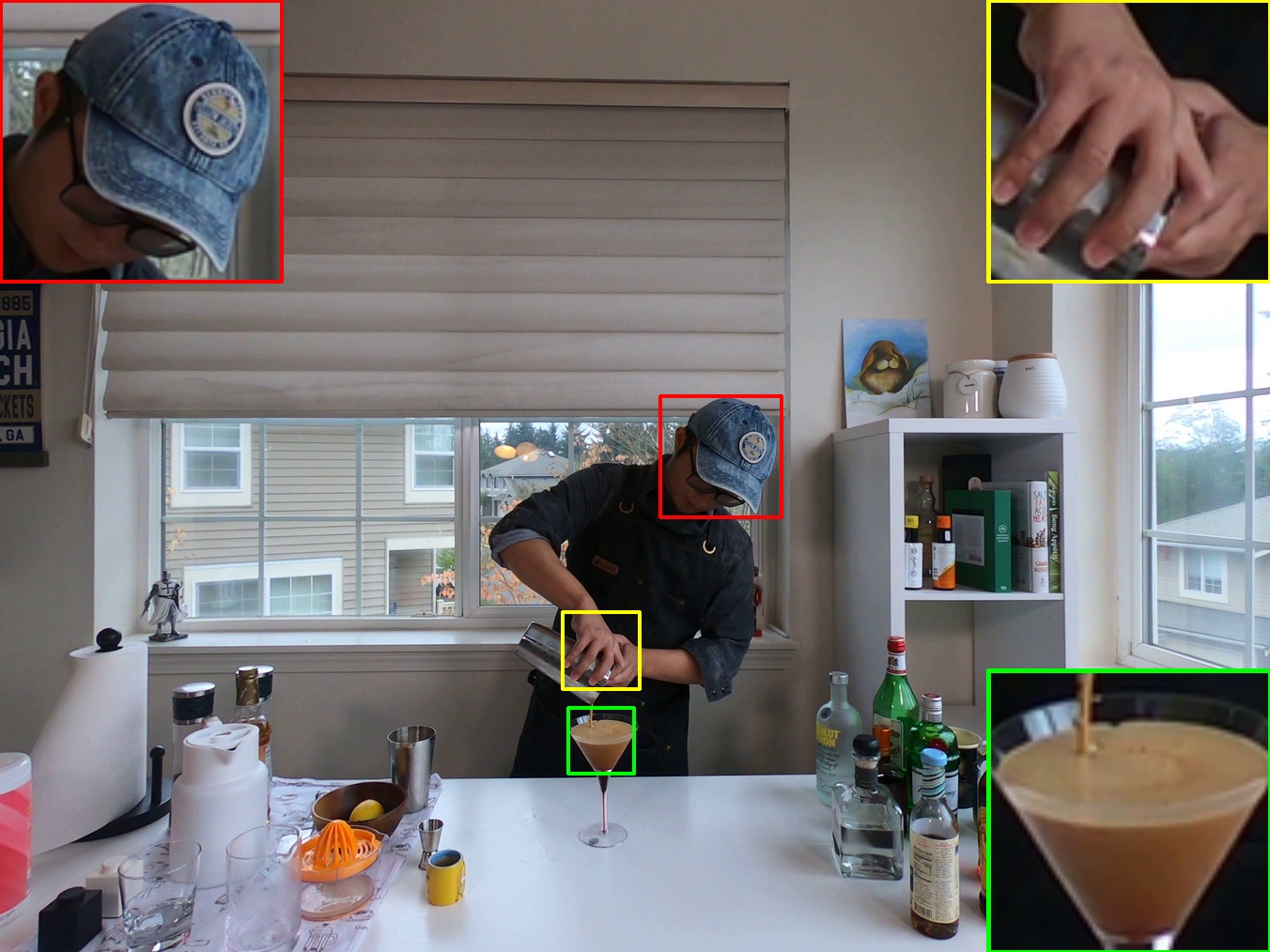}
   }   
   \caption{Qualitative comparison of ours with the benchmark methods on \textit{Coffee Martini} scene of N3DV dataset.}
   \label{fig:coffee}
   \vspace{-0.2cm}
\end{figure*}

\section{Per-scene Qualitative Results}

\begin{figure*}[!t]
   \centering
   \subfloat[TeTriRF~\cite{wu2024tetrirf}]{
      \includegraphics[width=0.24\textwidth]{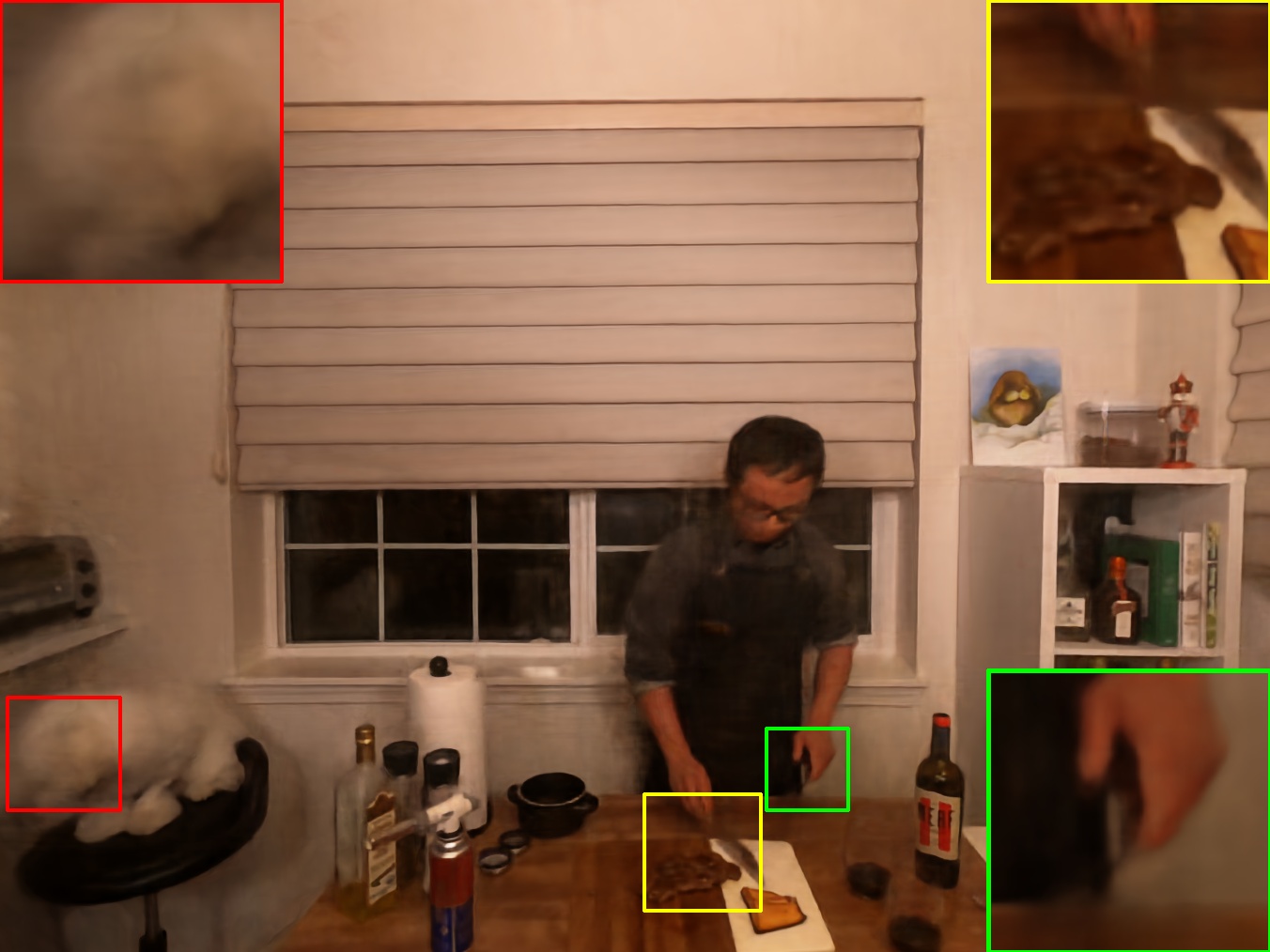}
   }
   \subfloat[3DGStream~\cite{sun20243dgstream}]{
      \includegraphics[width=0.24\textwidth]{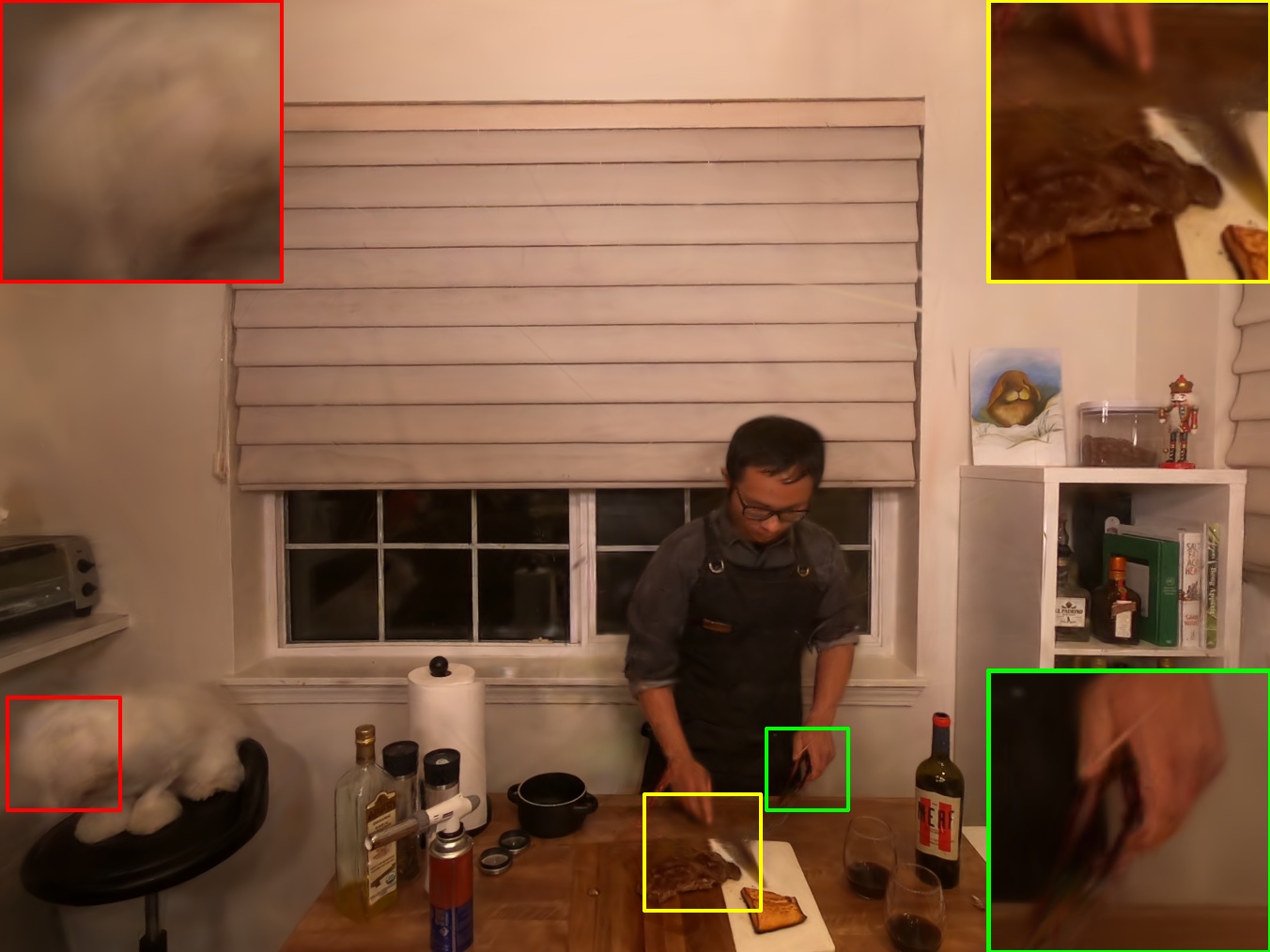}
   }
   \subfloat[HiCoM~\cite{gao2024hicom}]{
      \includegraphics[width=0.24\textwidth]{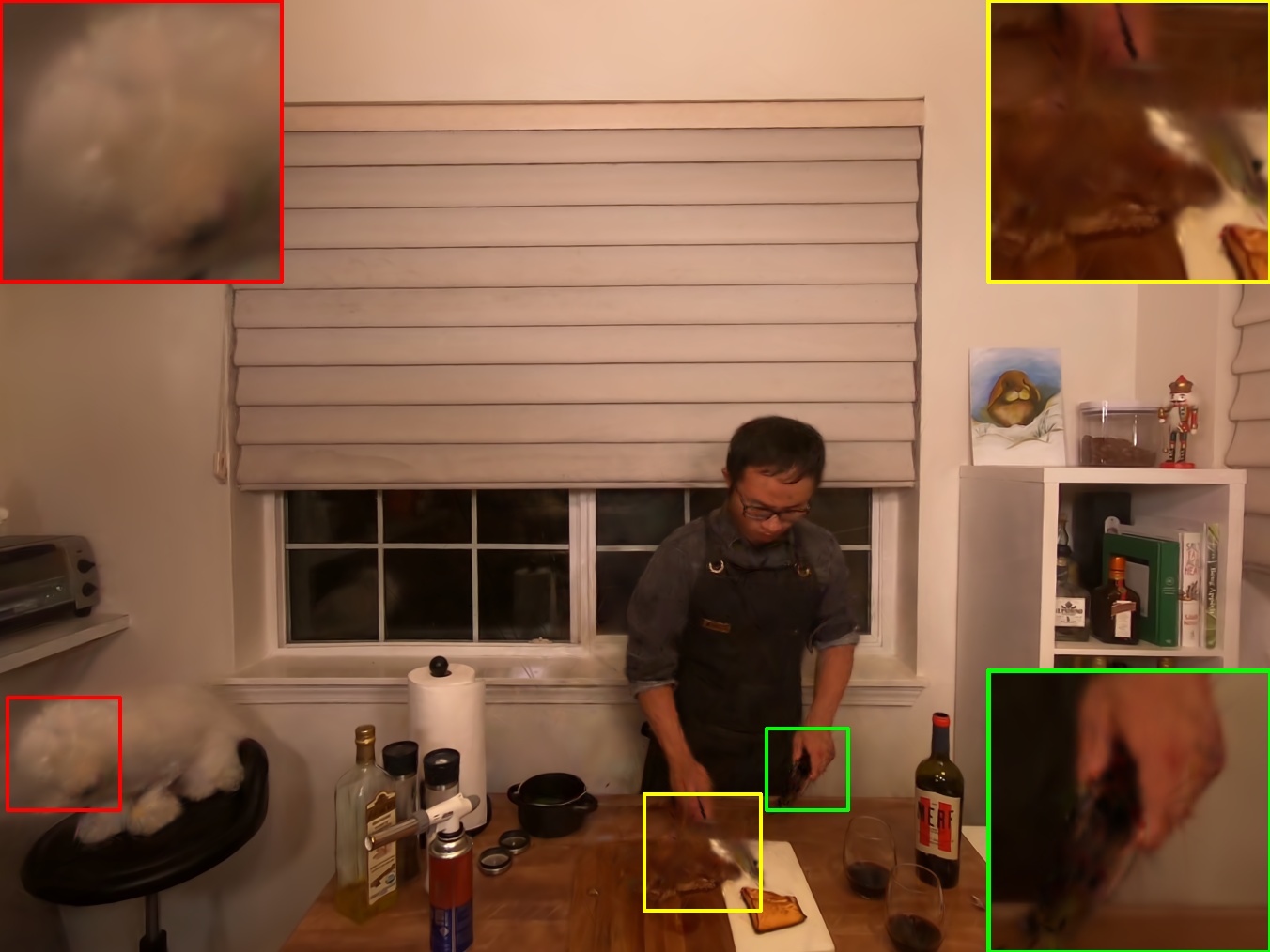}
   }
   \subfloat[VideoGS~\cite{wang2024v}]{
      \includegraphics[width=0.24\textwidth]{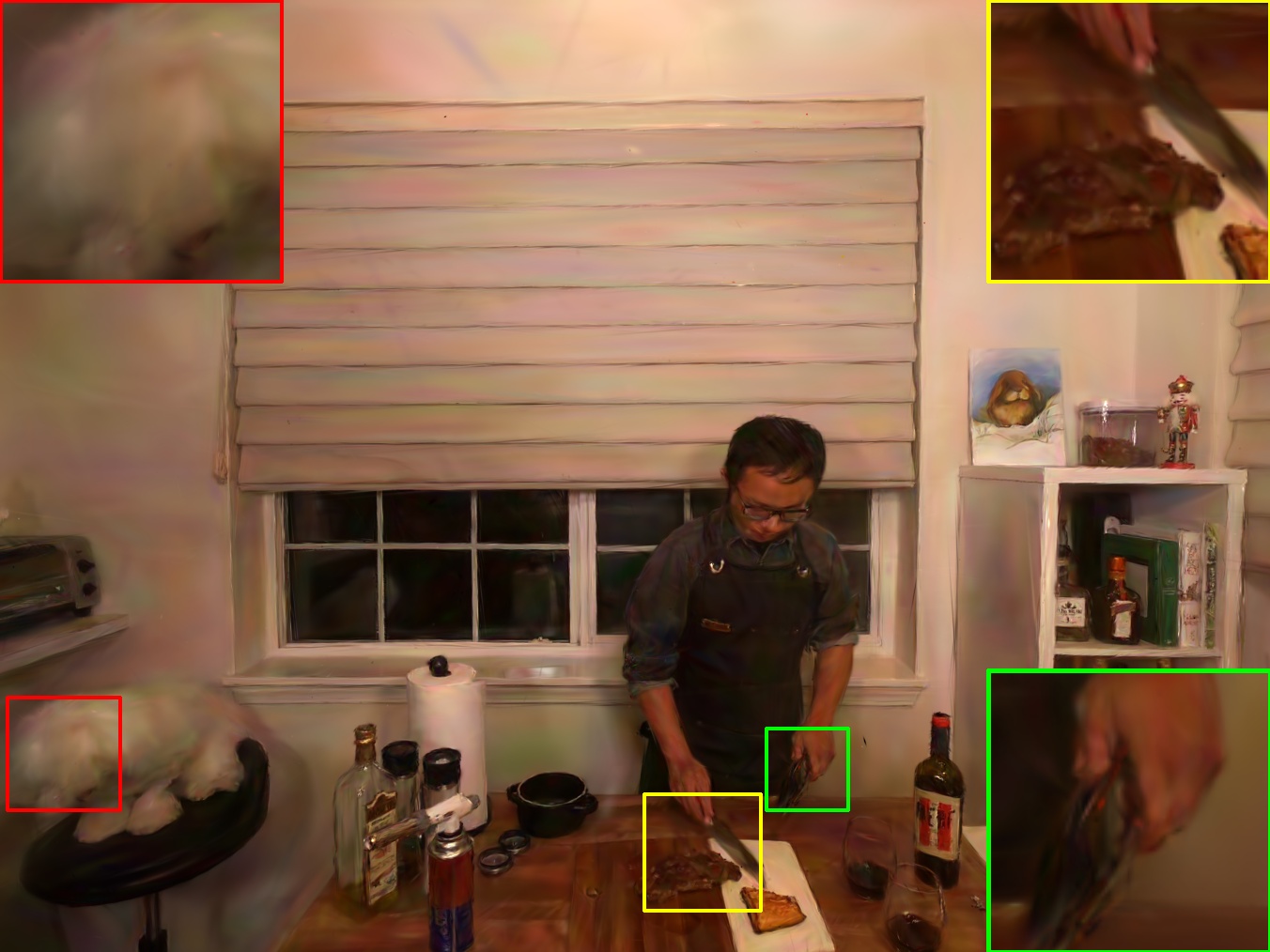}
   }
   \ 
   \subfloat[4DGC~\cite{hu20254dgc}]{
      \includegraphics[width=0.32\textwidth]{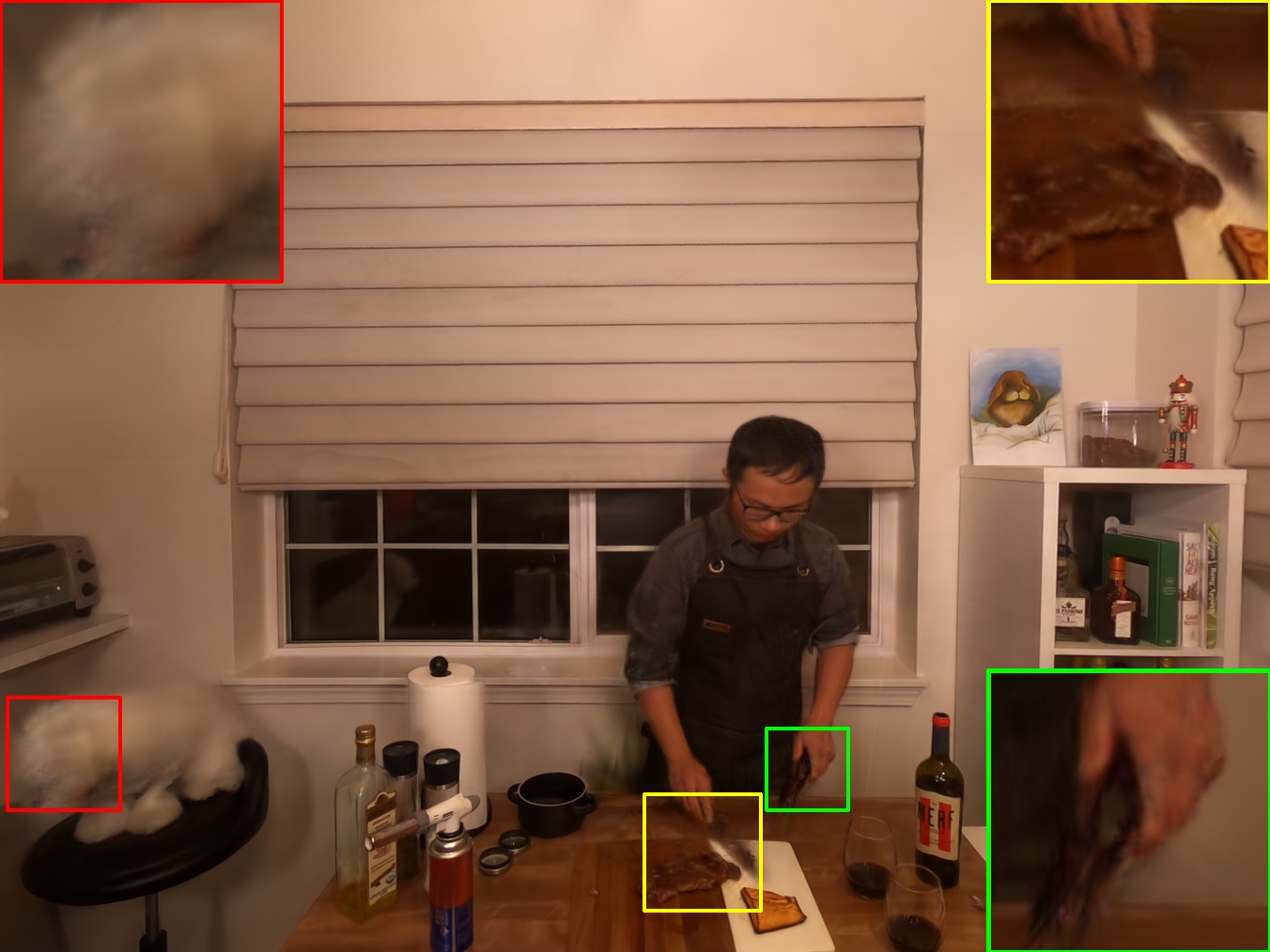}
   }
   \subfloat[Ours]{
      \includegraphics[width=0.32\textwidth]{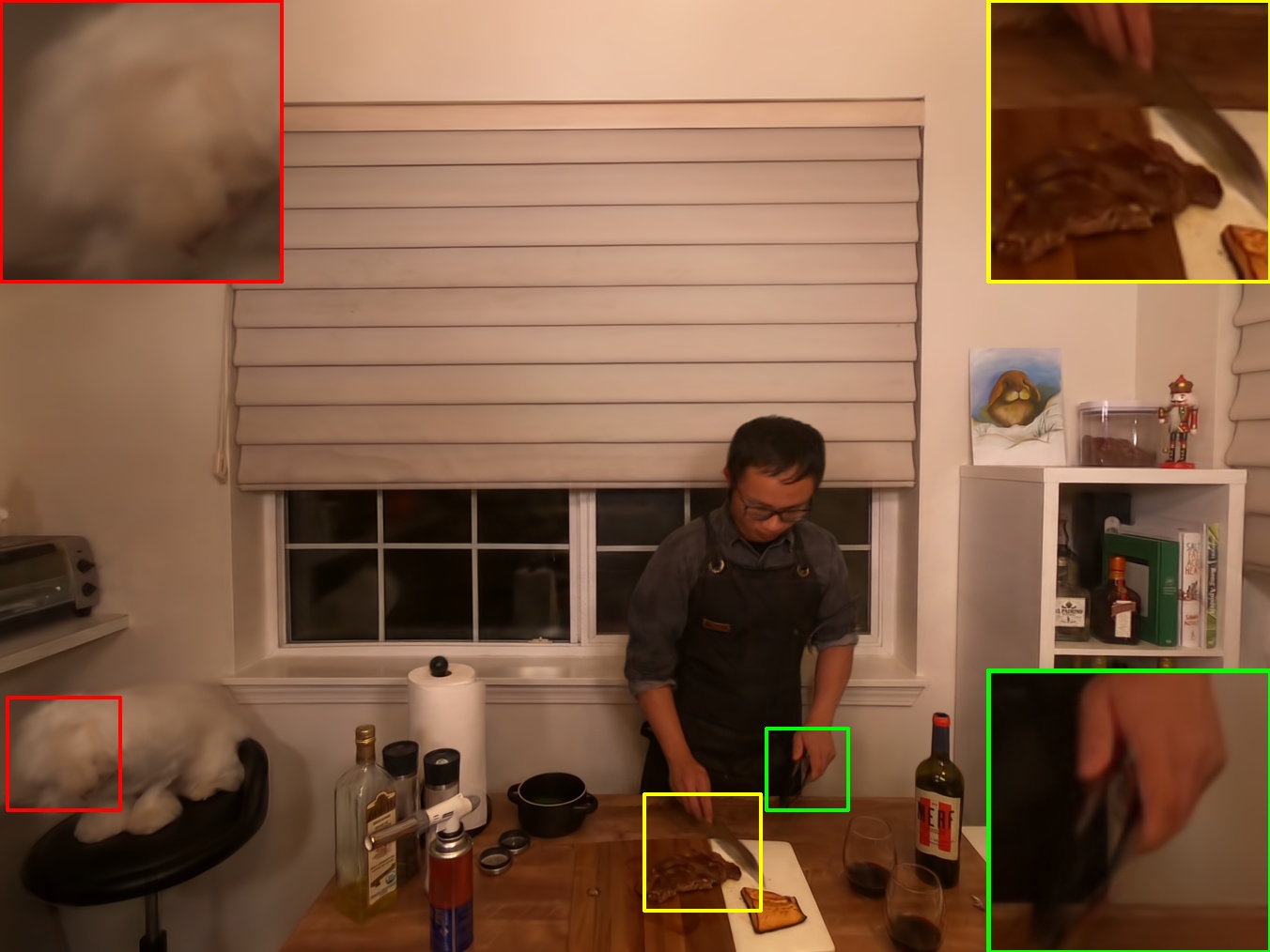}
   }
   \subfloat[GroudTruth]{
      \includegraphics[width=0.32\textwidth]{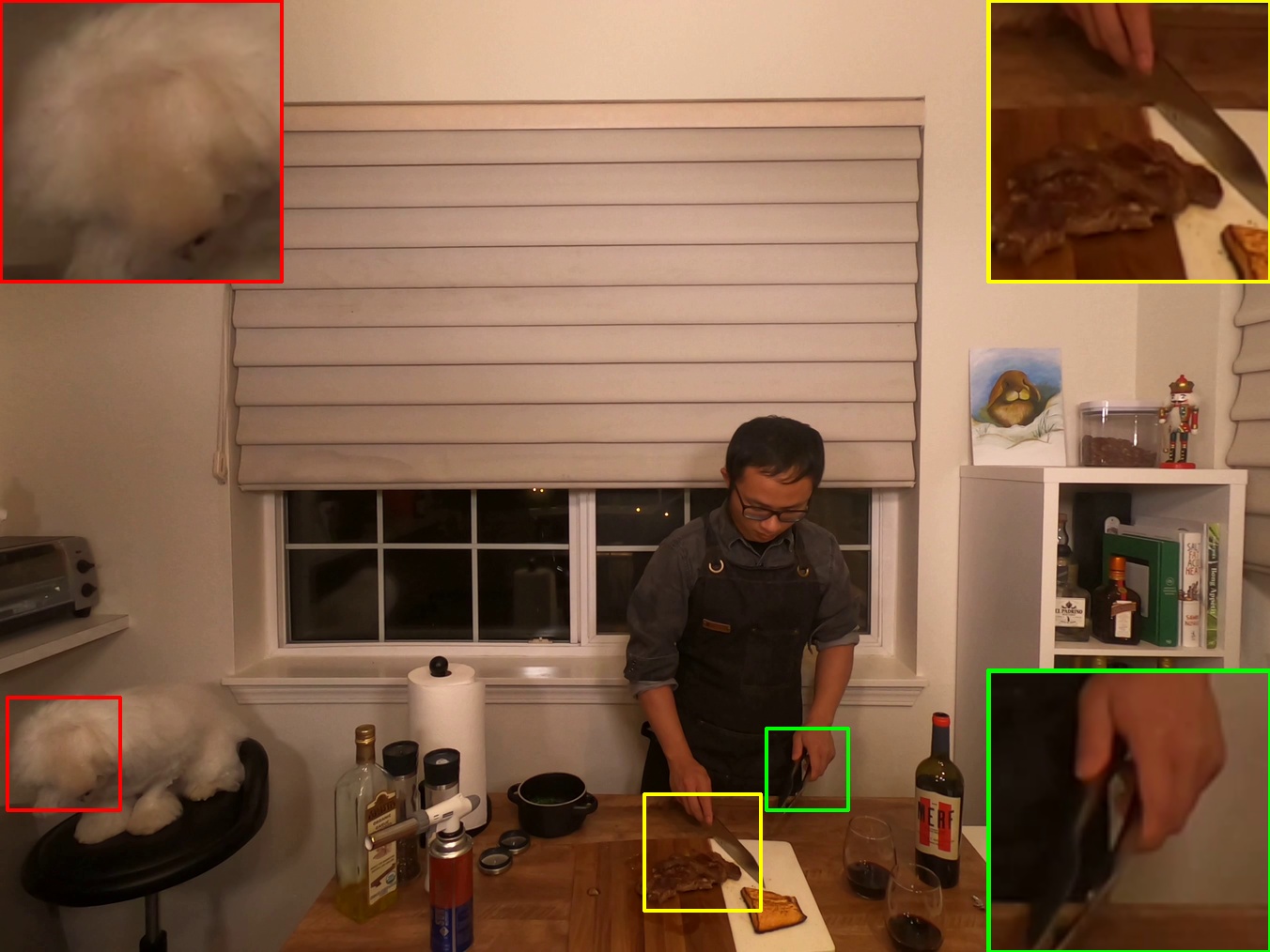}
   }  
   \caption{Qualitative comparison of ours with the benchmark methods on \textit{Cut Roasted Beef} scene of N3DV dataset.}
   \label{fig:cut}
   \vspace{-0.2cm}
\end{figure*}

\begin{figure*}[!t]
   \centering
   \subfloat[TeTriRF~\cite{wu2024tetrirf}]{
      \includegraphics[width=0.24\textwidth]{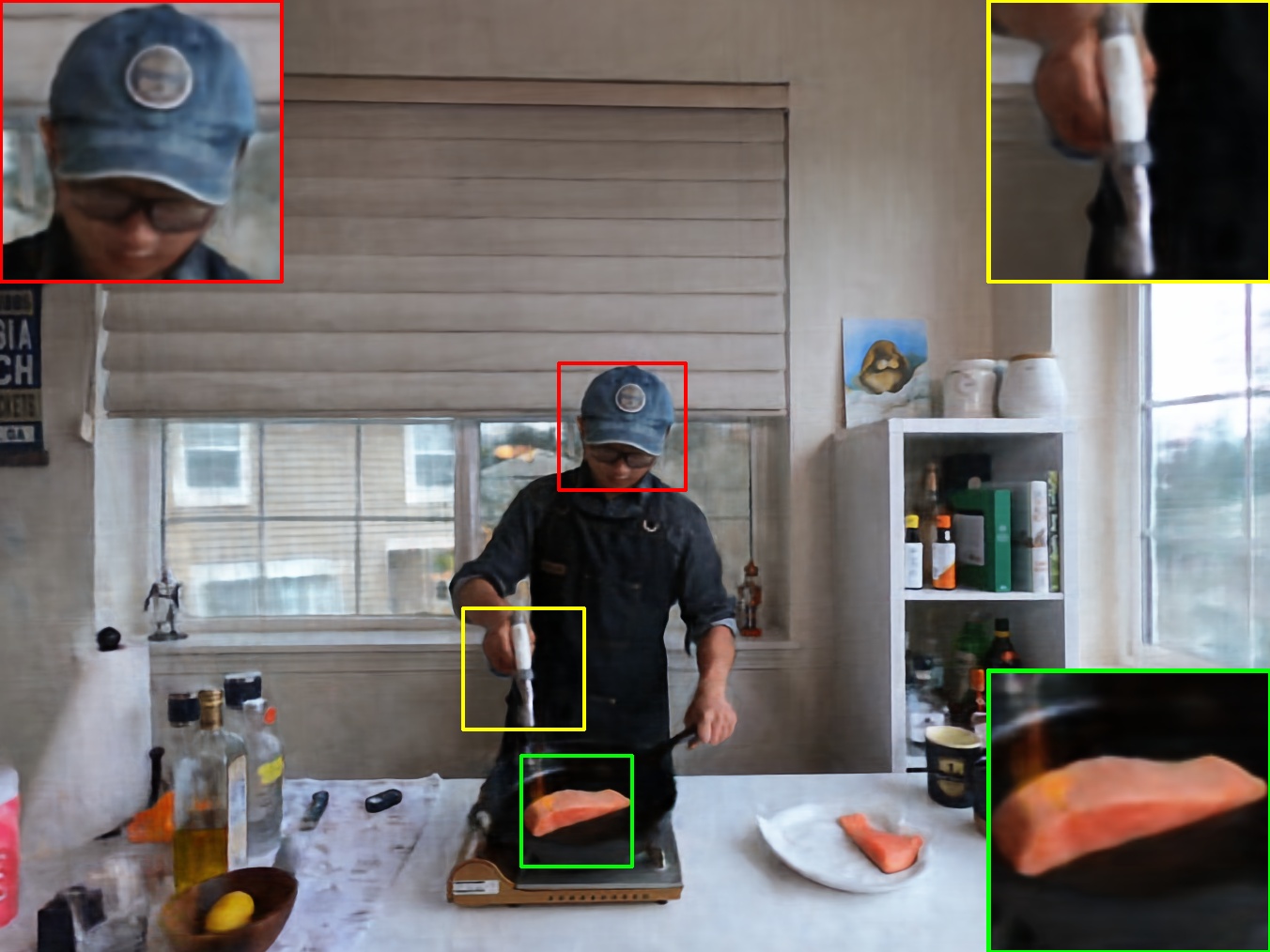}
   }
   \subfloat[3DGStream~\cite{sun20243dgstream}]{
      \includegraphics[width=0.24\textwidth]{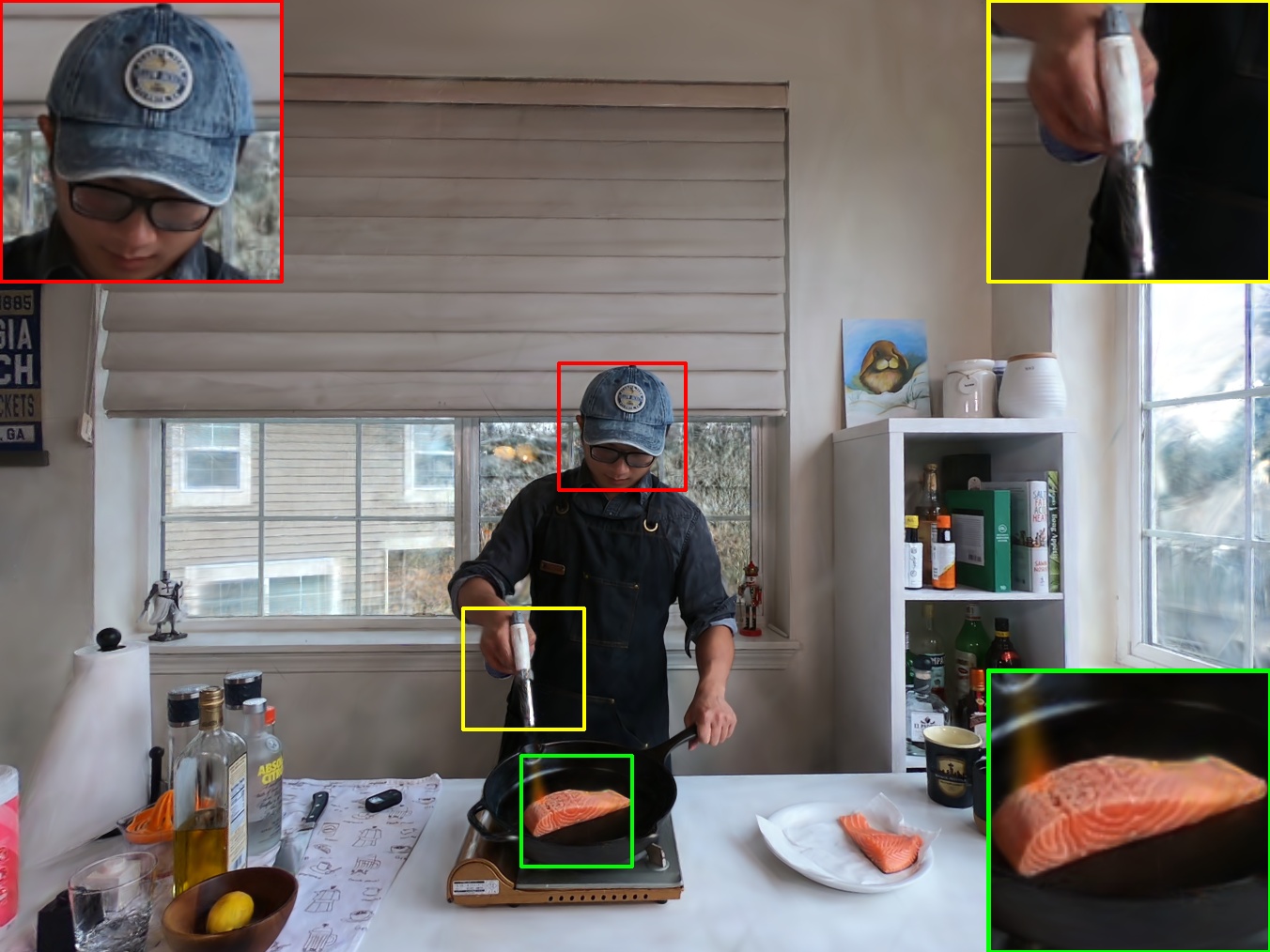}
   }
   \subfloat[HiCoM~\cite{gao2024hicom}]{
      \includegraphics[width=0.24\textwidth]{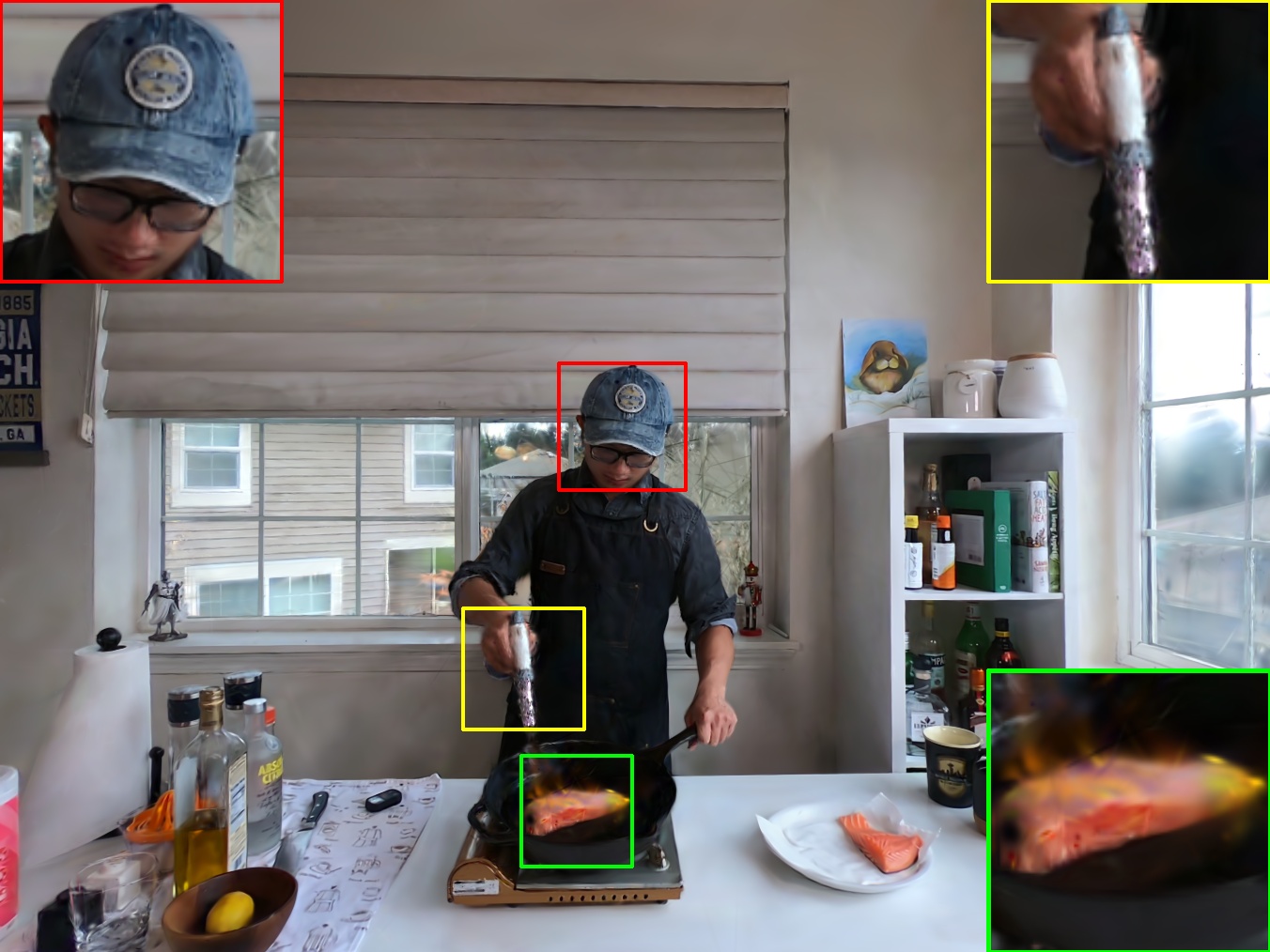}
   }
   \subfloat[VideoGS~\cite{wang2024v}]{
      \includegraphics[width=0.24\textwidth]{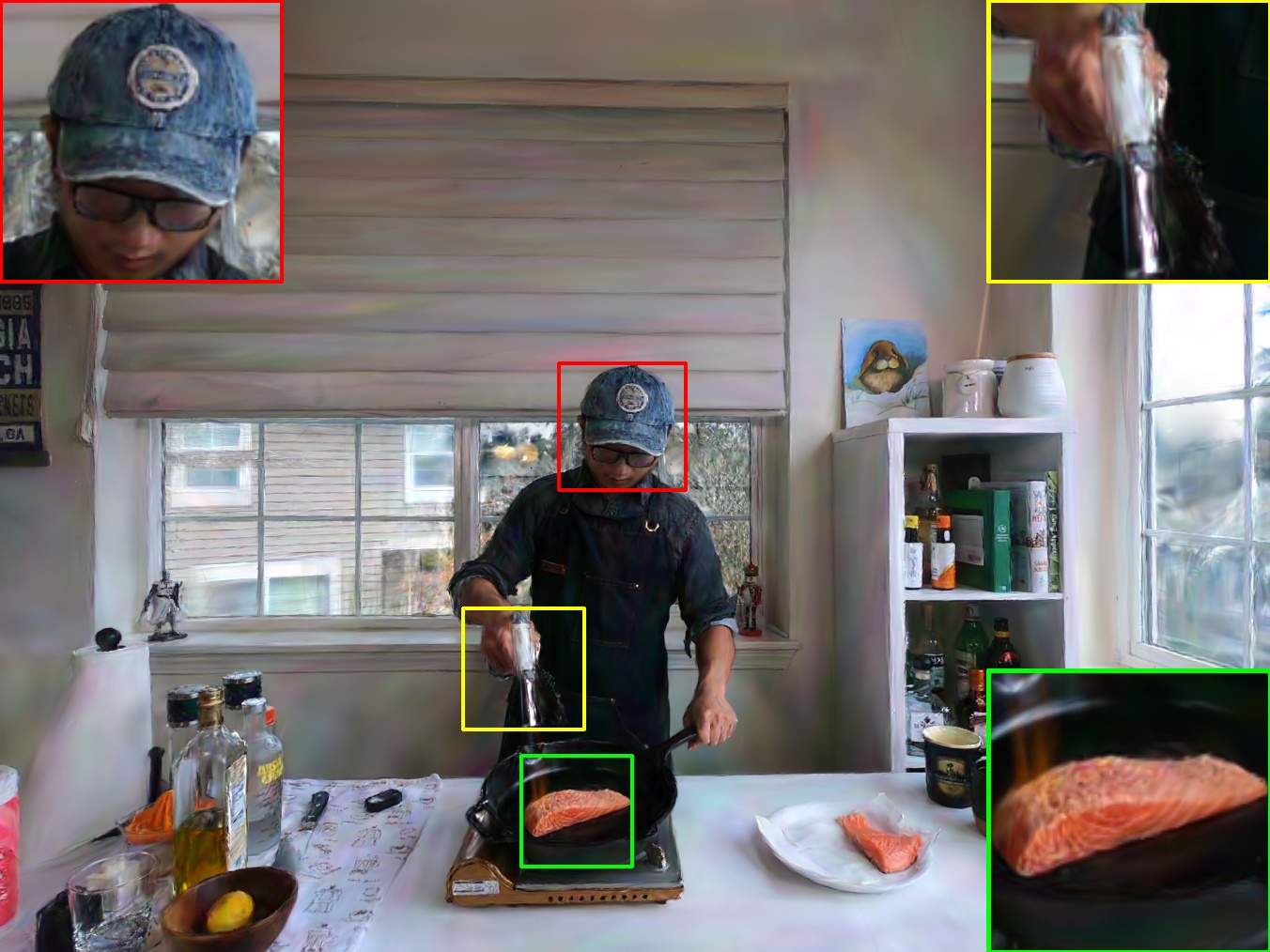}
   }
   \ 
   \subfloat[4DGC~\cite{hu20254dgc}]{
      \includegraphics[width=0.32\textwidth]{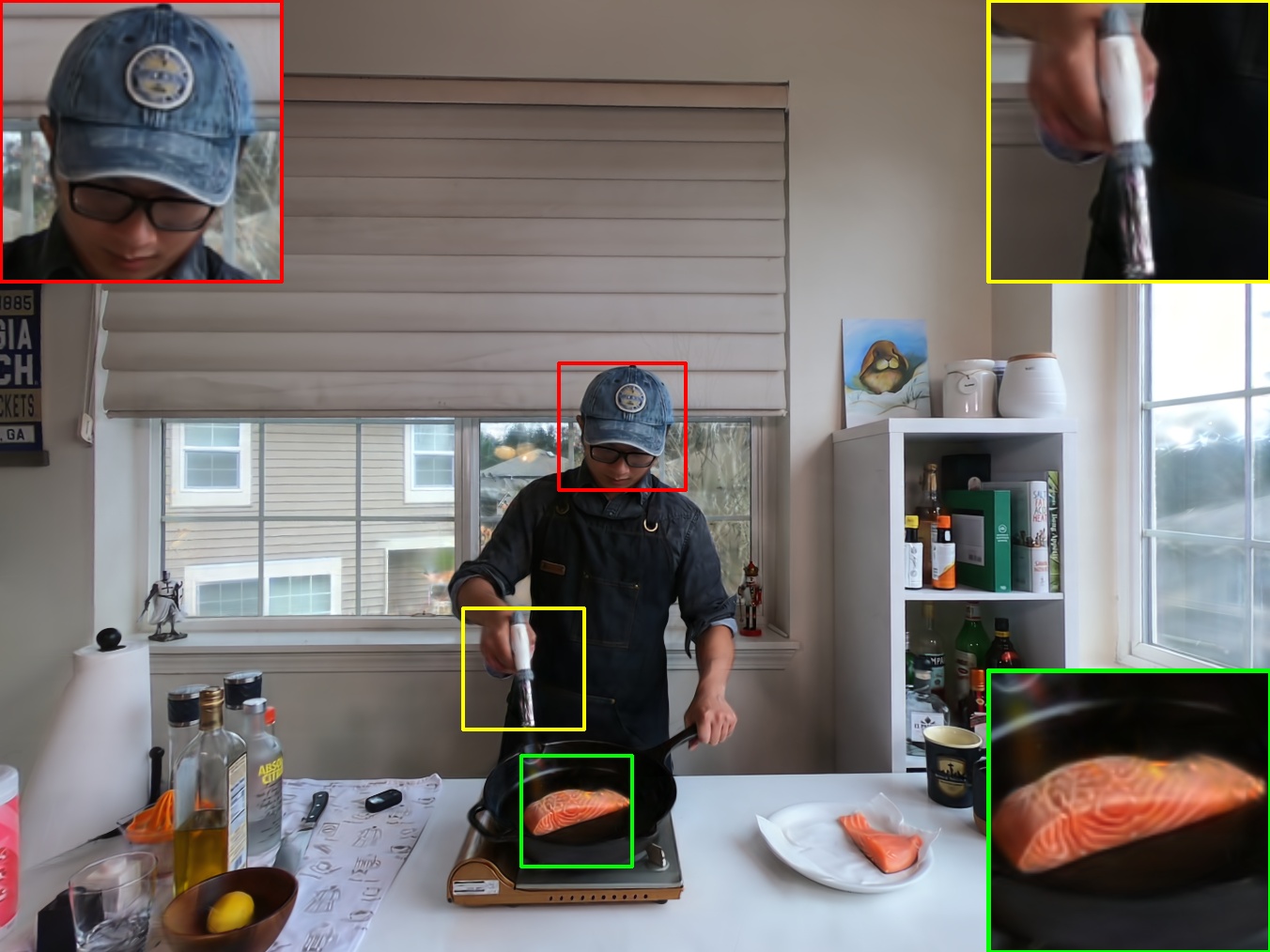}
   }
   \subfloat[Ours]{
      \includegraphics[width=0.32\textwidth]{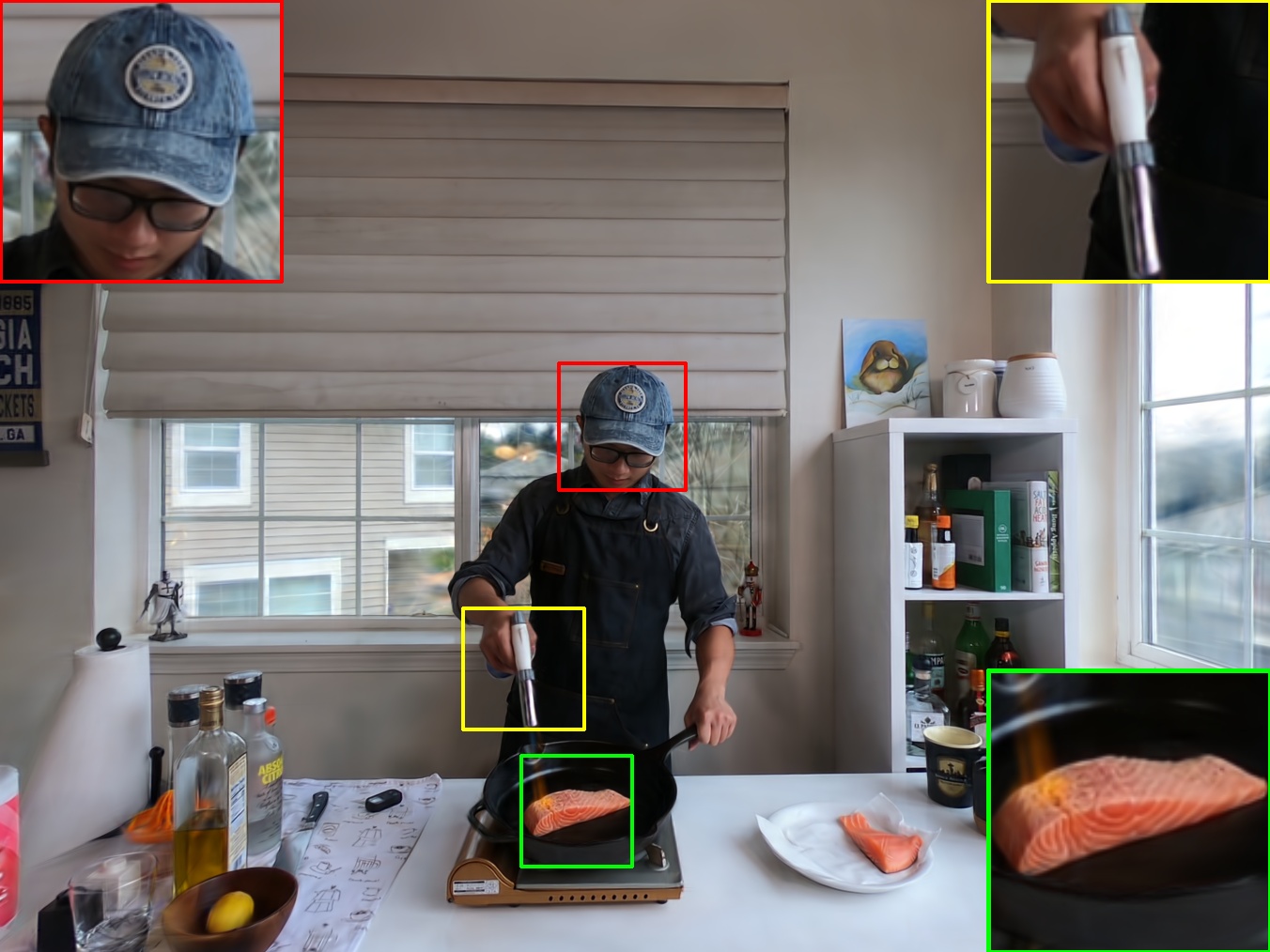}
   }
   \subfloat[GroudTruth]{
      \includegraphics[width=0.32\textwidth]{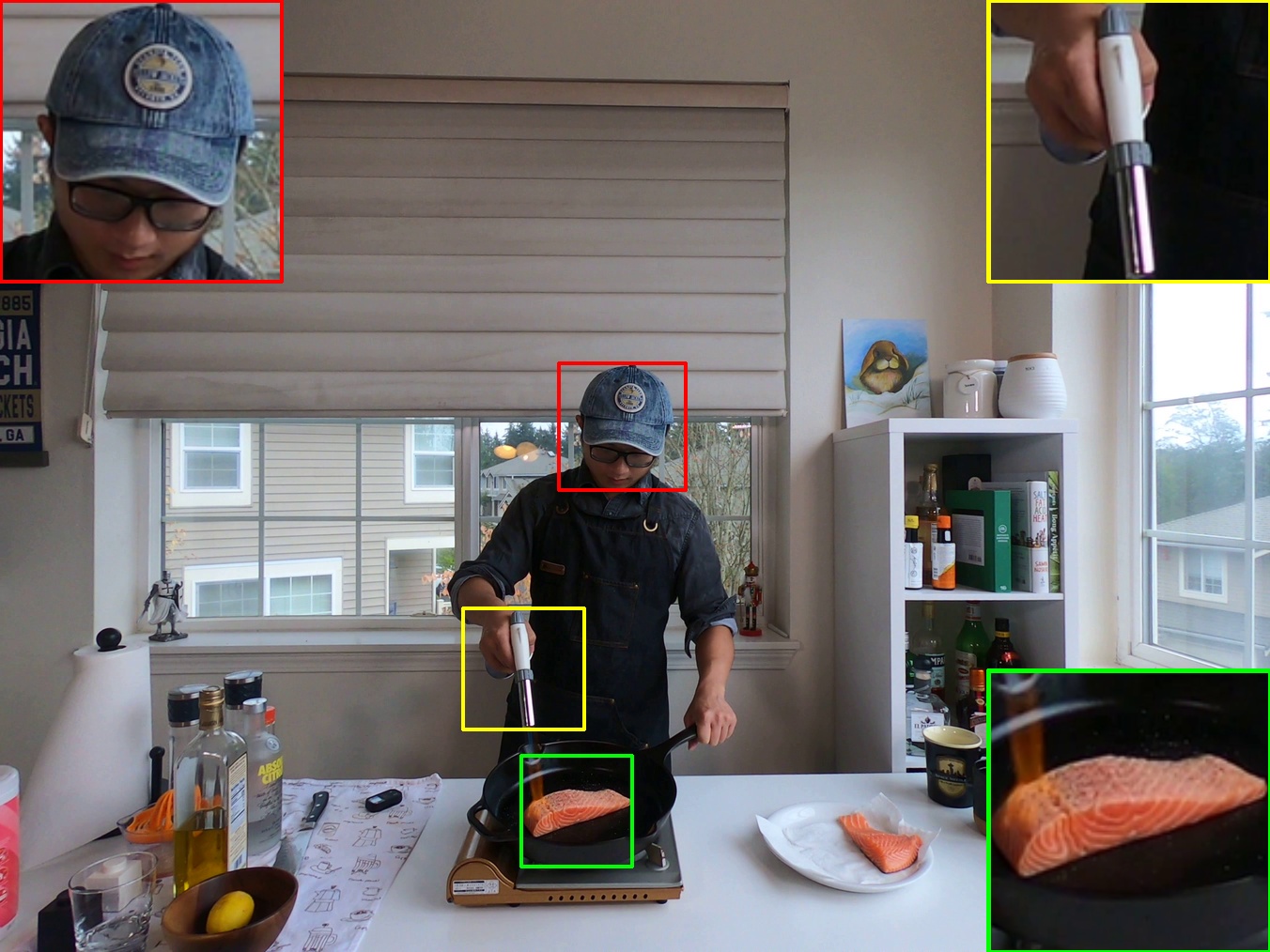}
   }  
   \caption{Qualitative comparison of ours with the benchmark methods on \textit{Flame Salmon} scene of N3DV dataset.}
   \label{fig:salmon}
   \vspace{-0.2cm}
\end{figure*}

\begin{figure*}[!t]
   \centering
   \subfloat[TeTriRF~\cite{wu2024tetrirf}]{
      \includegraphics[width=0.24\textwidth]{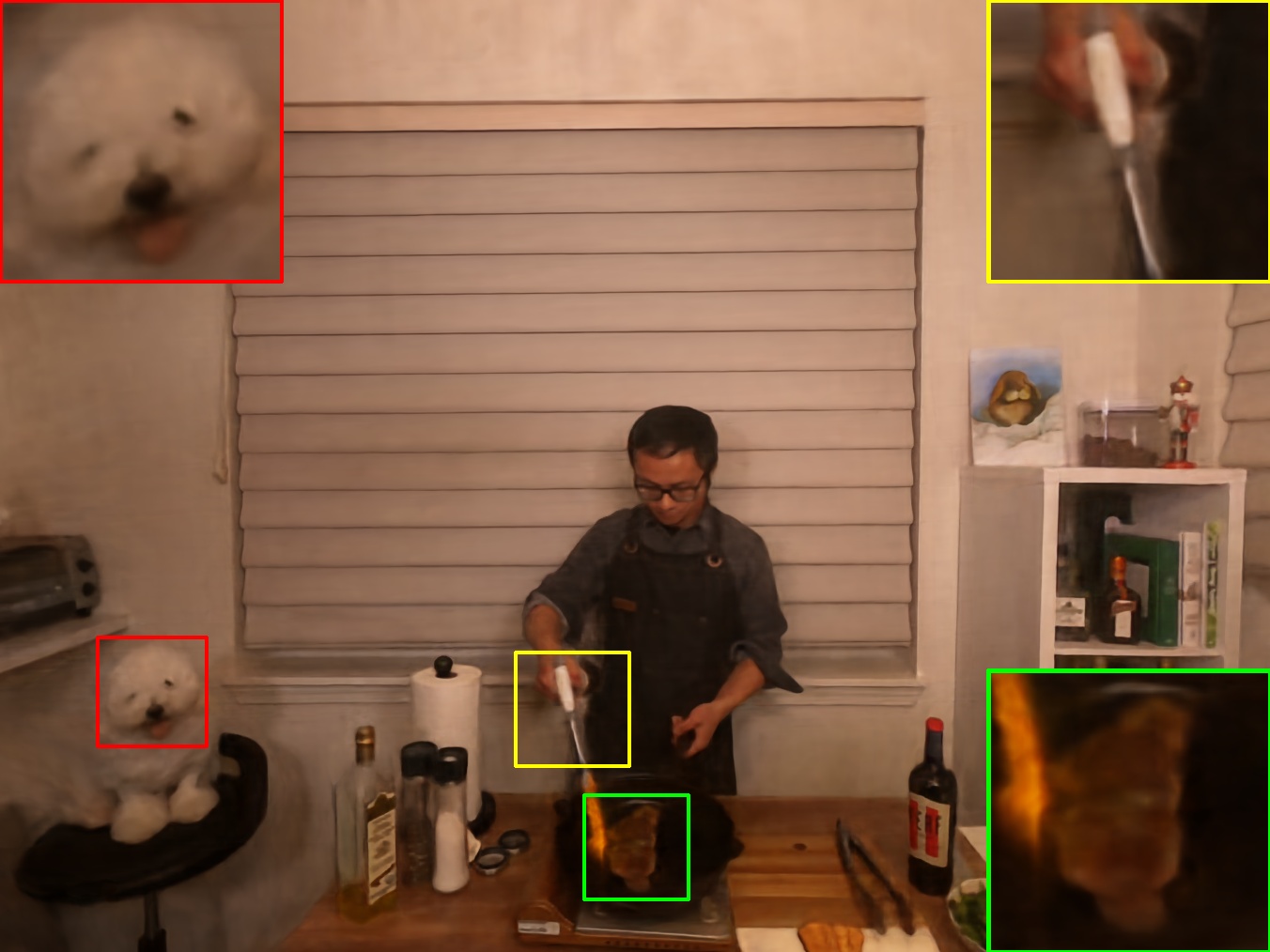}
   }
   \subfloat[3DGStream~\cite{sun20243dgstream}]{
      \includegraphics[width=0.24\textwidth]{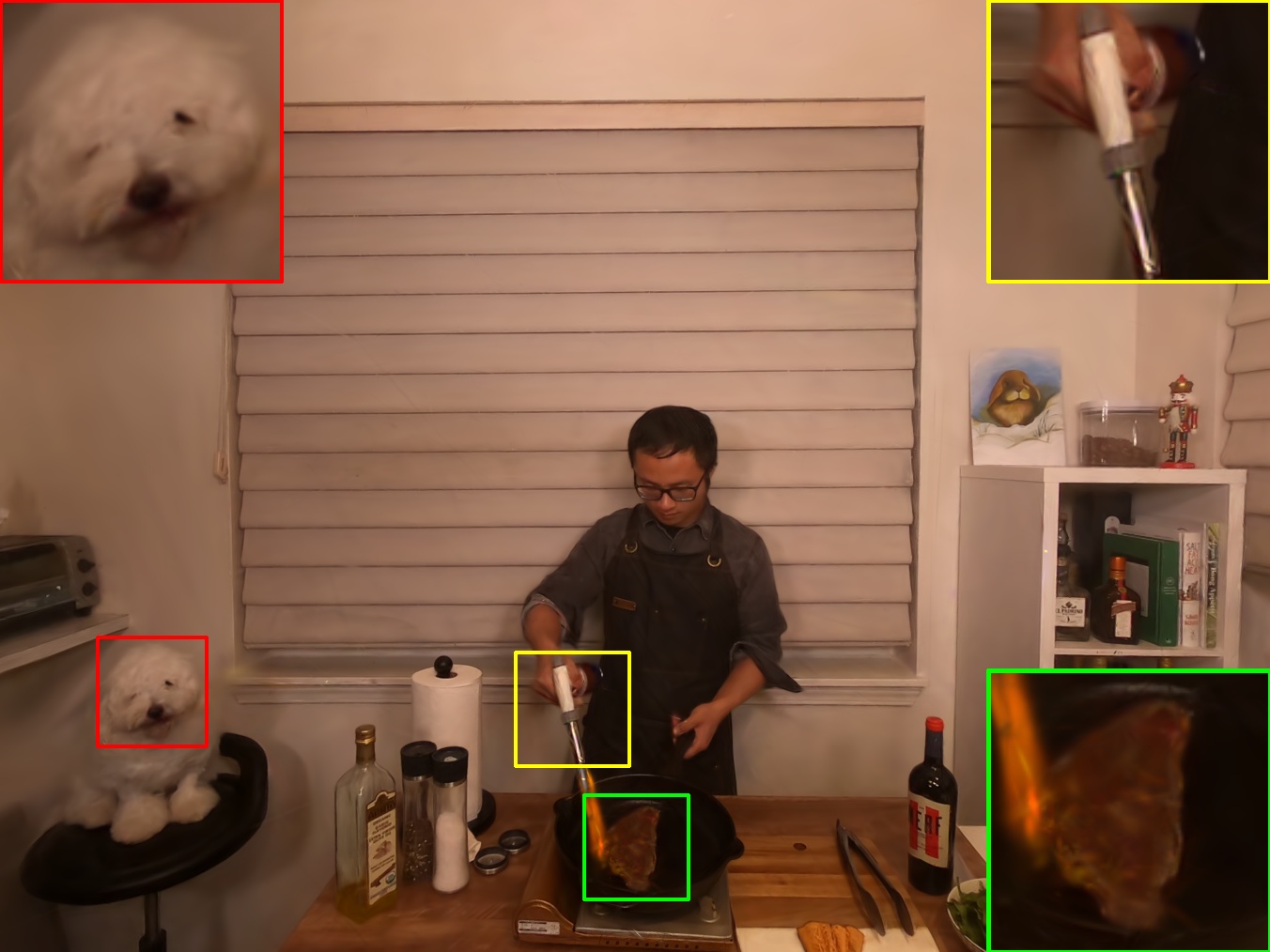}
   }
   \subfloat[HiCoM~\cite{gao2024hicom}]{
      \includegraphics[width=0.24\textwidth]{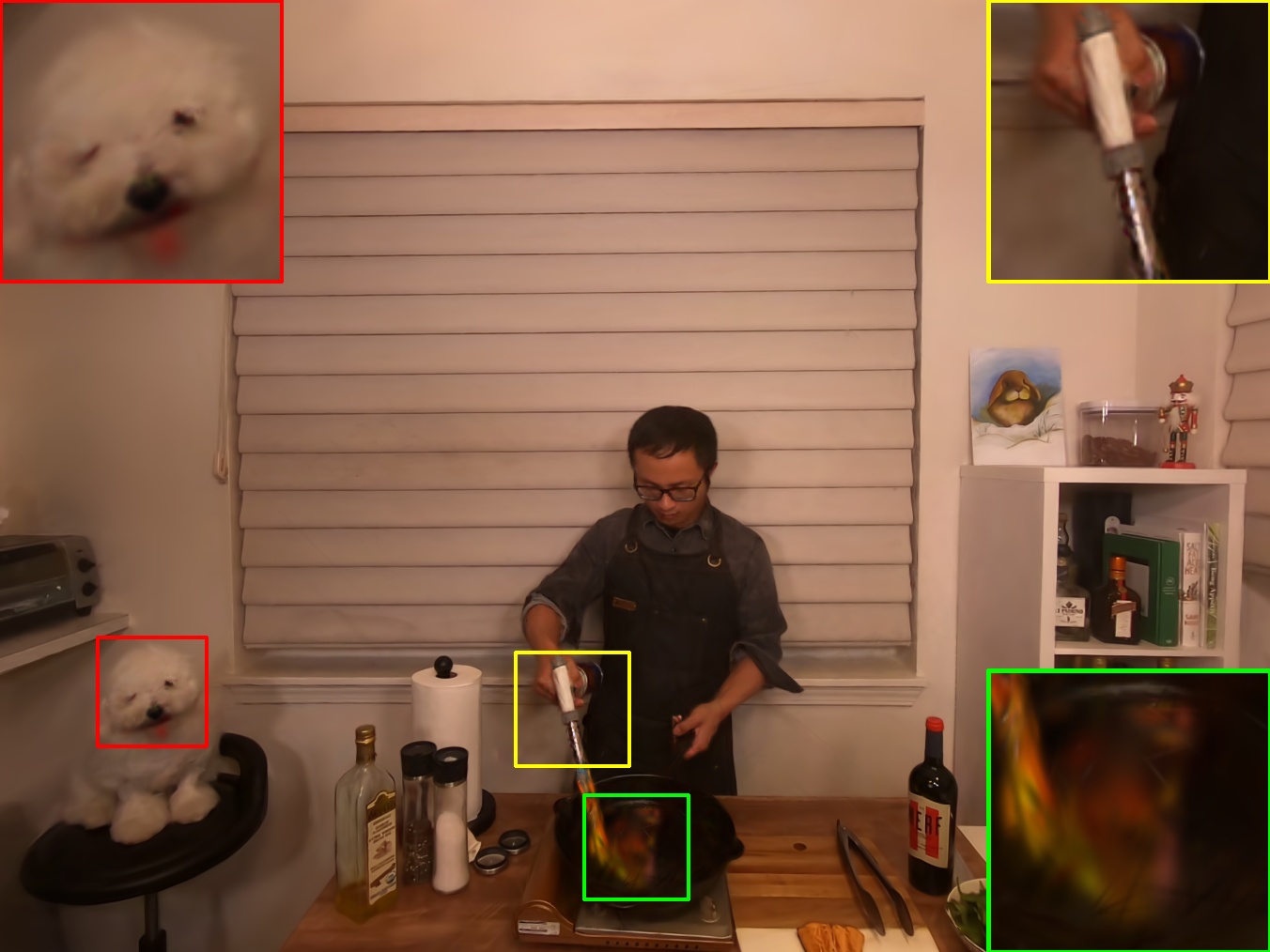}
   }
   \subfloat[VideoGS~\cite{wang2024v}]{
      \includegraphics[width=0.24\textwidth]{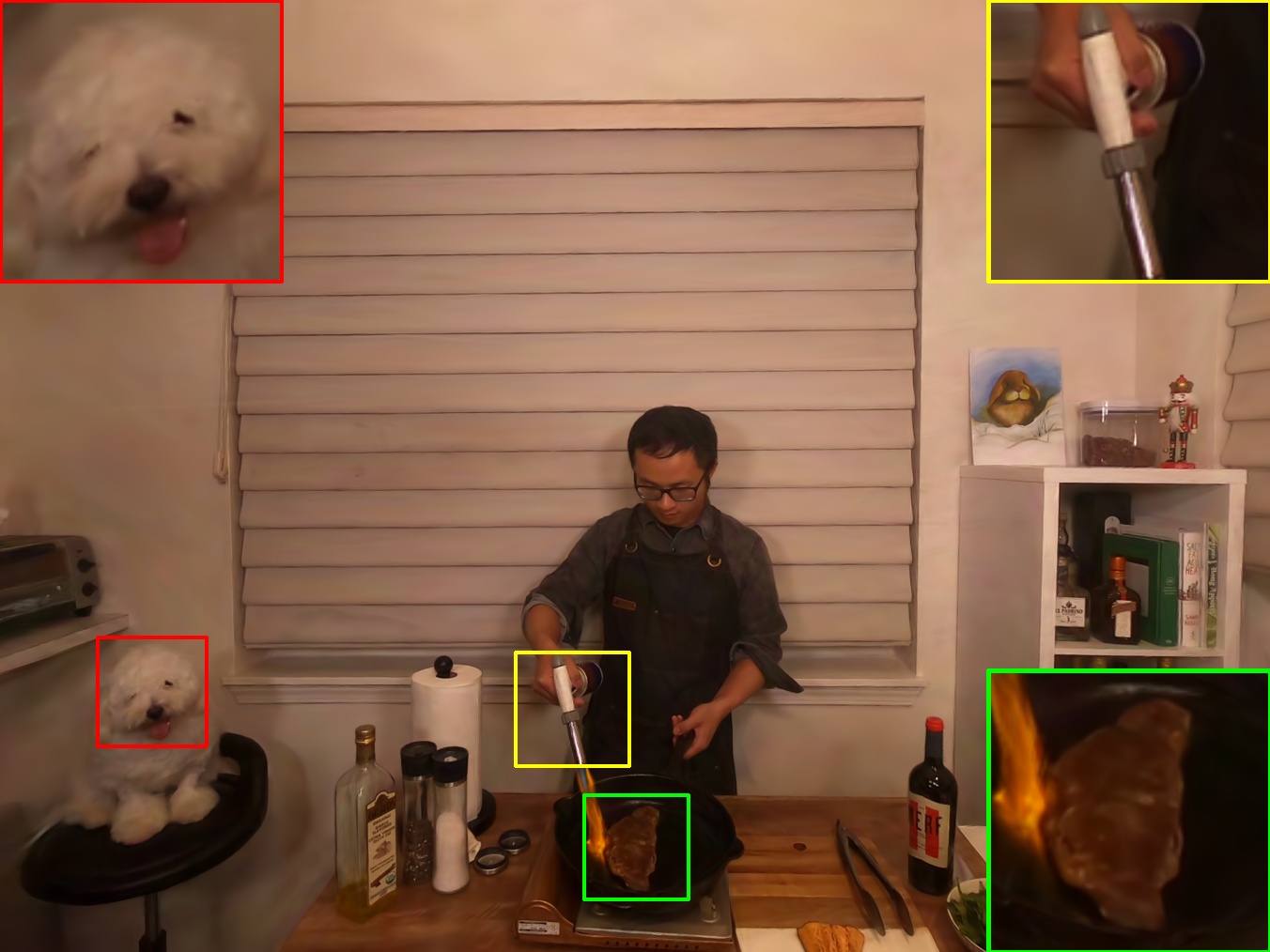}
   }
   \ 
   \subfloat[4DGC~\cite{hu20254dgc}]{
      \includegraphics[width=0.32\textwidth]{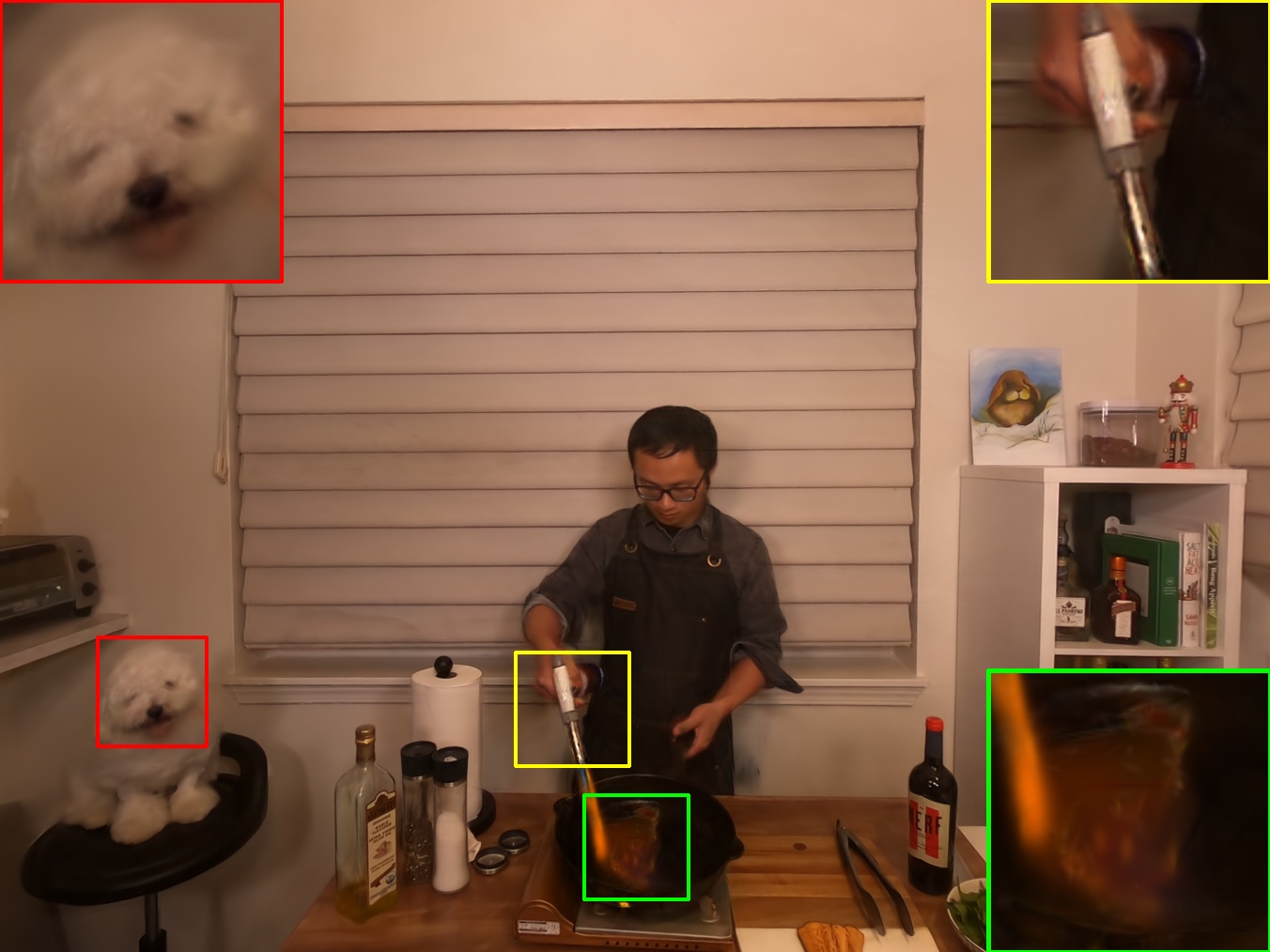}
   }
   \subfloat[Ours]{
      \includegraphics[width=0.32\textwidth]{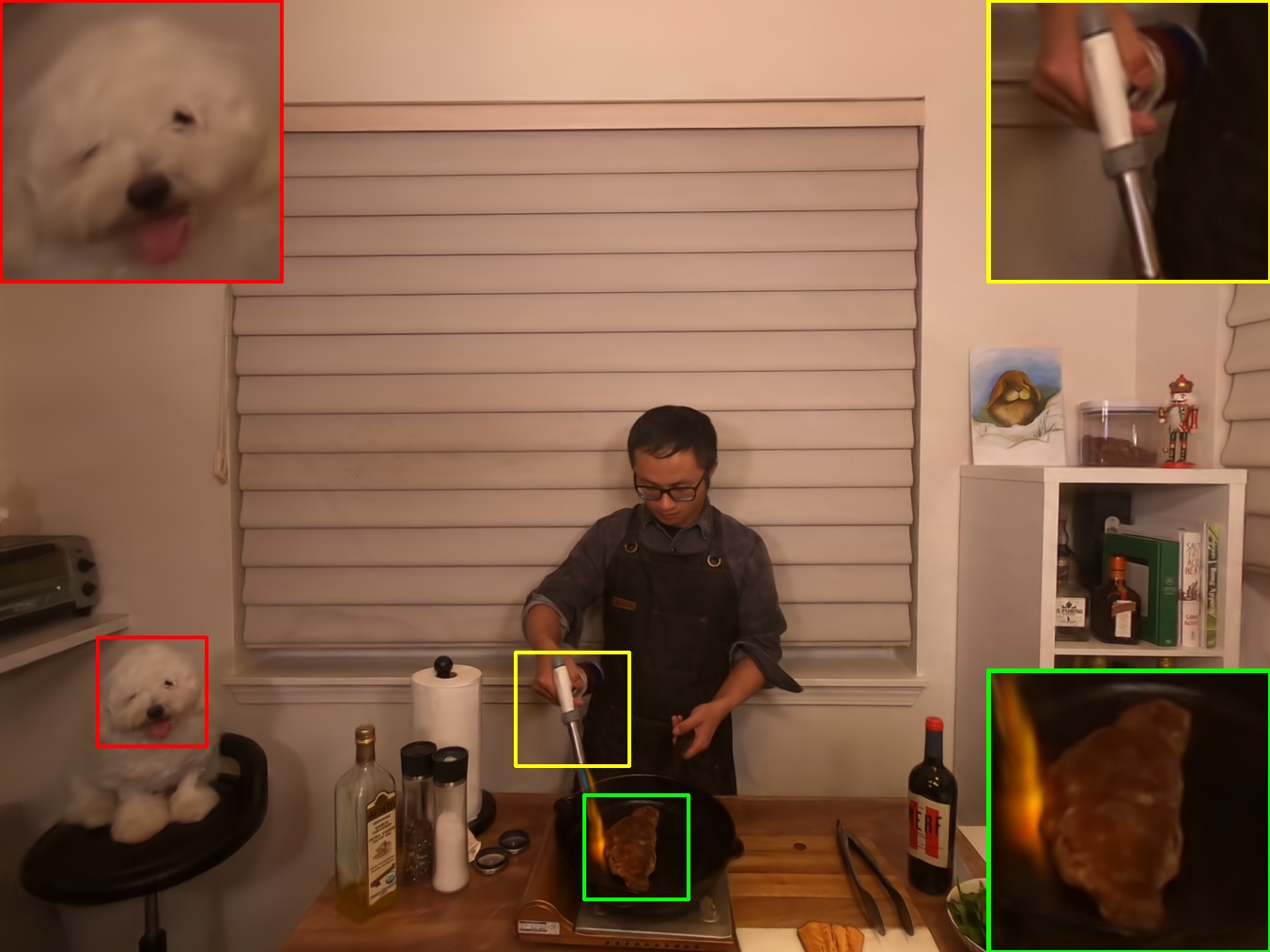}
   }
   \subfloat[GroudTruth]{
      \includegraphics[width=0.32\textwidth]{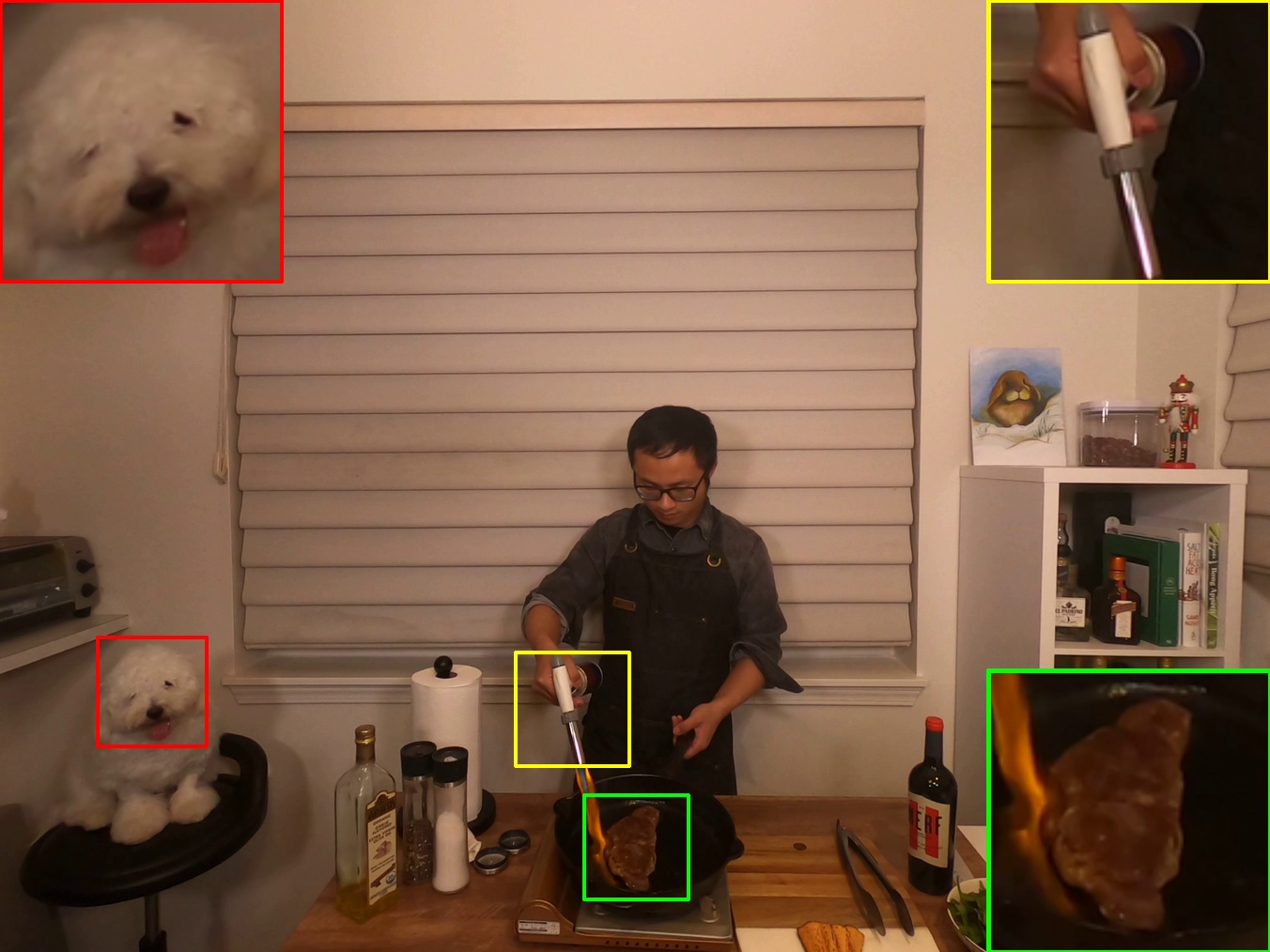}
   }   
   \caption{Qualitative comparison of ours with the benchmark methods on \textit{Flame Steak} scene of N3DV dataset.}
   \label{fig:steak}
   \vspace{-0.2cm}
\end{figure*}

\begin{figure*}[!t]
   \centering
   \subfloat[TeTriRF~\cite{wu2024tetrirf}]{
      \includegraphics[width=0.24\textwidth]{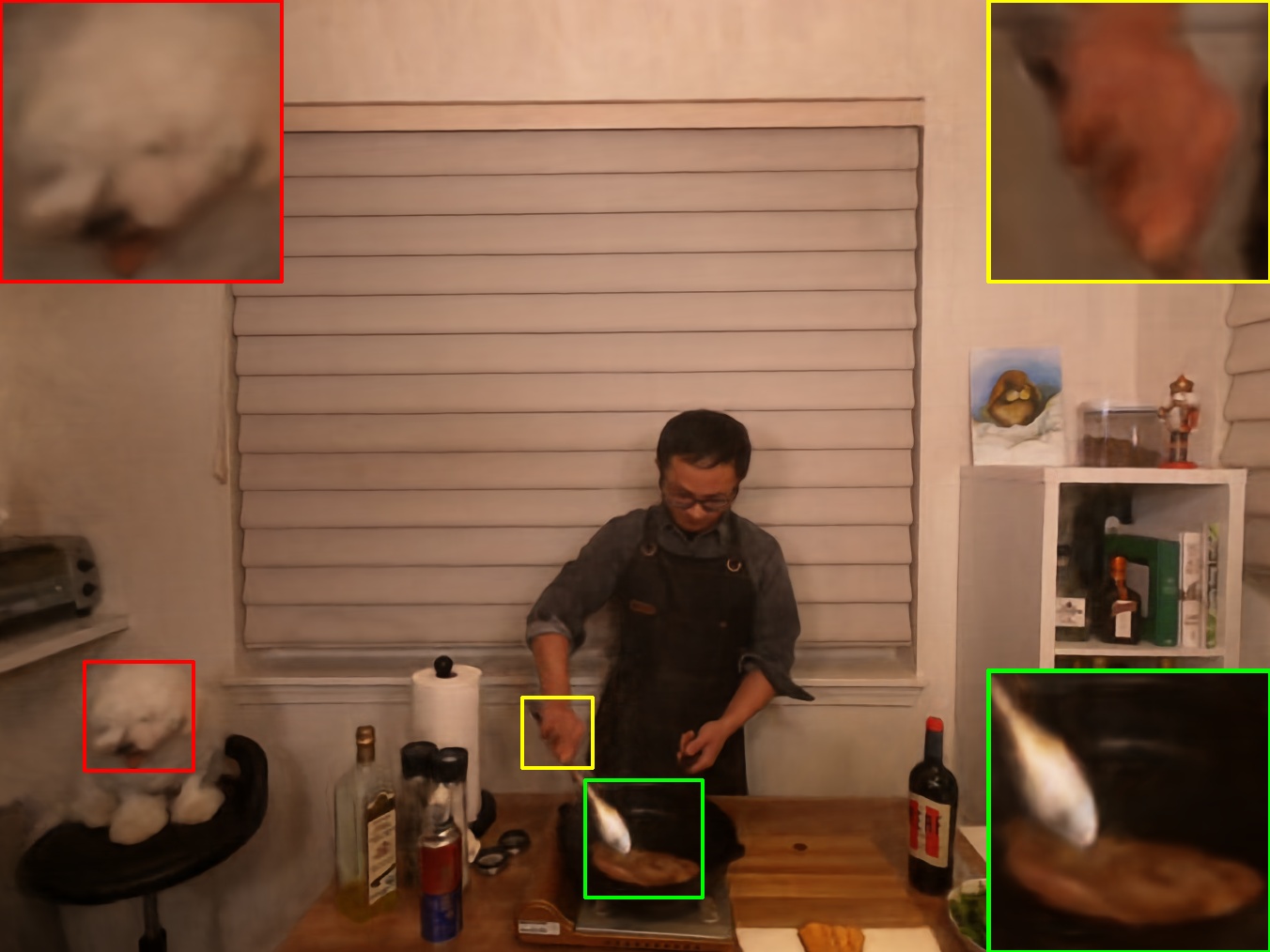}
   }
   \subfloat[3DGStream~\cite{sun20243dgstream}]{
      \includegraphics[width=0.24\textwidth]{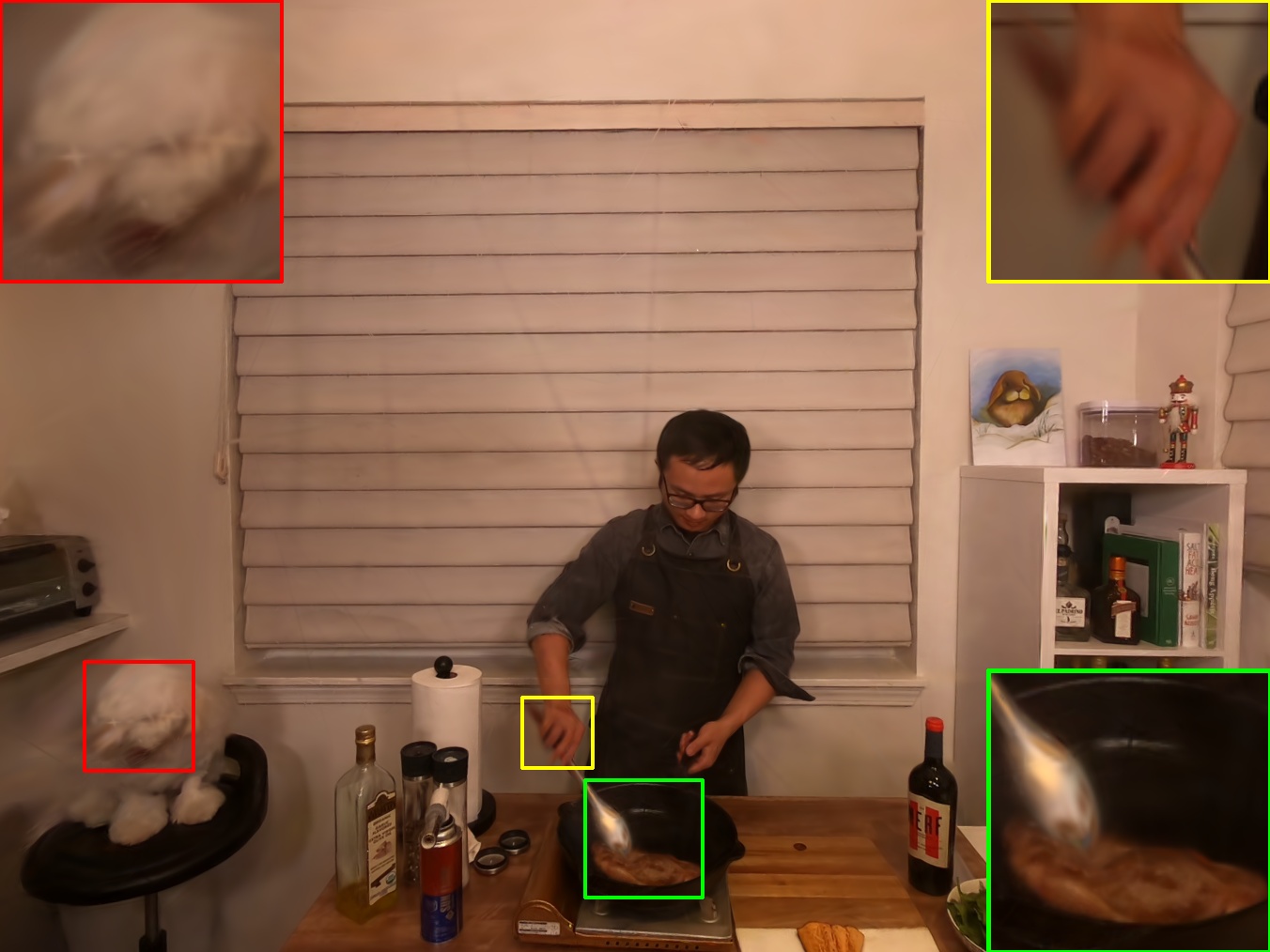}
   }
   \subfloat[HiCoM~\cite{gao2024hicom}]{
      \includegraphics[width=0.24\textwidth]{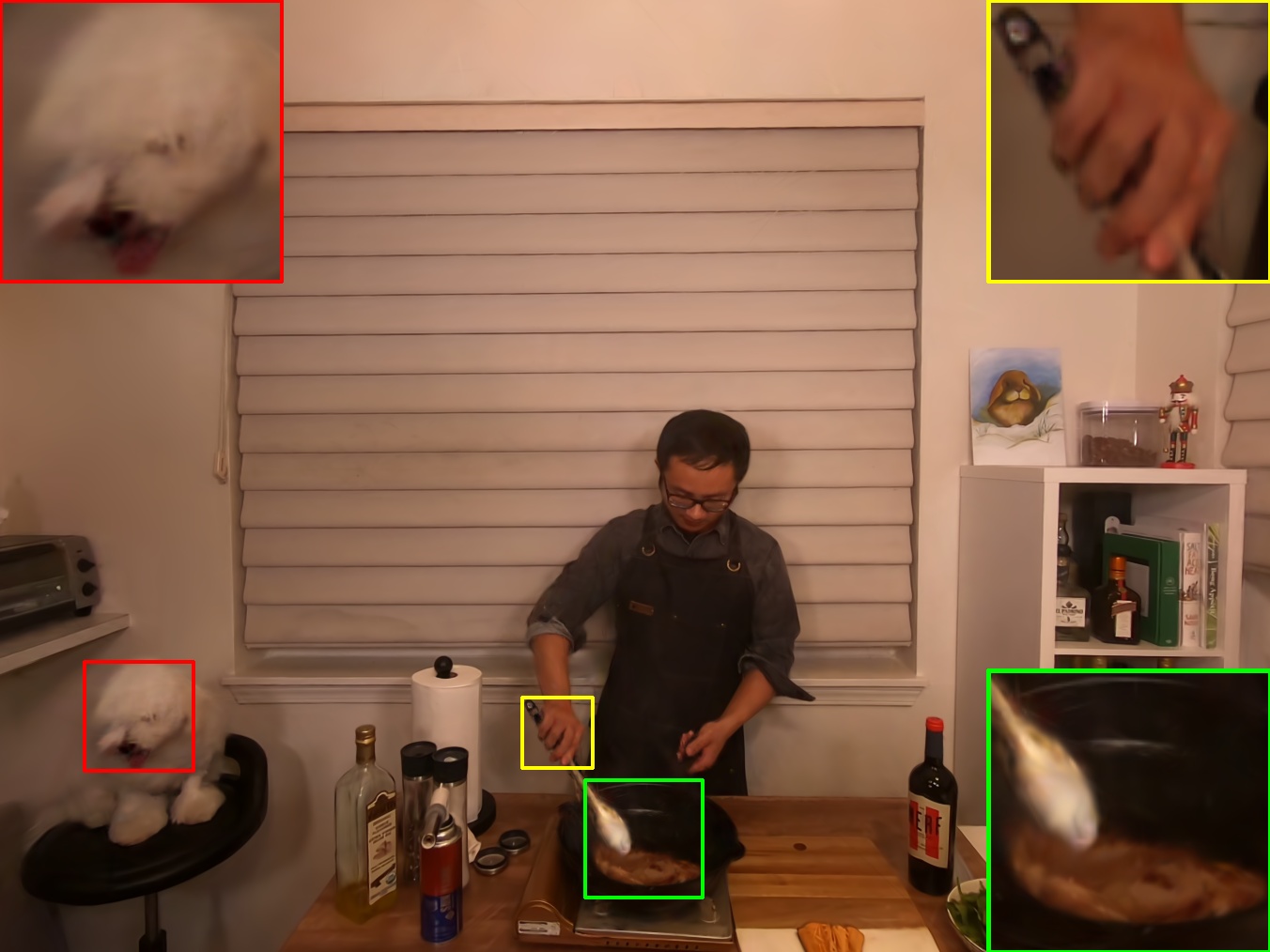}
   }
   \subfloat[VideoGS~\cite{wang2024v}]{
      \includegraphics[width=0.24\textwidth]{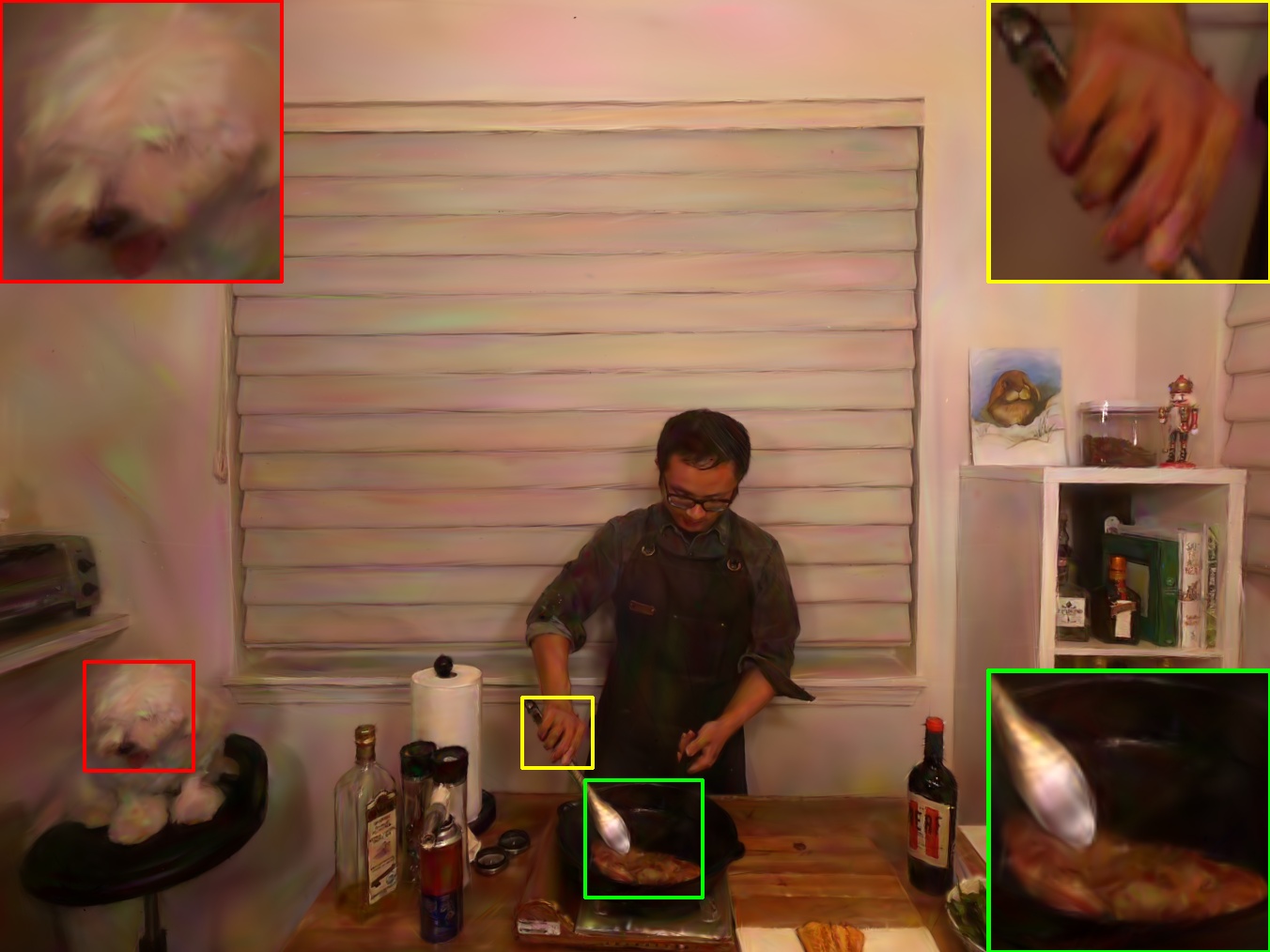}
   }
   \ 
   \subfloat[4DGC~\cite{hu20254dgc}]{
      \includegraphics[width=0.32\textwidth]{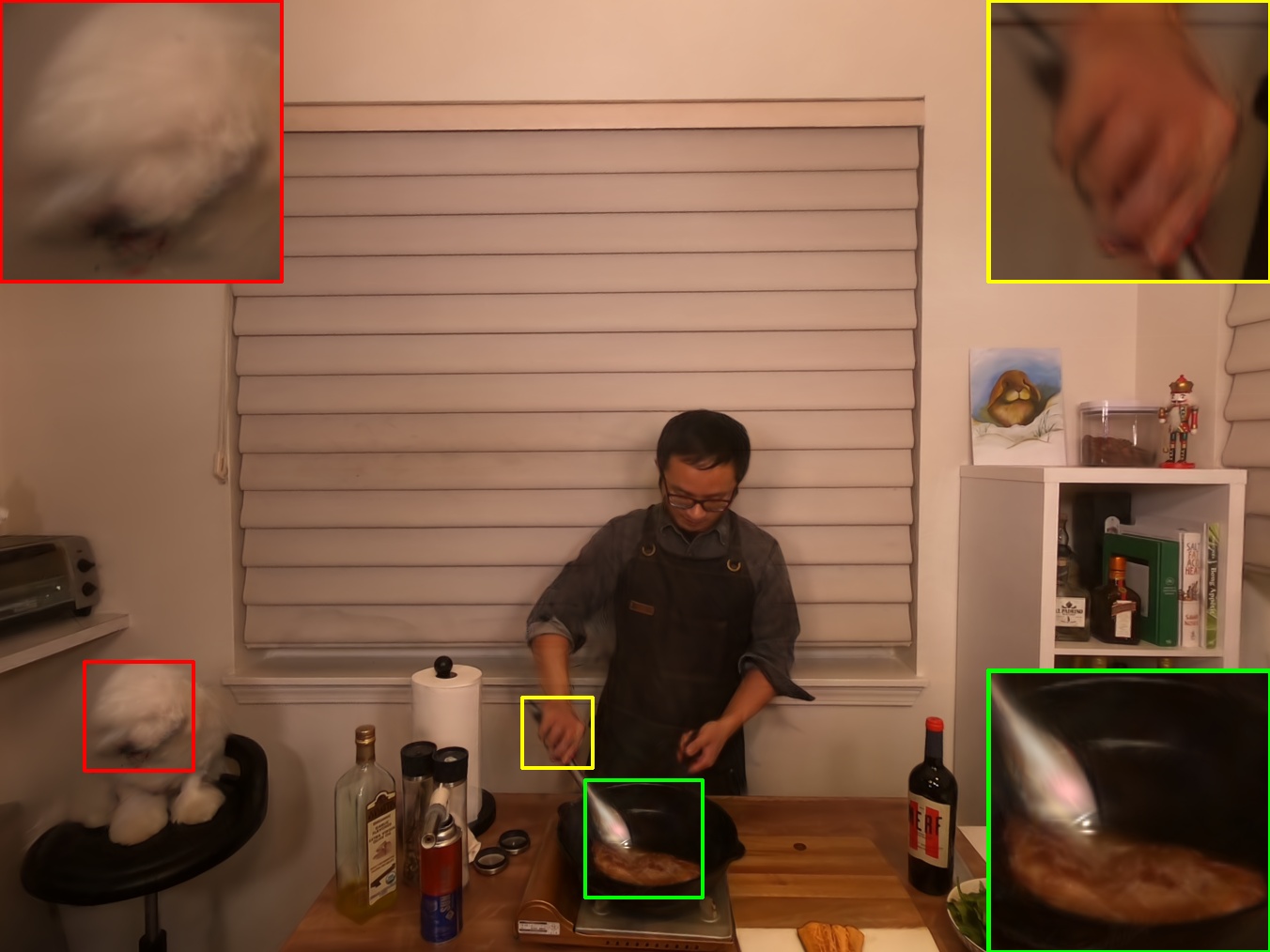}
   }
   \subfloat[Ours]{
      \includegraphics[width=0.32\textwidth]{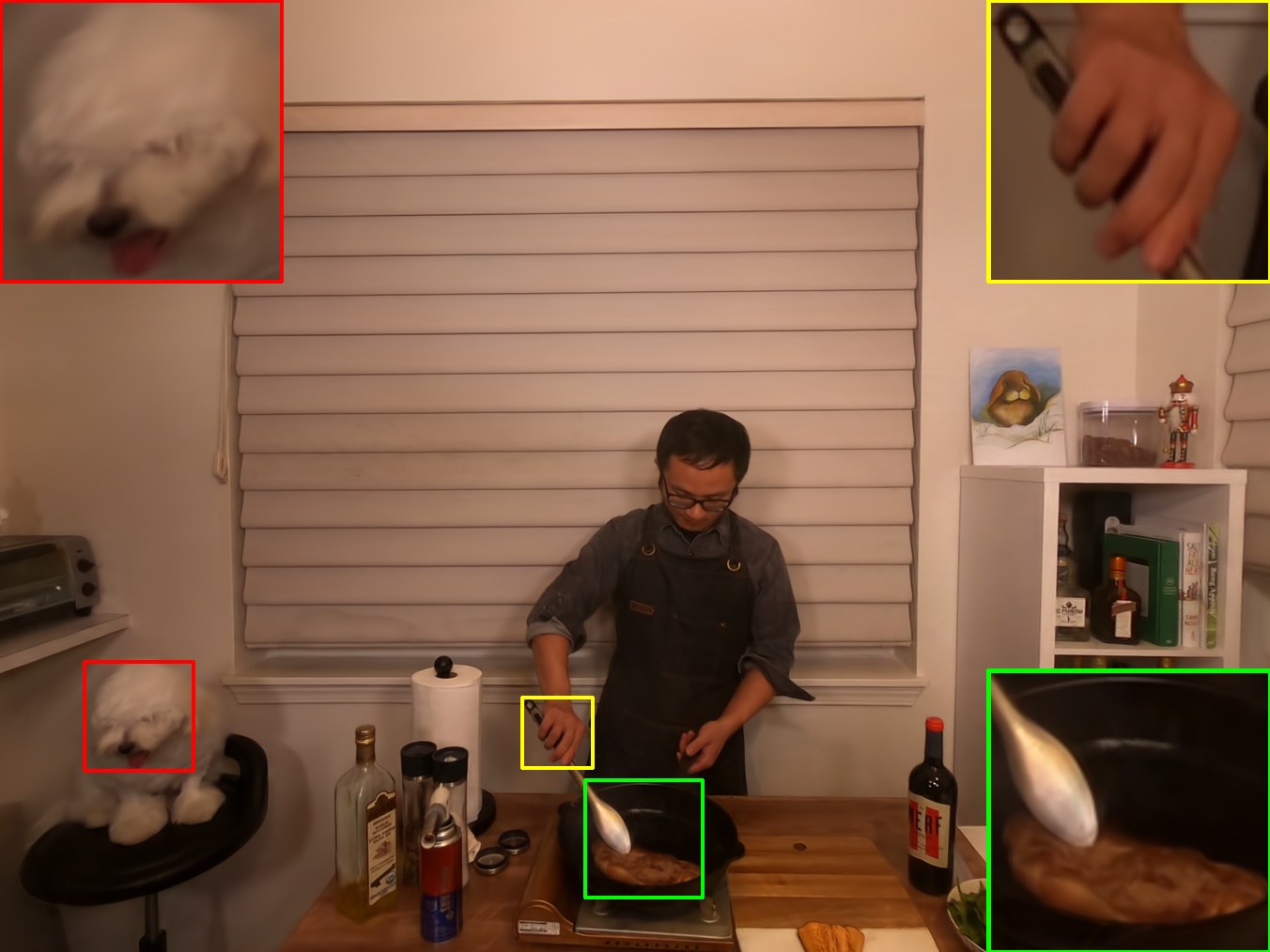}
   }
   \subfloat[GroudTruth]{
      \includegraphics[width=0.32\textwidth]{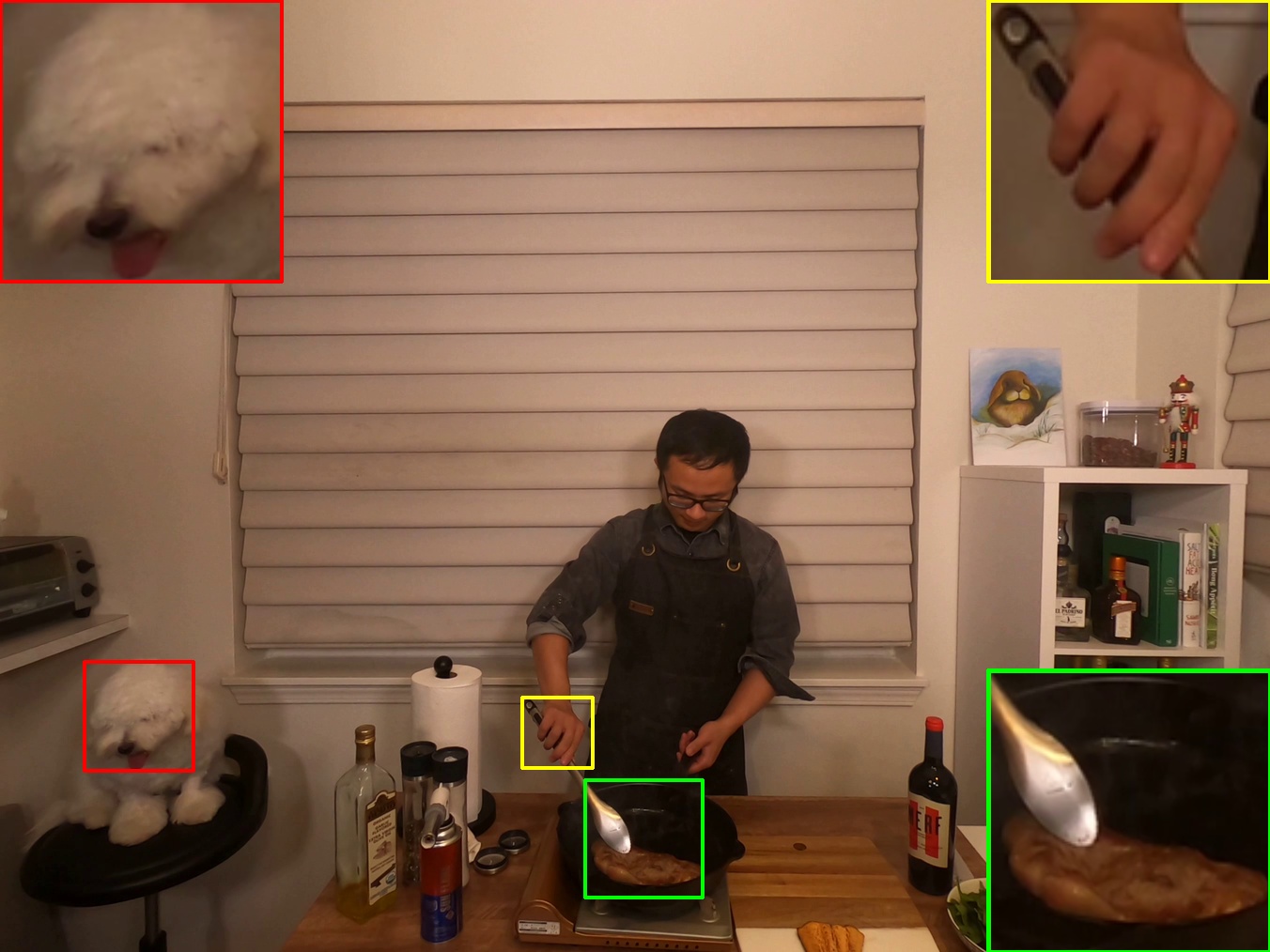}
   }   
   \caption{Qualitative comparison of ours with the benchmark methods on \textit{Sear Steak} scene of N3DV dataset.}
   \label{fig:sear}
   \vspace{-0.2cm}
\end{figure*}

\begin{figure*}[!t]
   \centering
   \subfloat[3DGStream]{
      \includegraphics[width=0.192\textwidth]{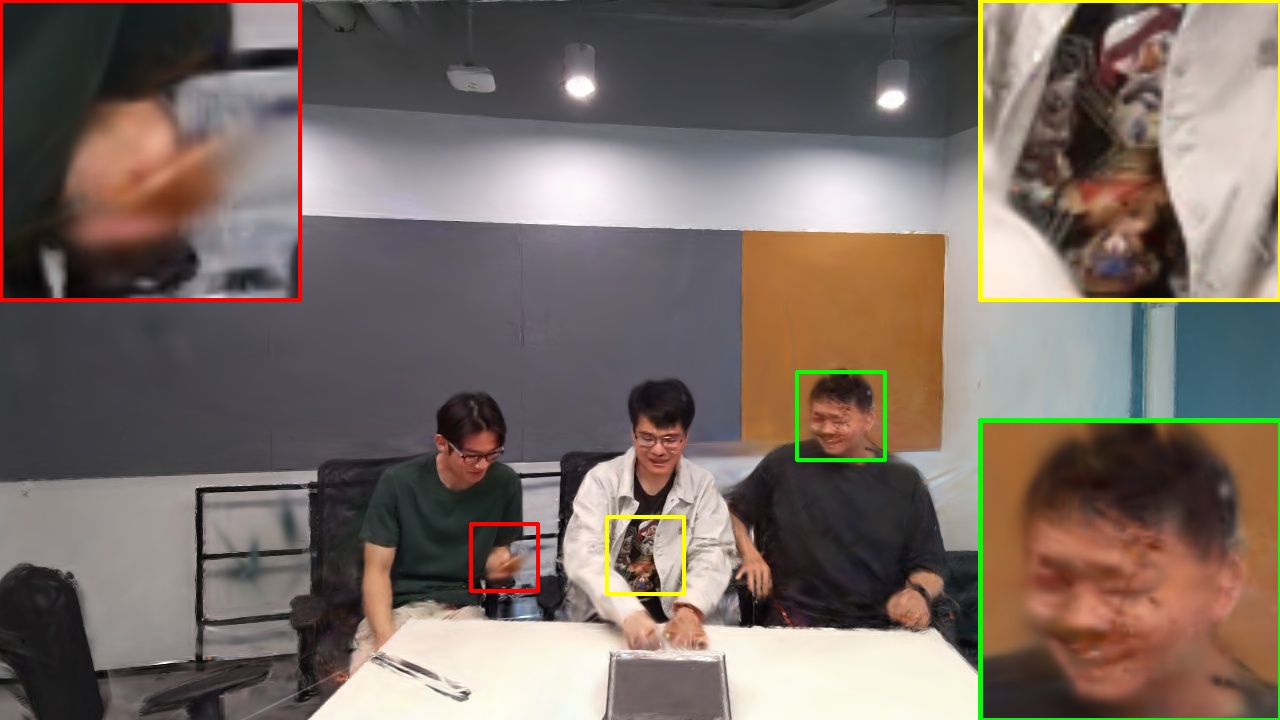}
   }
   \subfloat[HiCoM]{
      \includegraphics[width=0.192\textwidth]{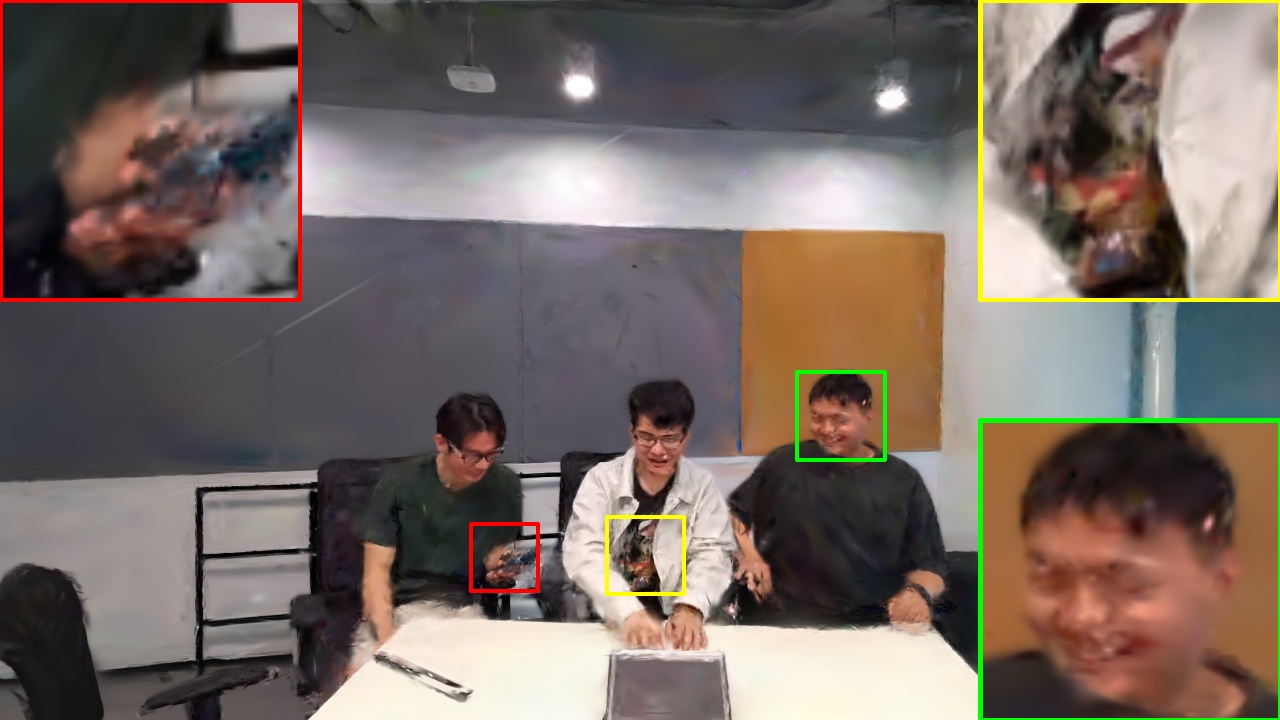}
   }
   \subfloat[4DGC]{
      \includegraphics[width=0.192\textwidth]{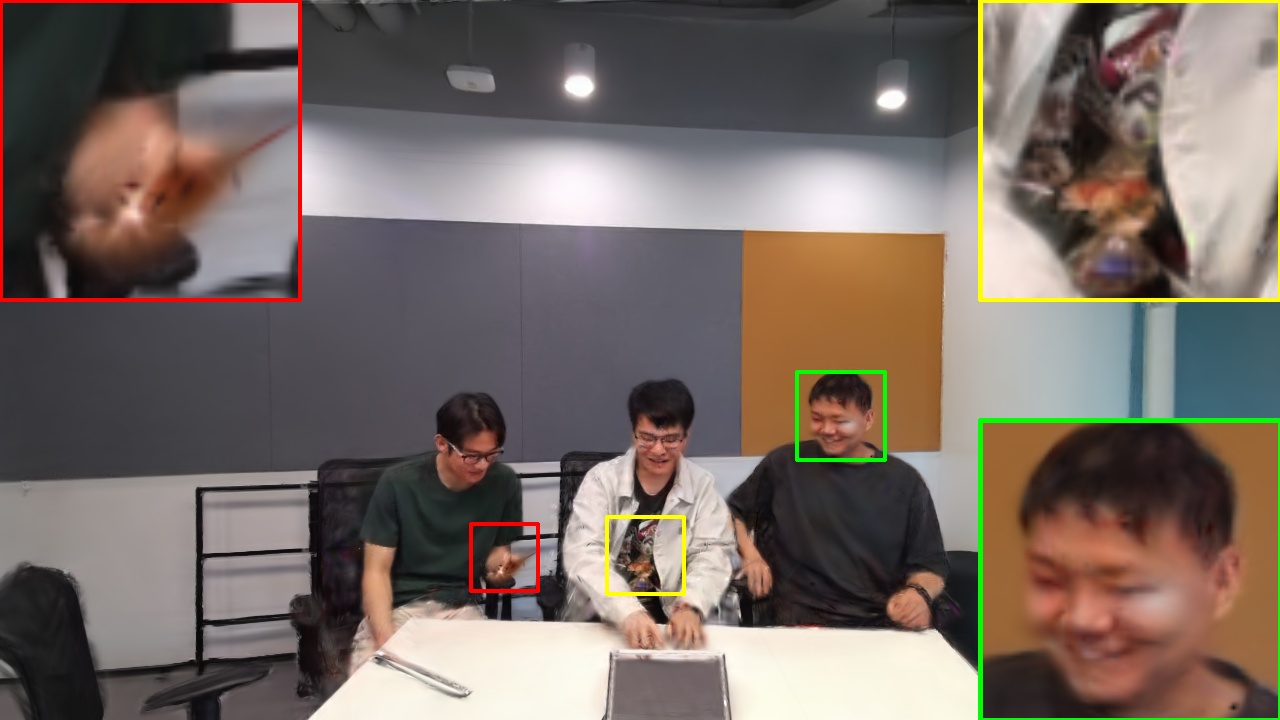}
   }
   \subfloat[Ours]{
      \includegraphics[width=0.192\textwidth]{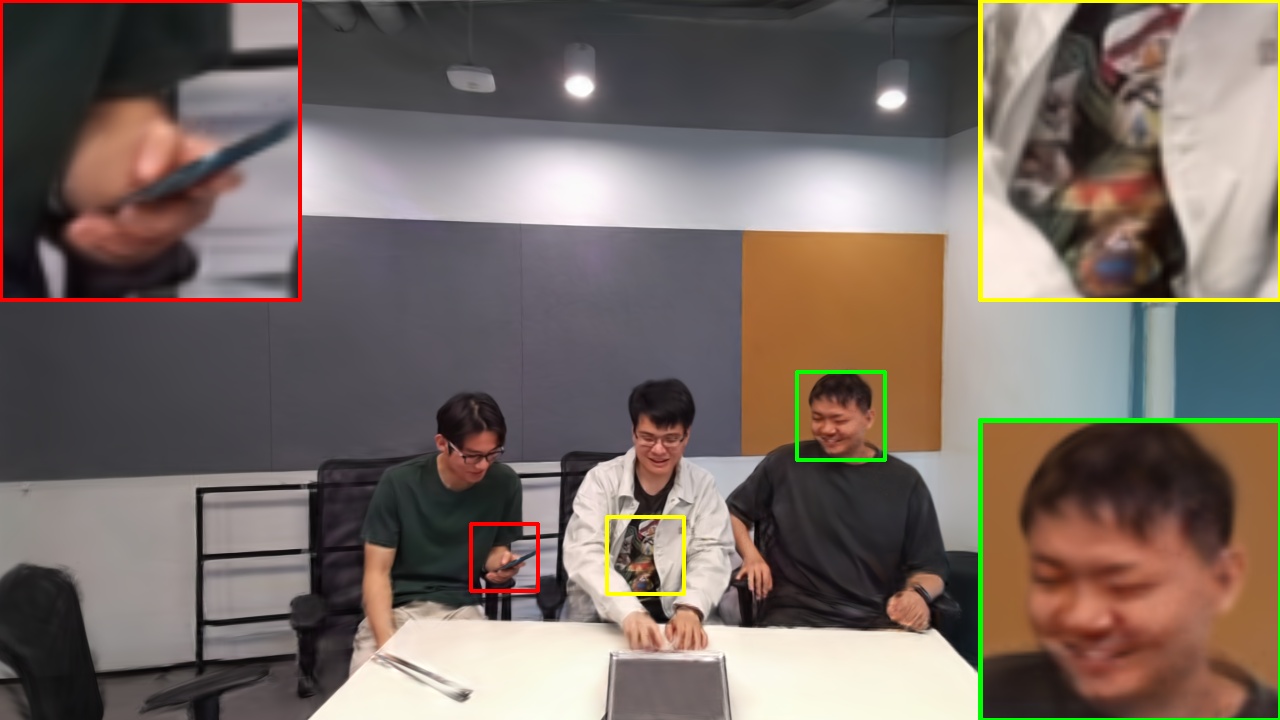}
   }
   \subfloat[GroudTruth]{
      \includegraphics[width=0.192\textwidth]{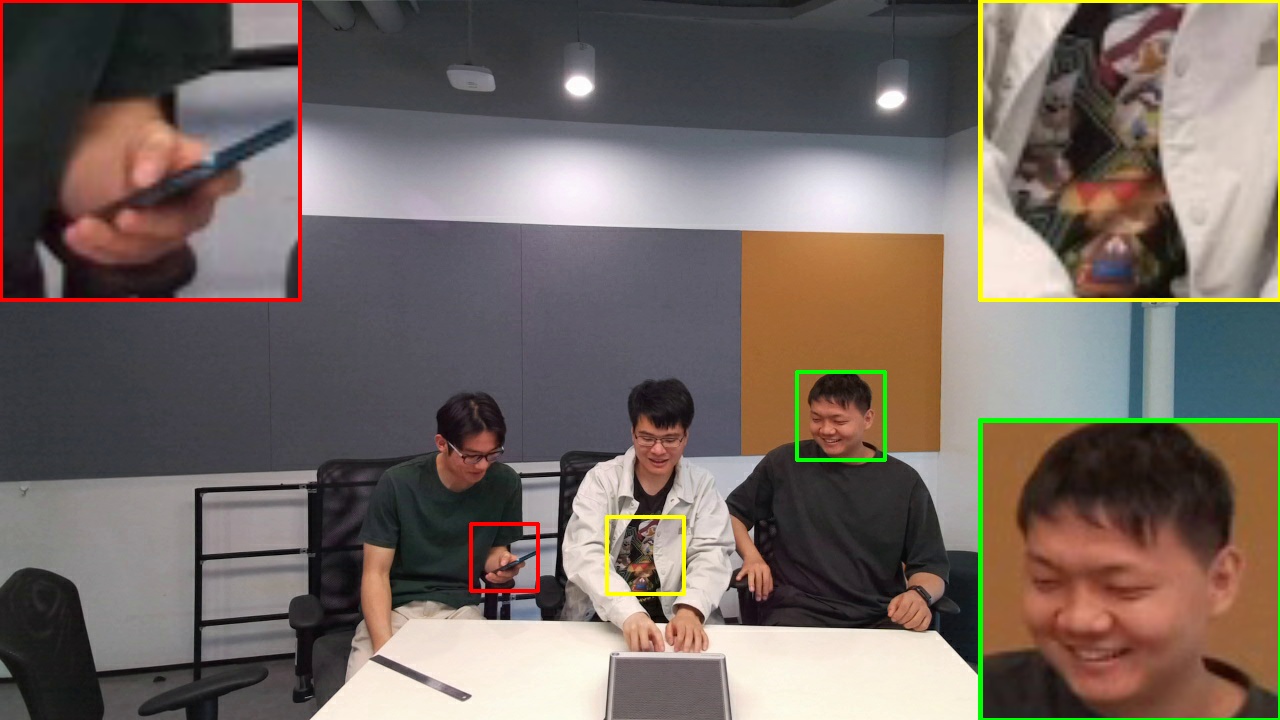}
   }
   \caption{Qualitative comparison of ours with the benchmark methods on \textit{Discussion} scene of MeetRoom dataset.}
   \label{fig:discussion}
   \vspace{-0.2cm}
\end{figure*}

\begin{figure*}[!t]
   \centering
   \subfloat[3DGStream]{
      \includegraphics[width=0.192\textwidth]{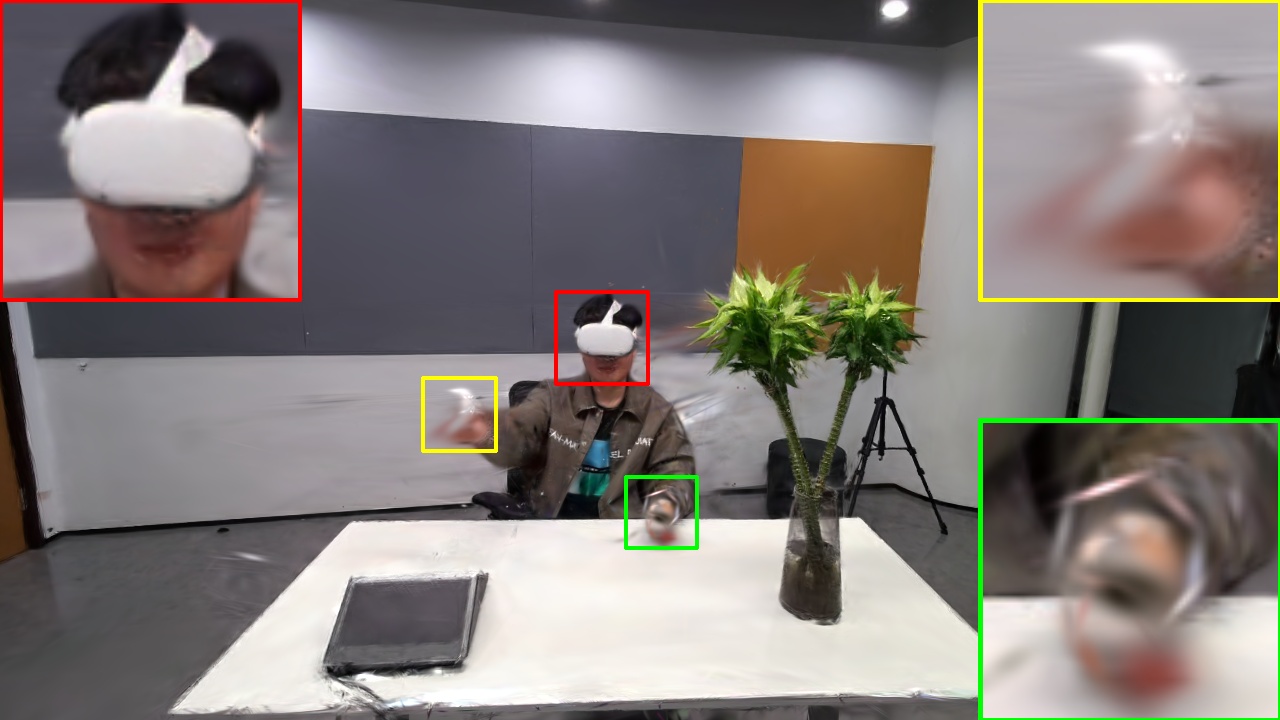}
   }
   \subfloat[HiCoM]{
      \includegraphics[width=0.192\textwidth]{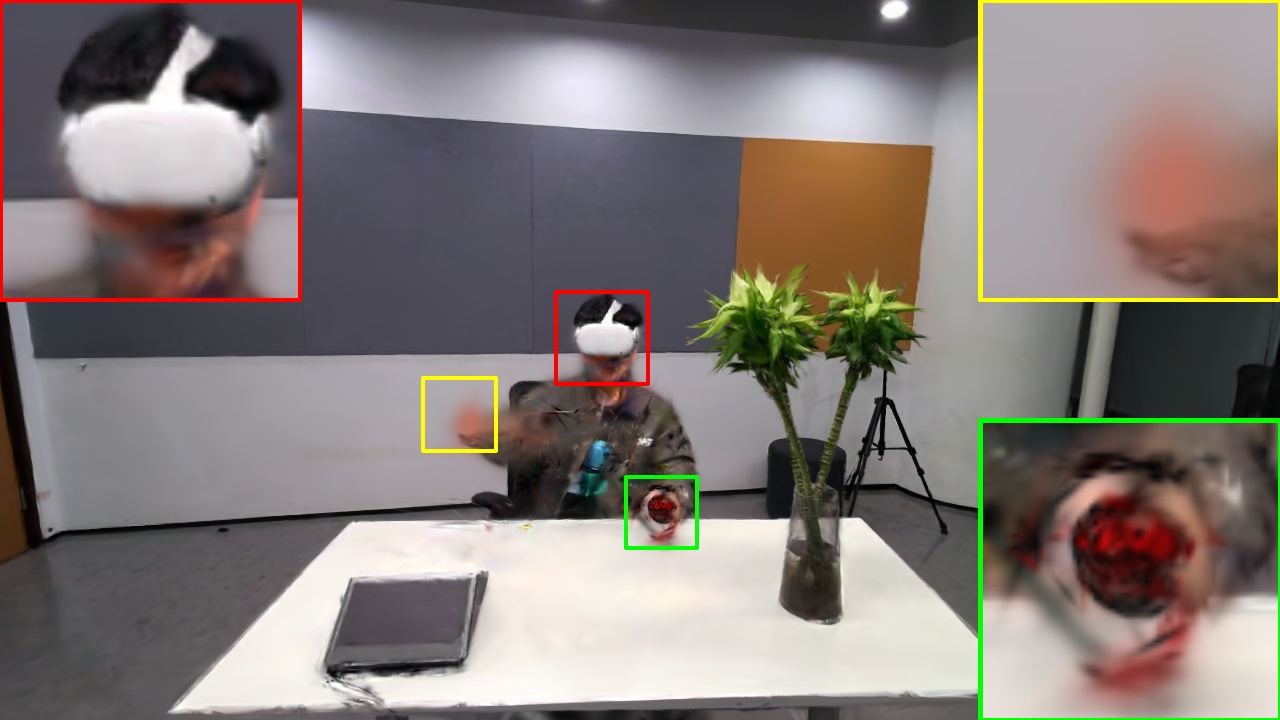}
   }
   \subfloat[4DGC]{
      \includegraphics[width=0.192\textwidth]{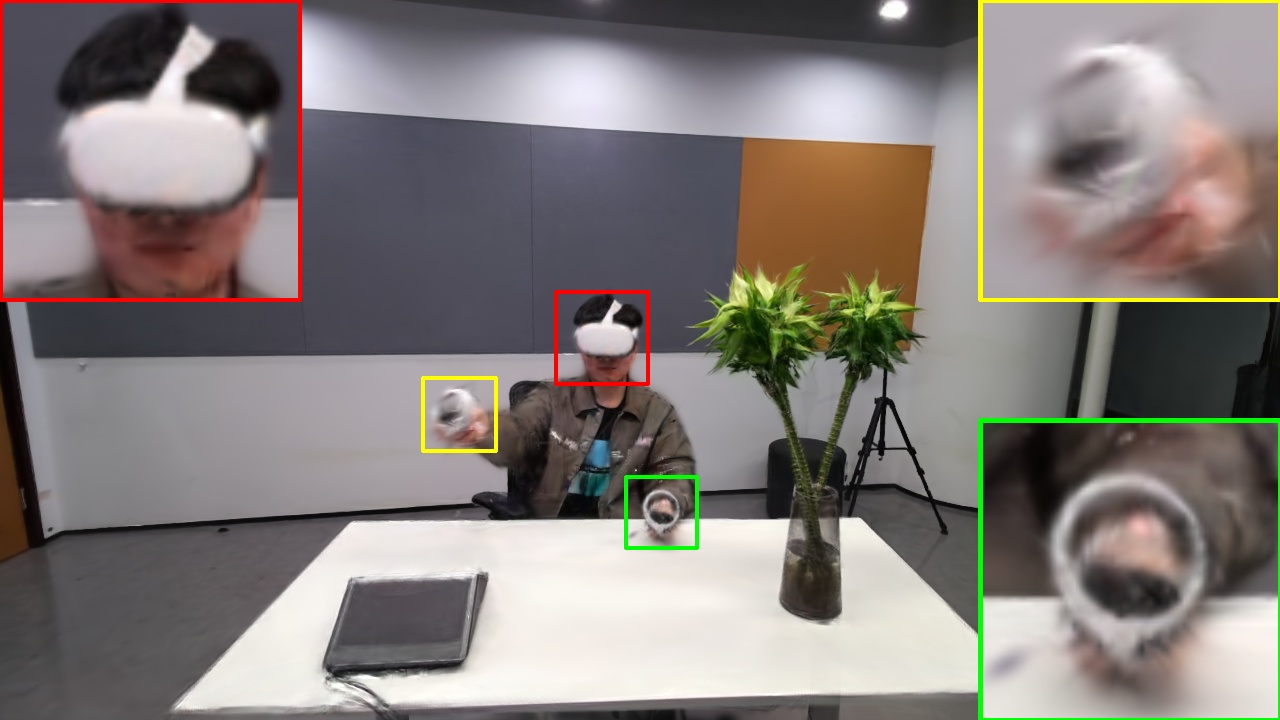}
   }
   \subfloat[Ours]{
      \includegraphics[width=0.192\textwidth]{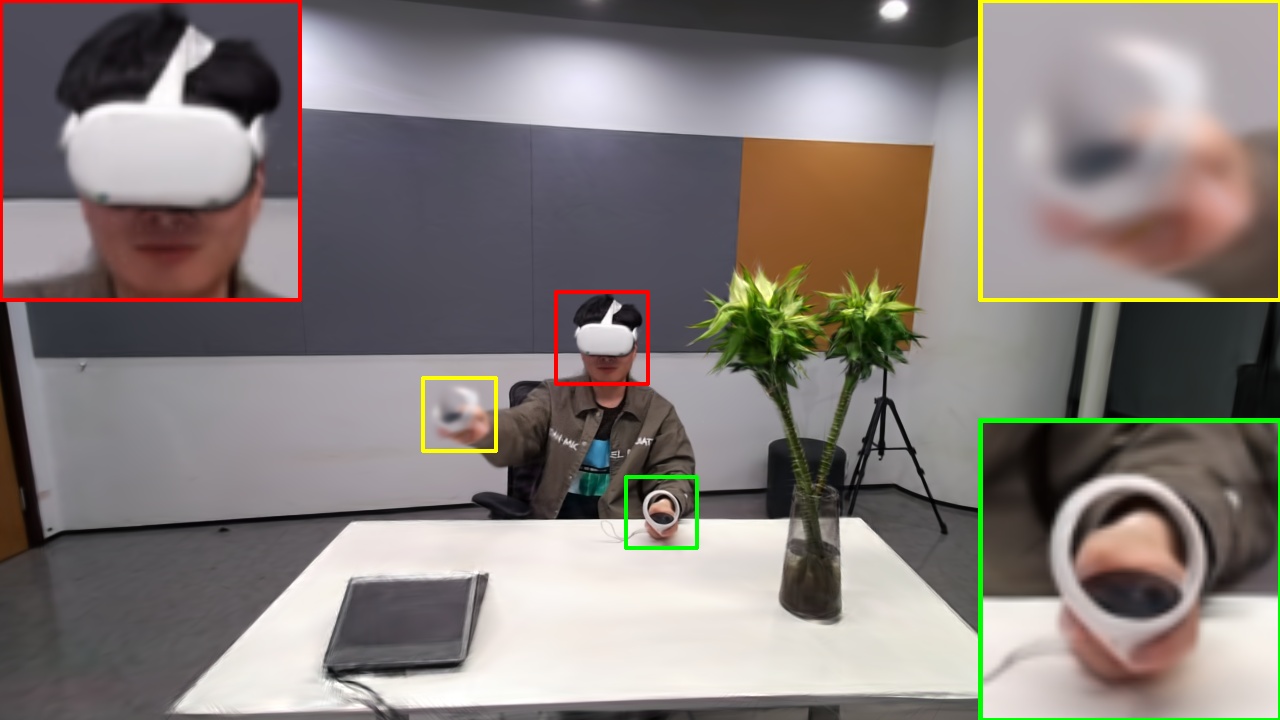}
   }
   \subfloat[GroudTruth]{
      \includegraphics[width=0.192\textwidth]{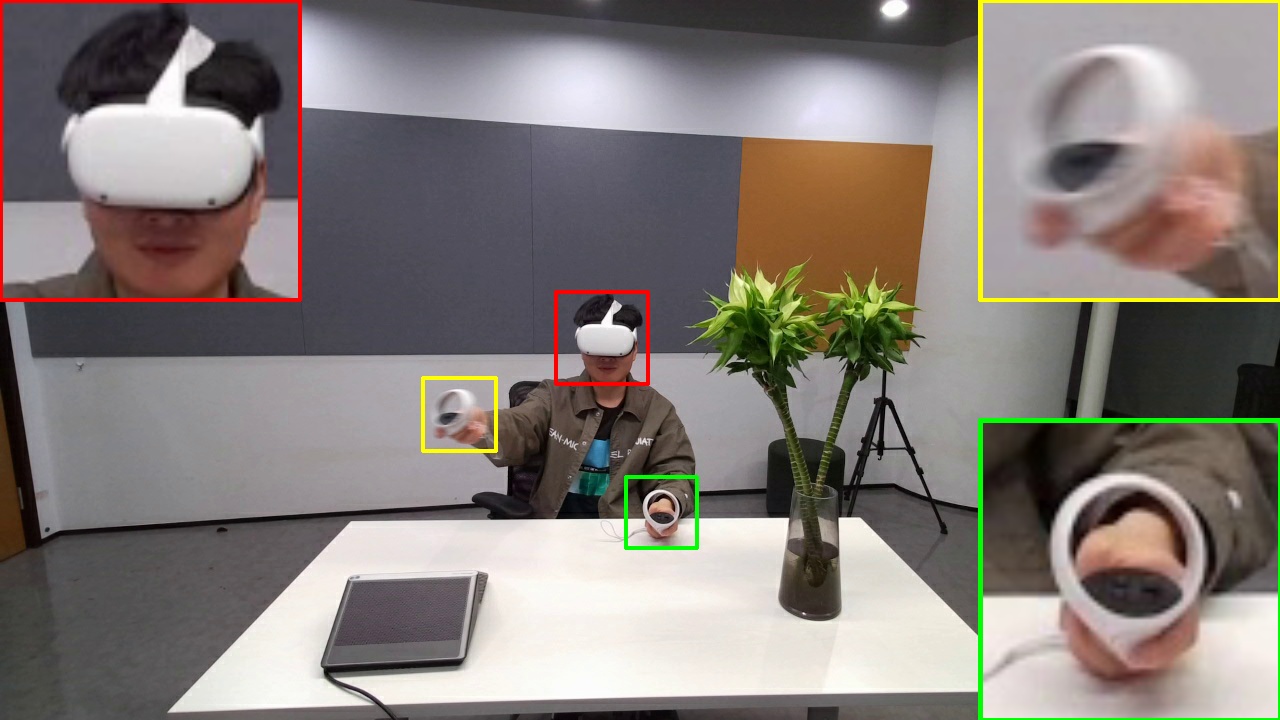}
   }
   \caption{Qualitative comparison of ours with the benchmark methods on \textit{Vrheadset} scene of MeetRoom dataset.}
   \label{fig:vrheadset}
   \vspace{-0.2cm}
\end{figure*}

In addition to the qualitative results in the paper, we show all scenes results in Fig.~\ref{fig:coffee} through Fig.~\ref{fig:vrheadset}. We have incorporated results of 3DGStream~\cite{sun20243dgstream} and VideoGS~\cite{wang2024v} for comparsion. VideoGS~\cite{wang2024v} also uses video codecs to compress each channel of all Gaussian attributes, generating 23 feature videos and incurring long decoding latency. Furthermore, direct compression of Gaussian attributes leads to significant performance degradation in both dynamic areas and static backgrounds. Instead, our method encodes Gaussian spatial attributes using lossless JPEG-XL and temporal features as a video. To mitigate decoding latency, five spatial images of the subsequent GOP are pre-decoded. Consequently, only the temporal feature video requires real-time decoding. Although compression of temporal features introduces a minor performance loss, several improvements minimize its impact. For example, we add a noise to the temporal features during training to emulate compression loss. Then, the deformation field serves as a denoiser, further improving performance. Nevertheless, the temporal regularization encourages static Gaussians in the temporal features to converge to 0, thereby minimizing the effect of temporal feature compression on static areas.

\section{Per-GOP Quantitative Results}

We show results for each GOP in the N3DV dataset of various metrics, including PSNR, SSIM, LPIPS(VGG) in Table~\ref{fig:per-gop}. StreamSTGS trains in GOP units, effectively solving the cumulative error generated by frame-by-frame training methods. As the GOP grows, there is no decrease in performance.

\section{Compression QP}

\begin{figure*}[!t]
   \centering
   \subfloat[QP=16]{
      \includegraphics[width=0.33\textwidth]{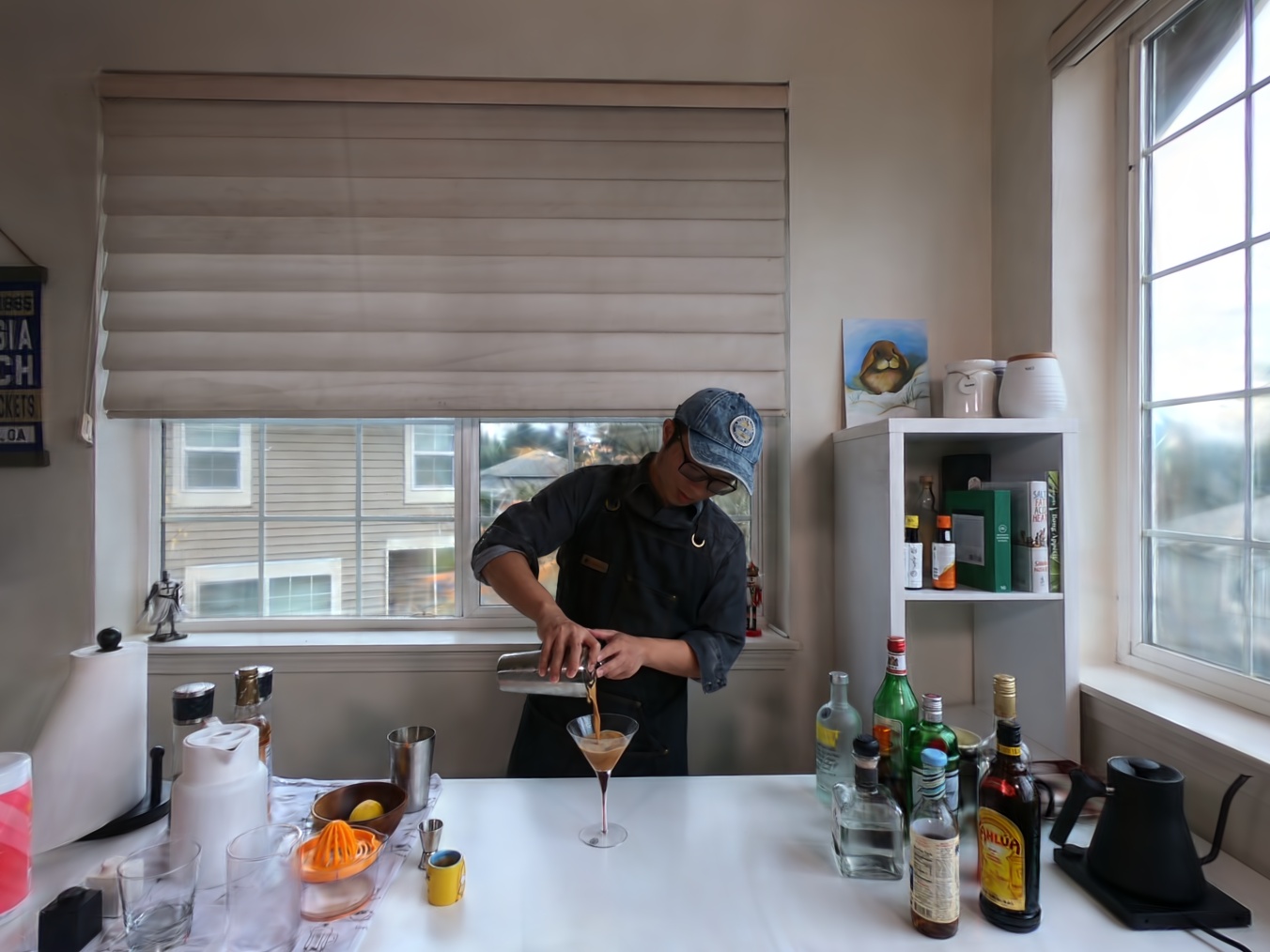}
   }
   \subfloat[QP=20]{
      \includegraphics[width=0.33\textwidth]{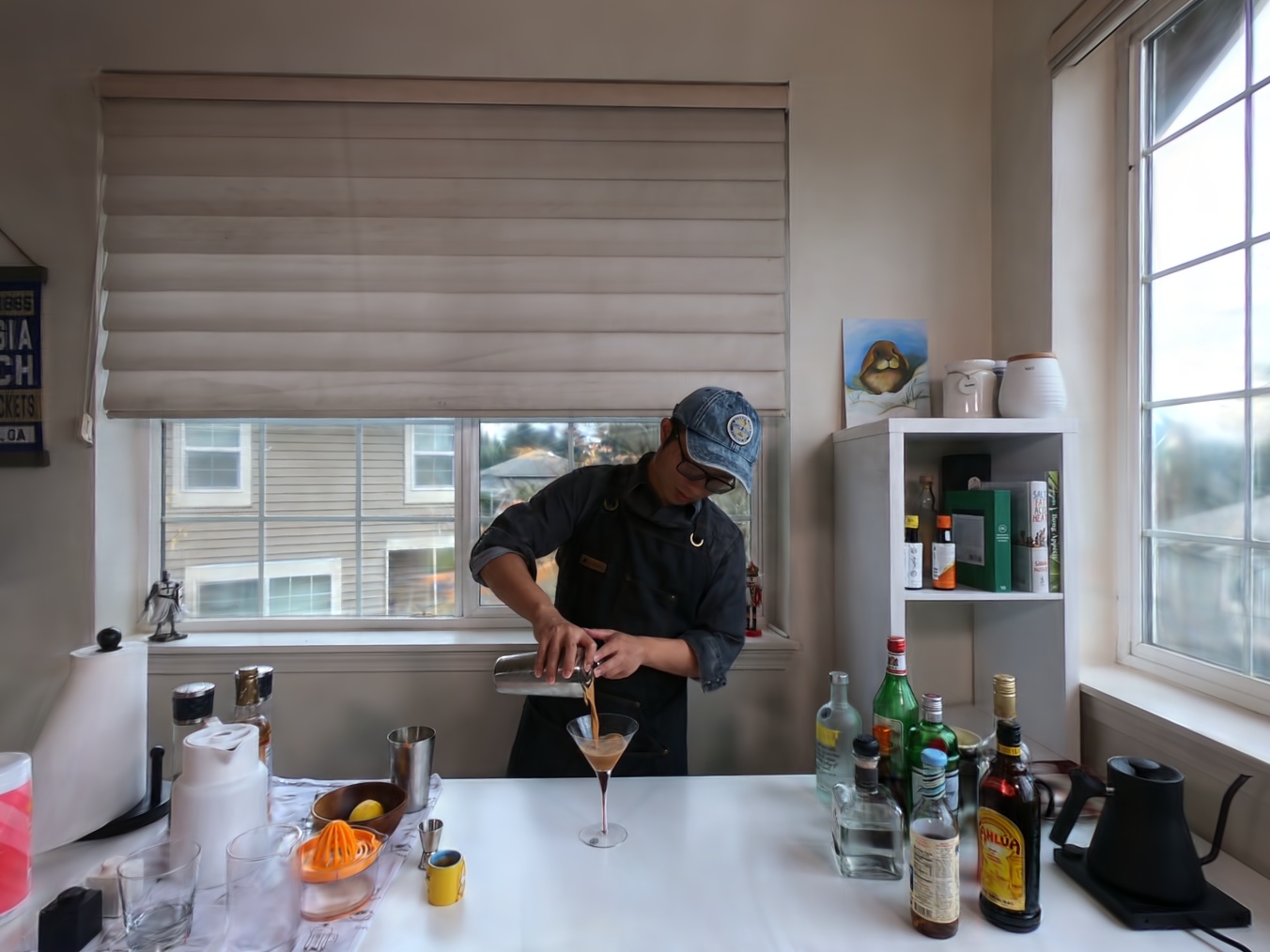}
   }
   \subfloat[QP=24]{
      \includegraphics[width=0.33\textwidth]{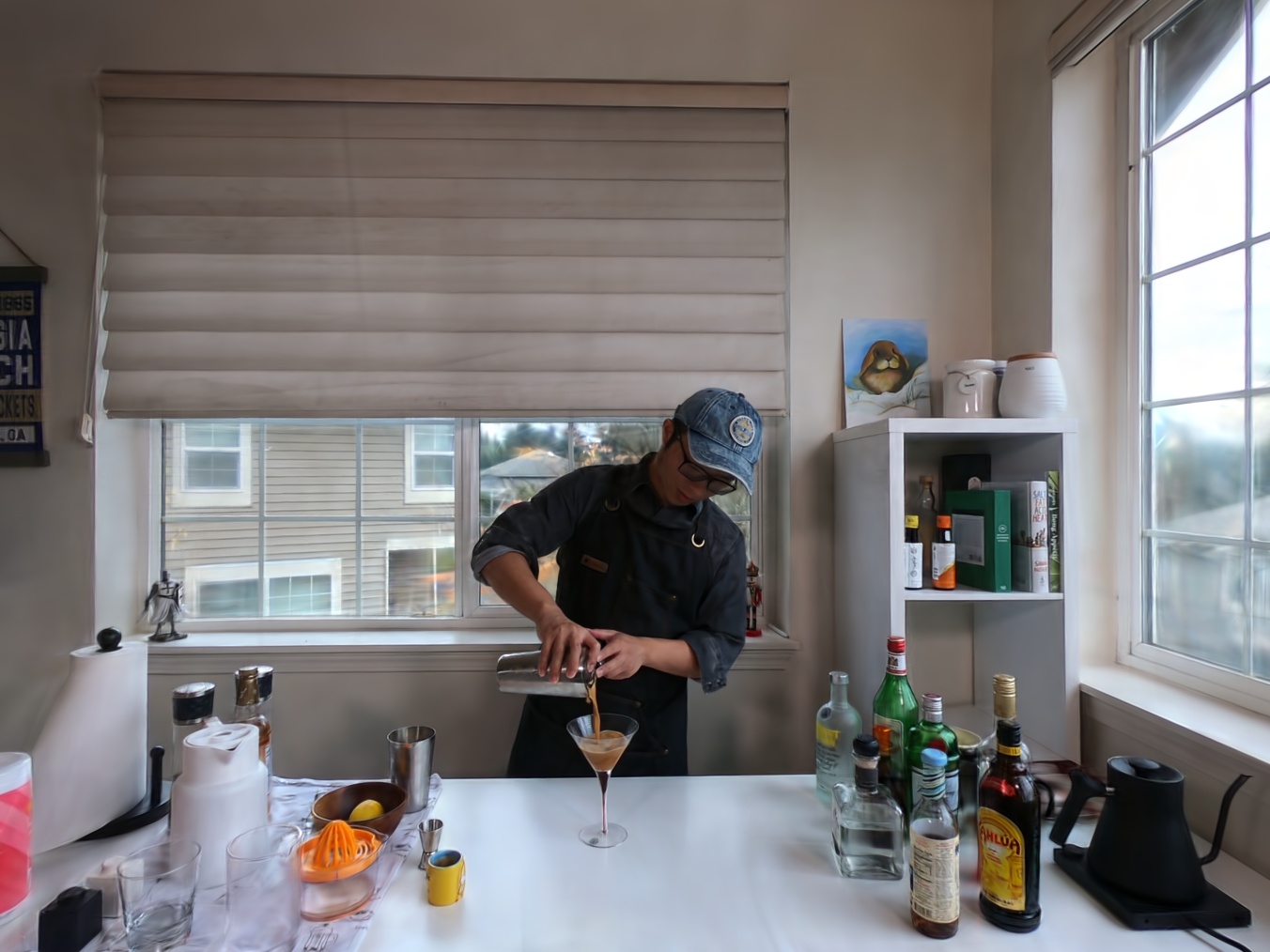}
   }
   \ 
   \subfloat[QP=28]{
      \includegraphics[width=0.33\textwidth]{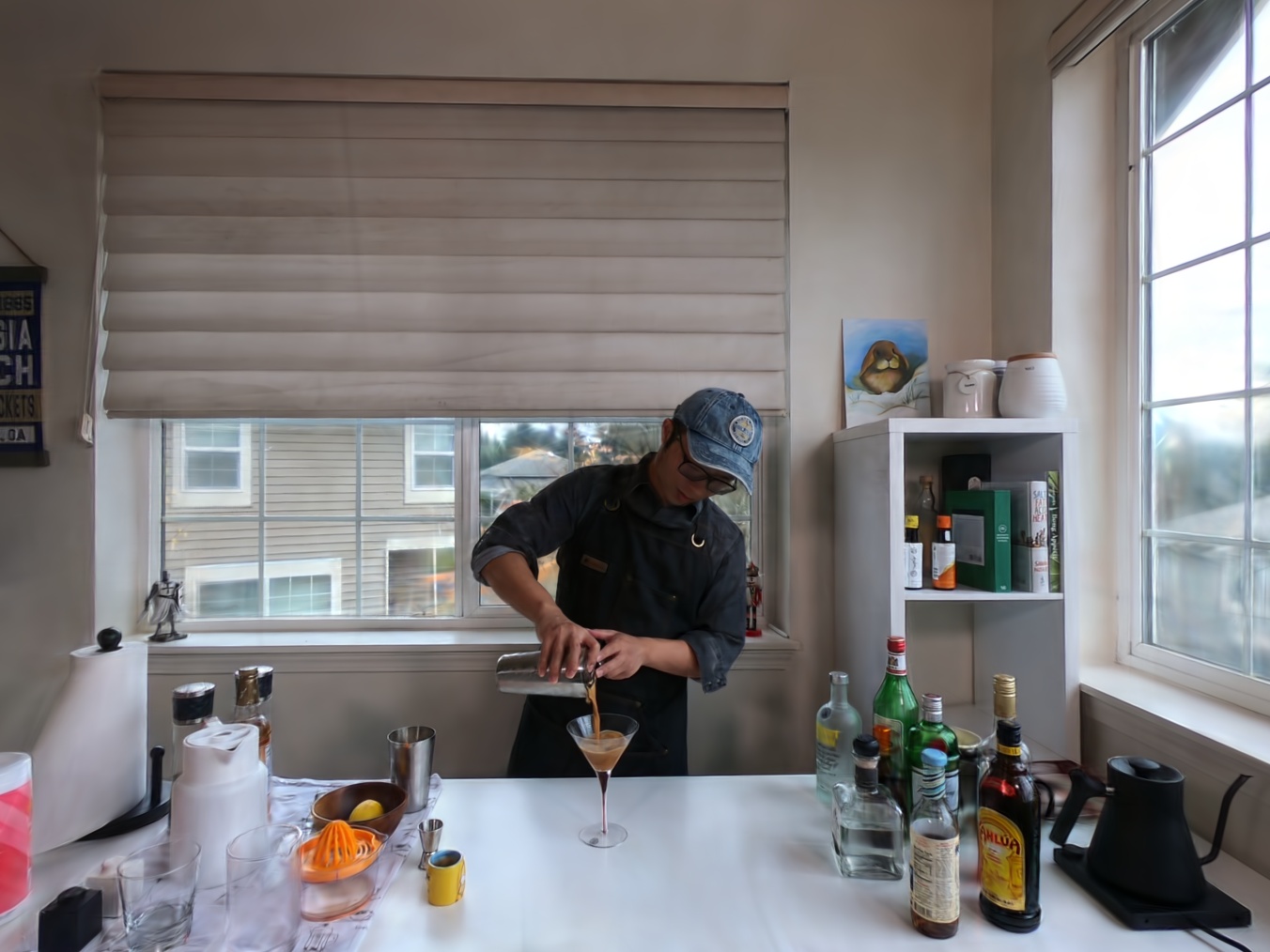}
   }
   \subfloat[QP=32]{
      \includegraphics[width=0.33\textwidth]{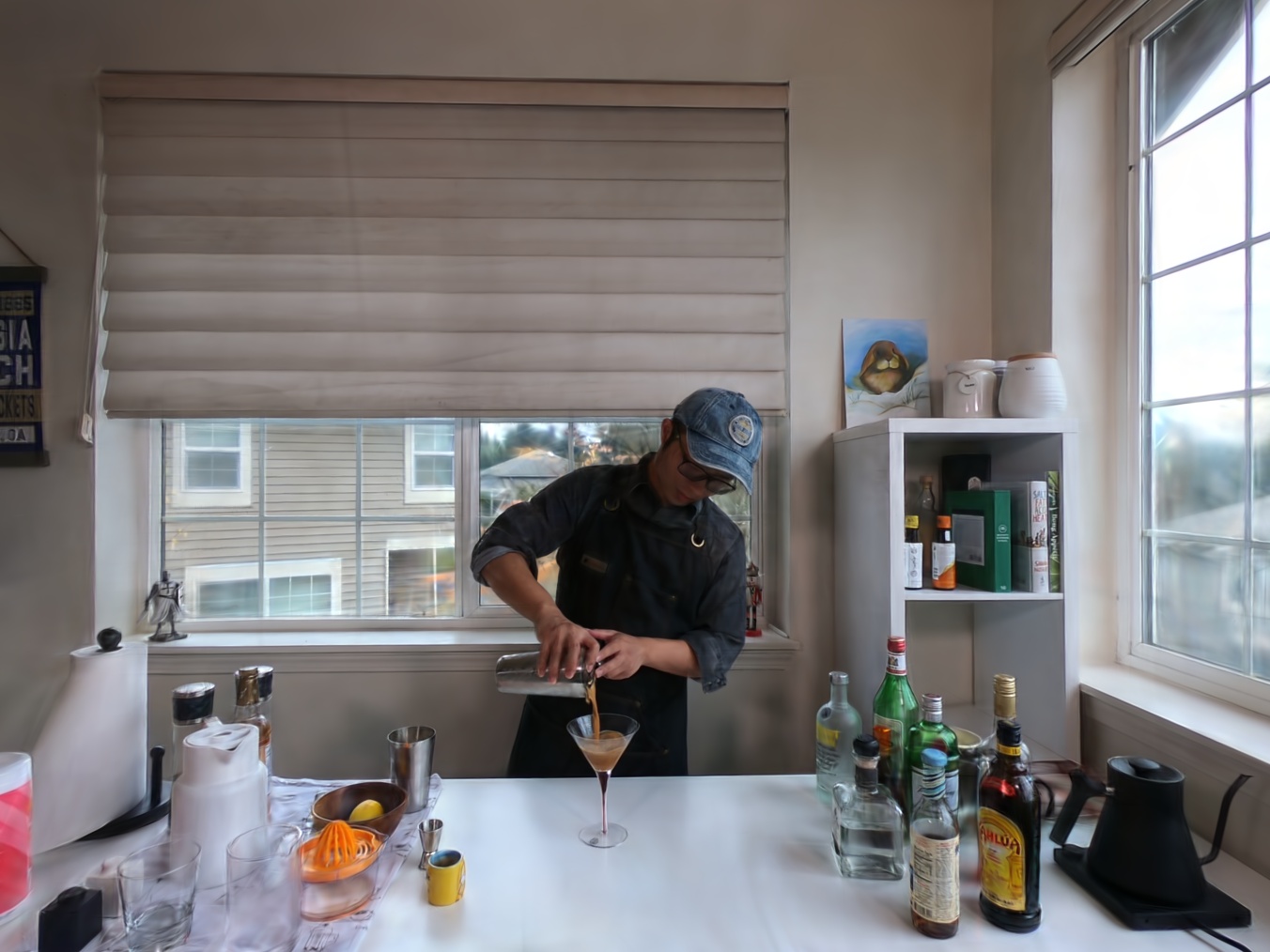}
   }
   \subfloat[GroudTruth]{
      \includegraphics[width=0.33\textwidth]{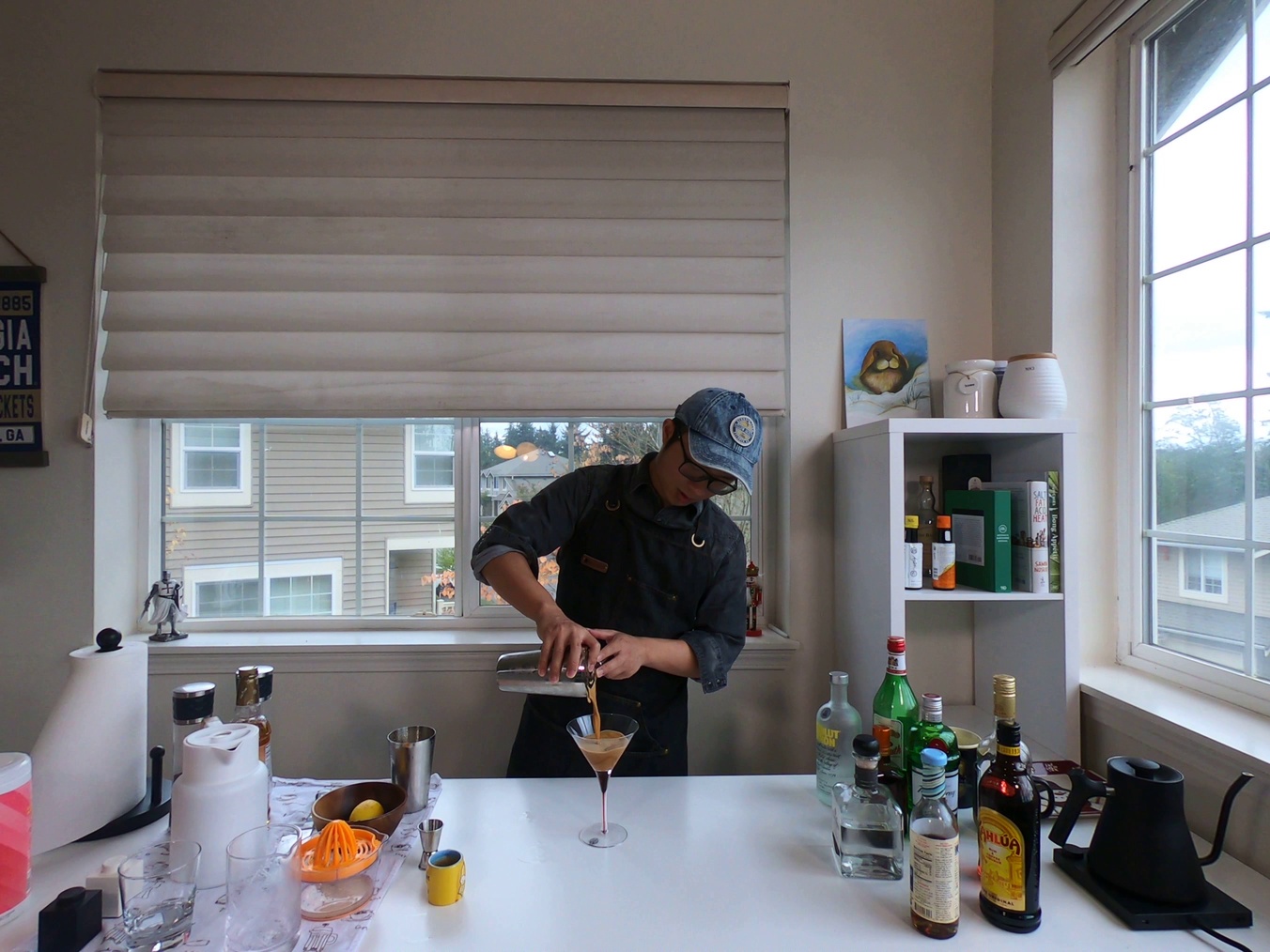}
   }
   \caption{Qualitative comparison under different compression qp on \textit{Coffee Martini} scene of N3DV dataset.}
   \label{fig:qp-coffee}
   \vspace{-0.2cm}
\end{figure*}

\begin{figure*}[!t]
   \centering
   \subfloat[QP=16]{
      \includegraphics[width=0.33\textwidth]{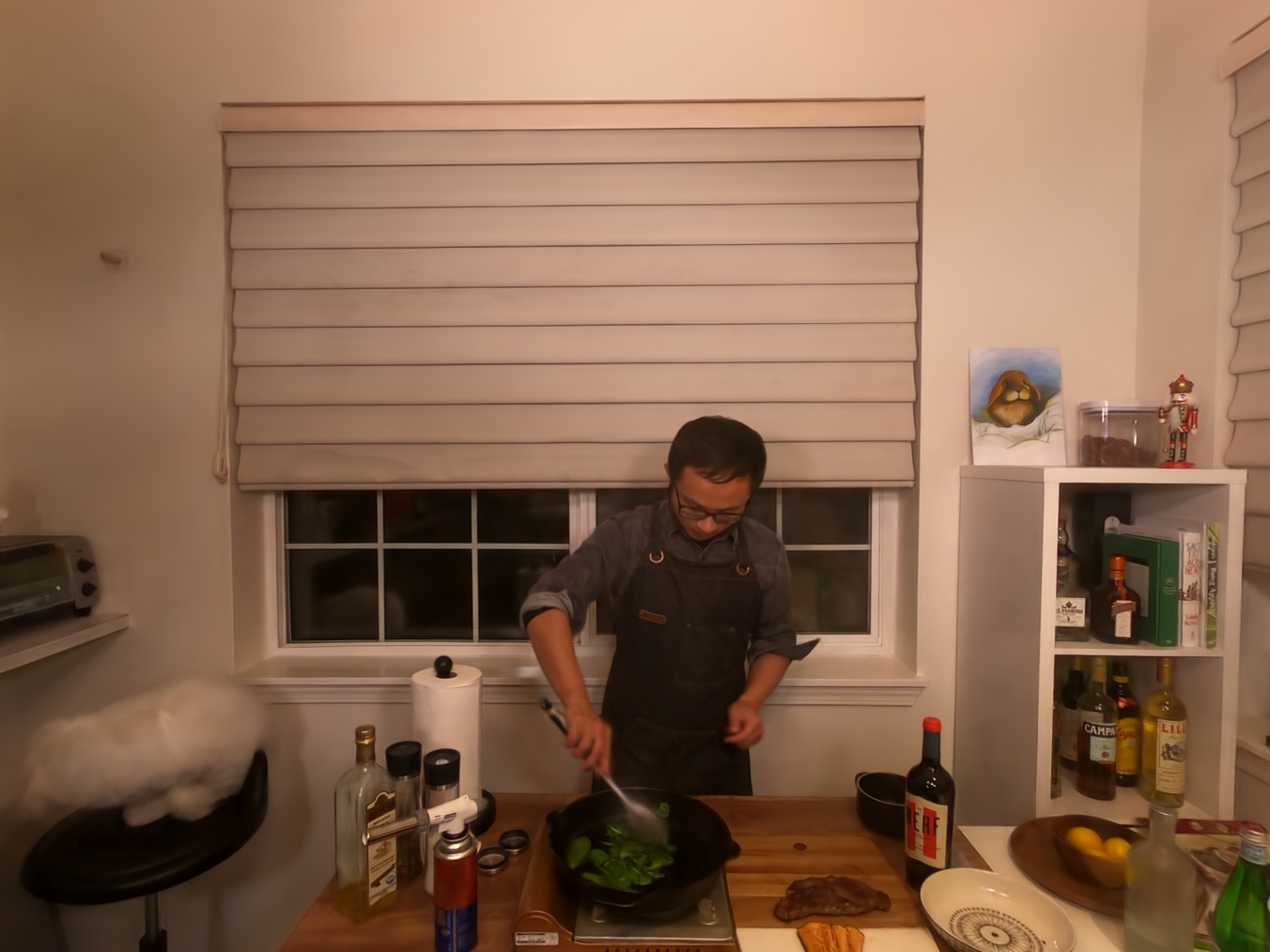}
   }
   \subfloat[QP=20]{
      \includegraphics[width=0.33\textwidth]{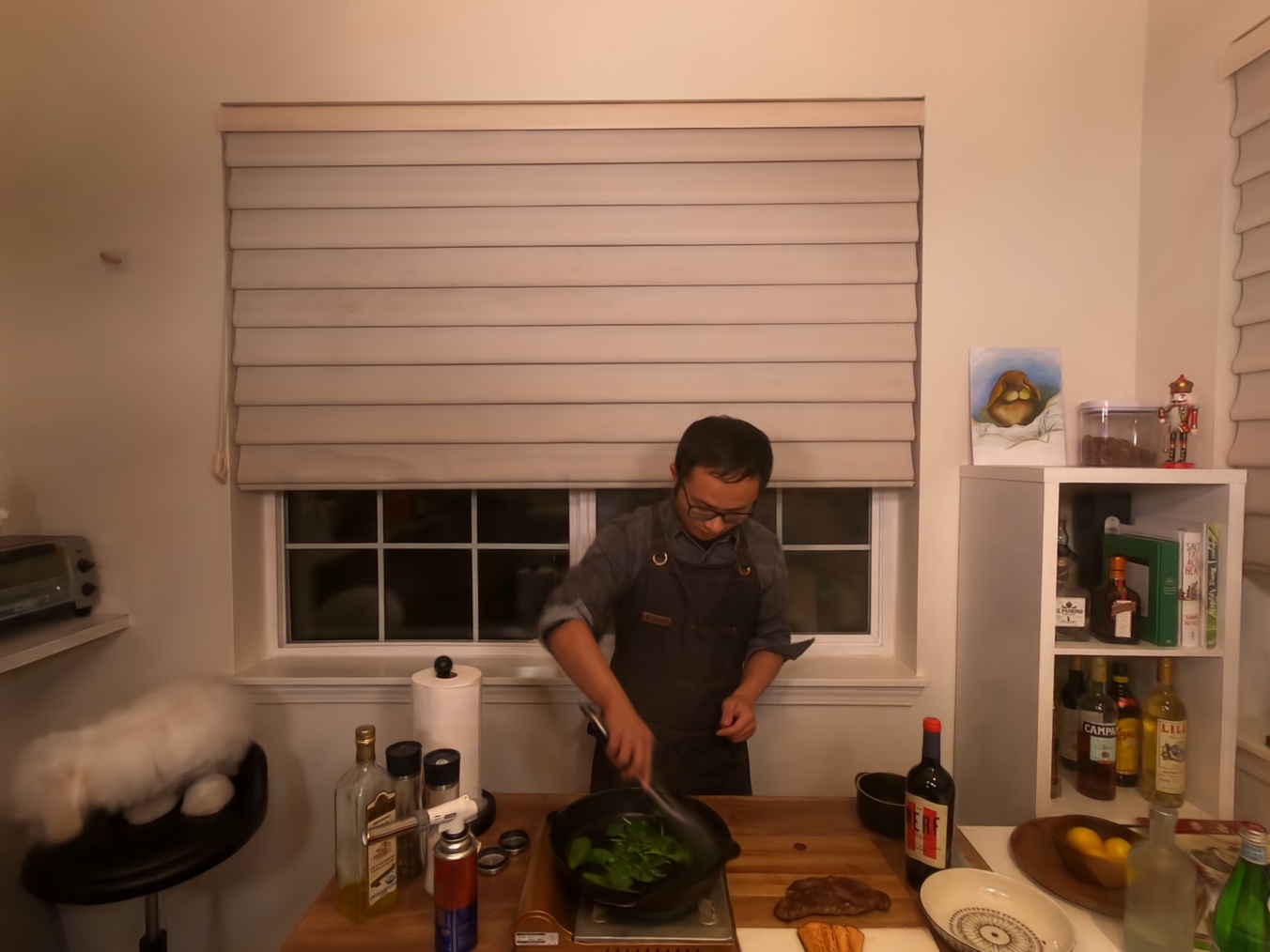}
   }
   \subfloat[QP=24]{
      \includegraphics[width=0.33\textwidth]{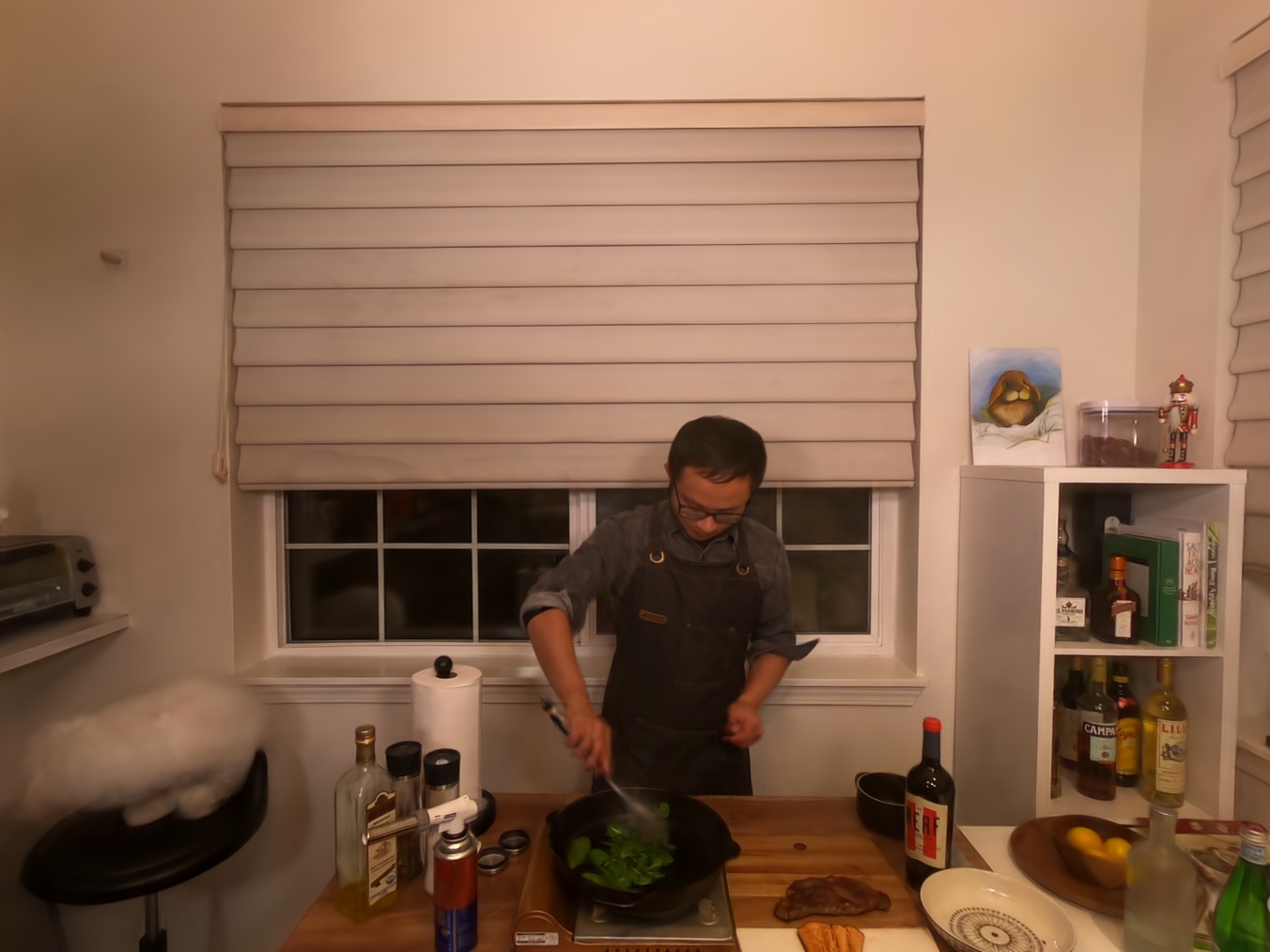}
   }
   \ 
   \subfloat[QP=28]{
      \includegraphics[width=0.33\textwidth]{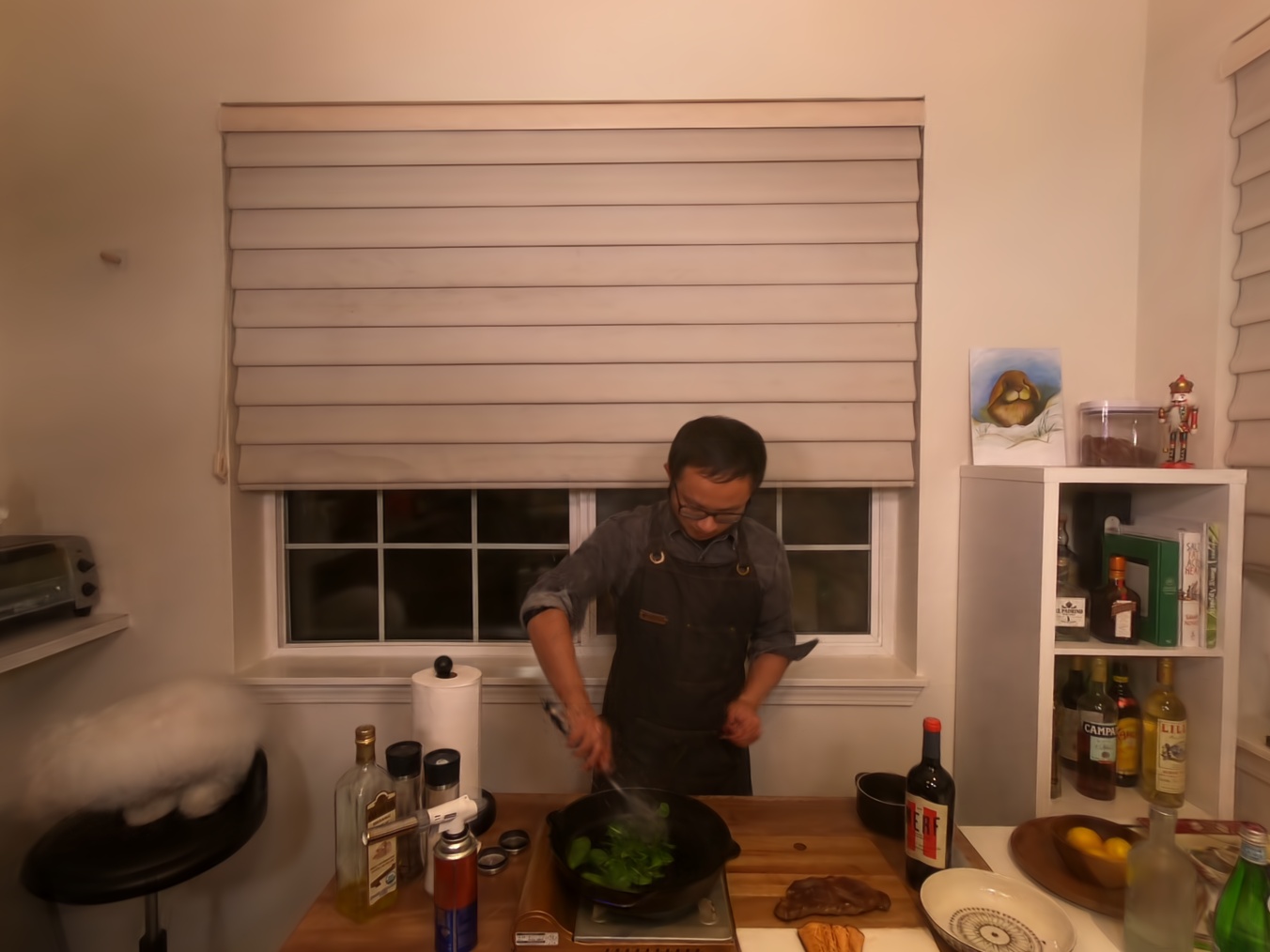}
   }
   \subfloat[QP=32]{
      \includegraphics[width=0.33\textwidth]{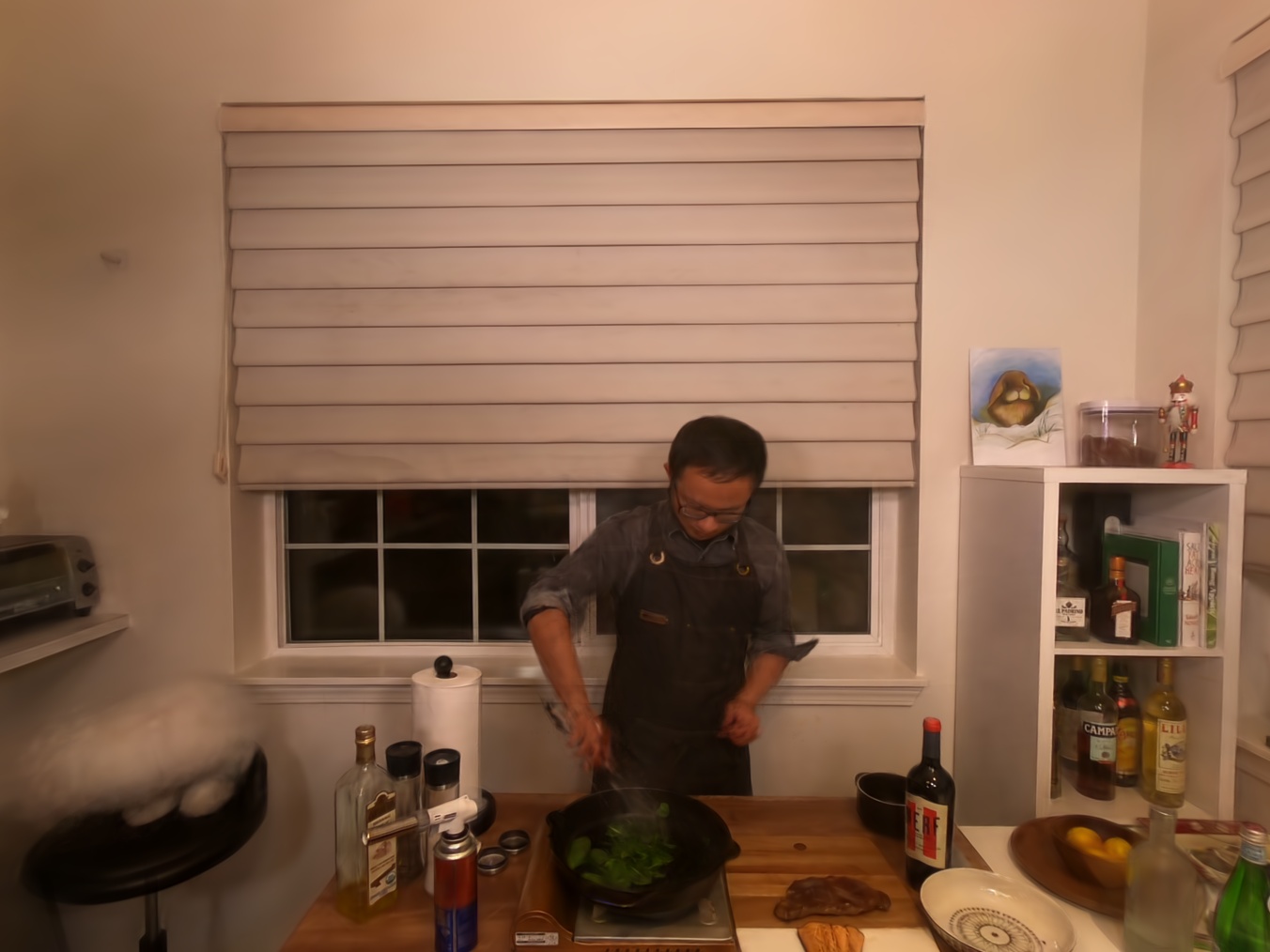}
   }
   \subfloat[GroudTruth]{
      \includegraphics[width=0.33\textwidth]{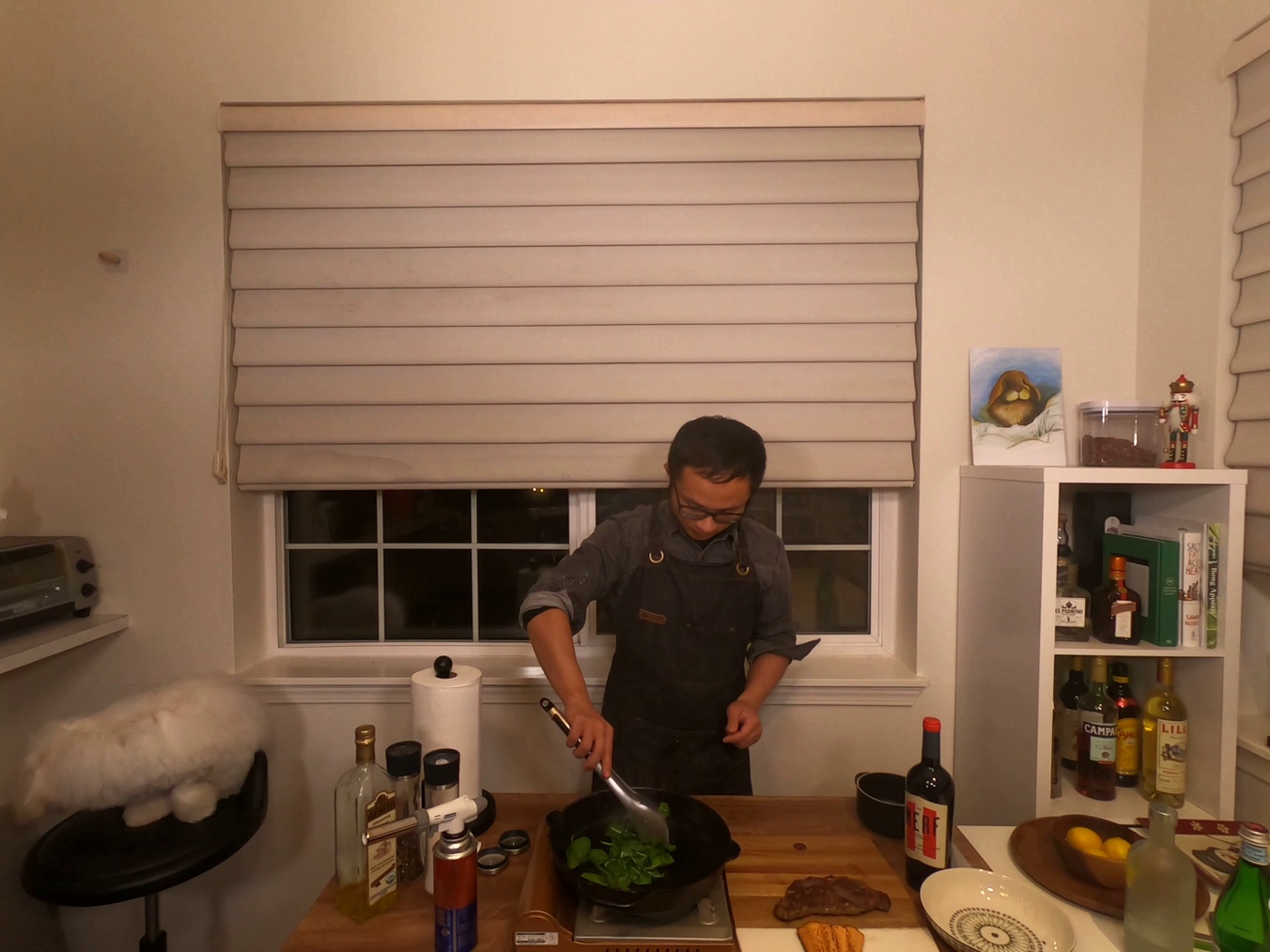}
   }
   \caption{Qualitative comparison under different compression qp on \textit{Cook Spinach} scene of N3DV dataset.}
   \label{fig:qp-cook}
   \vspace{-0.2cm}
\end{figure*}

\begin{figure*}[!t]
   \centering
   \subfloat[QP=16]{
      \includegraphics[width=0.33\textwidth]{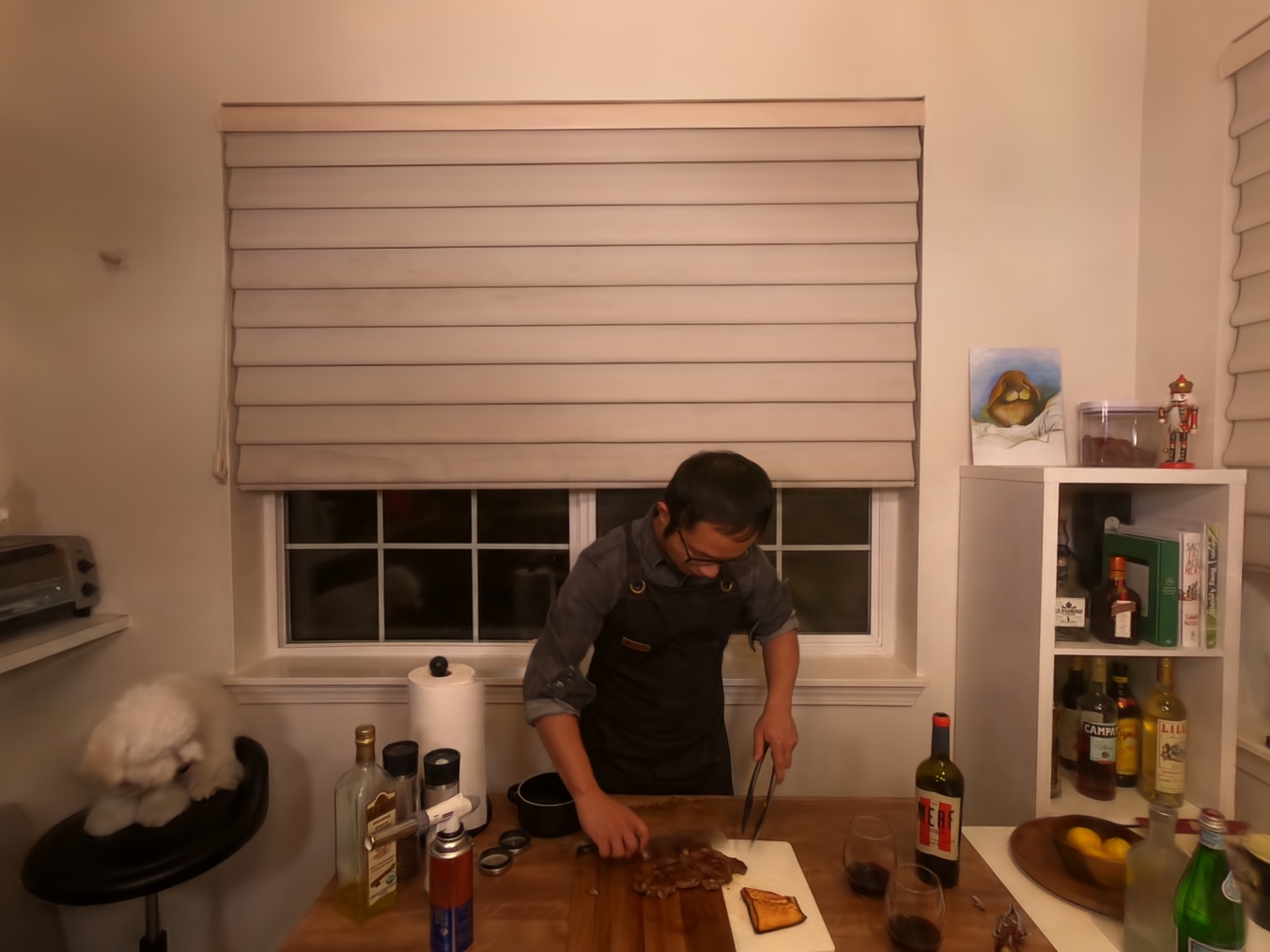}
   }
   \subfloat[QP=20]{
      \includegraphics[width=0.33\textwidth]{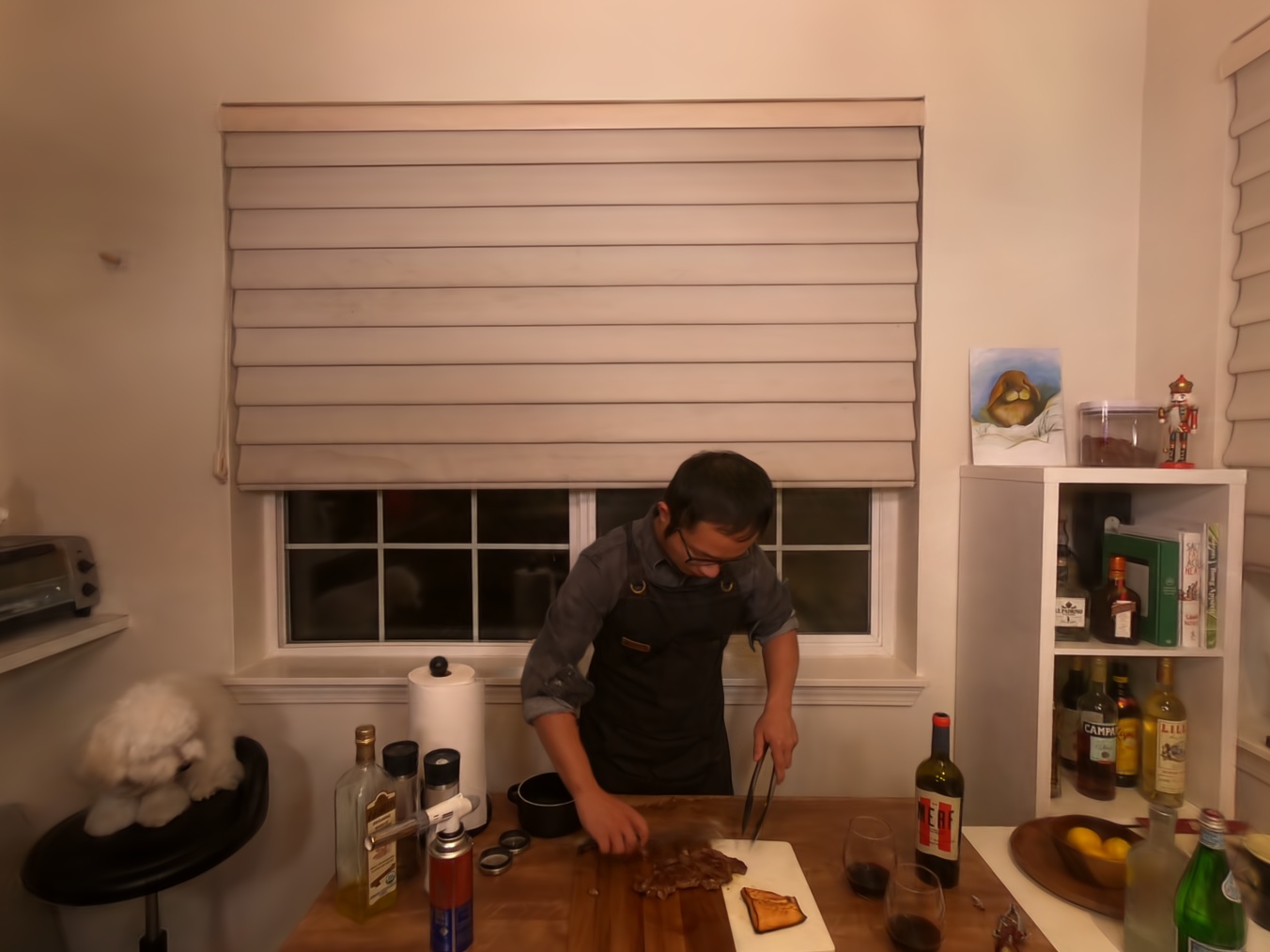}
   }
   \subfloat[QP=24]{
      \includegraphics[width=0.33\textwidth]{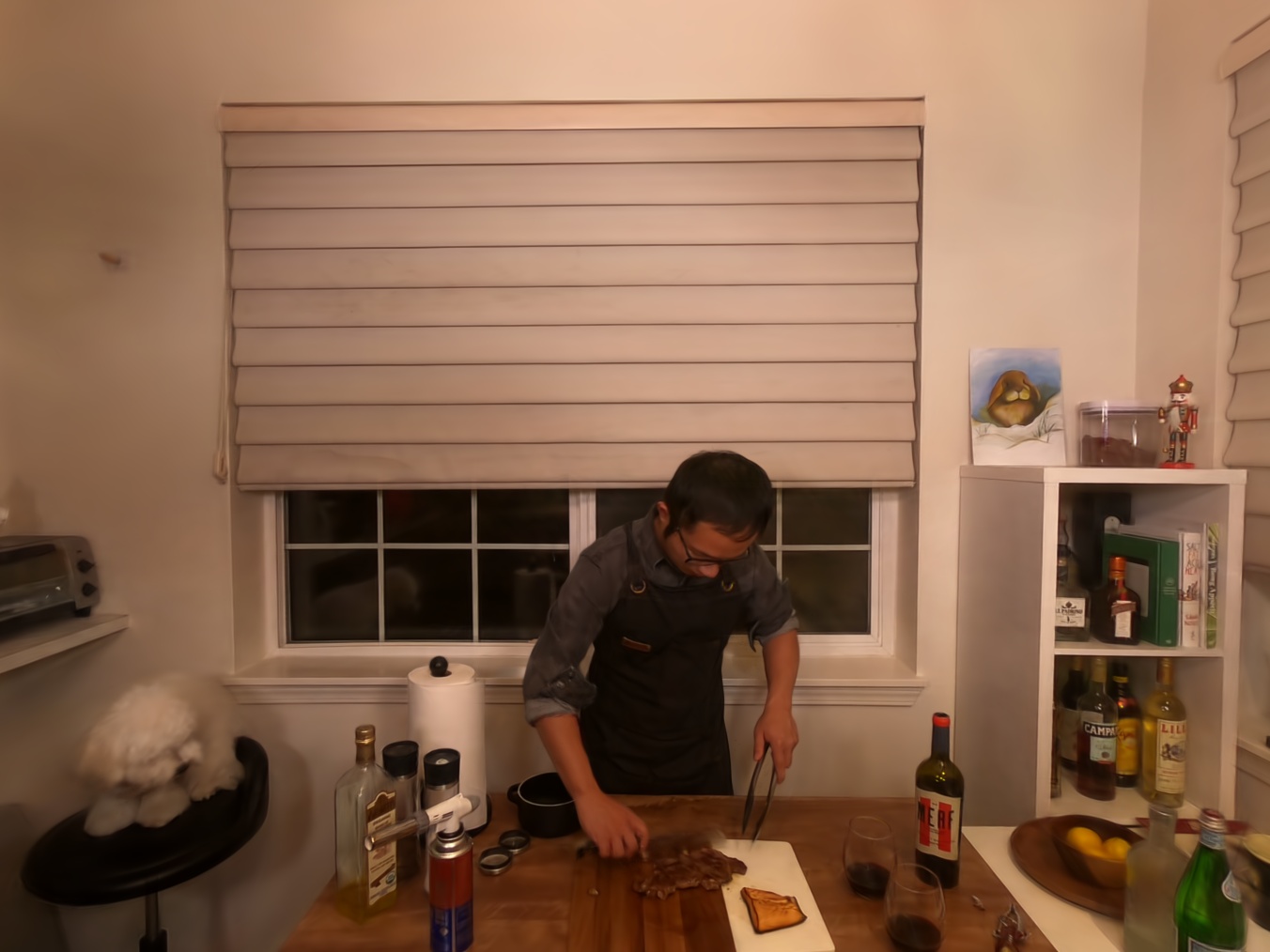}
   }
   \ 
   \subfloat[QP=28]{
      \includegraphics[width=0.33\textwidth]{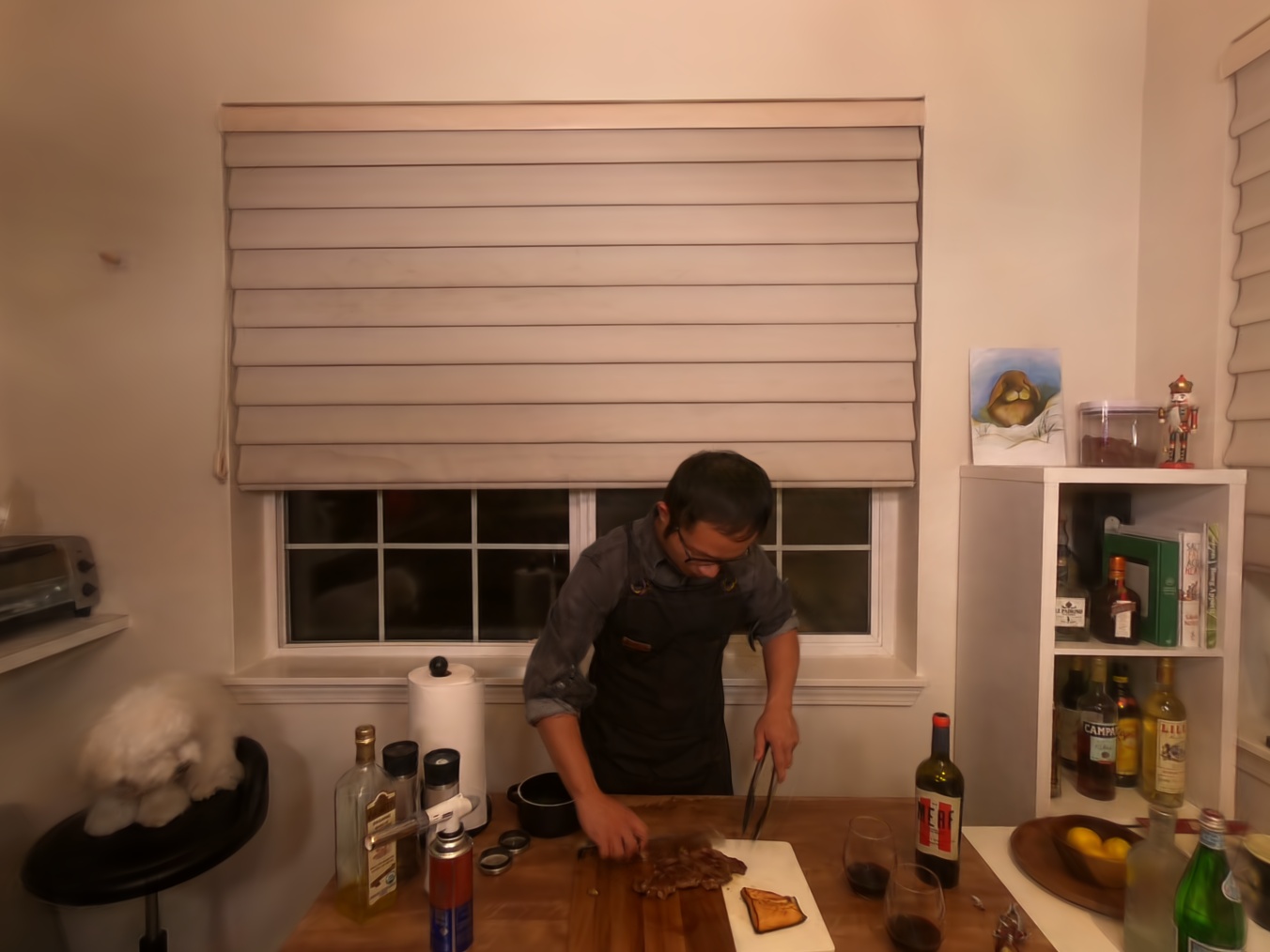}
   }
   \subfloat[QP=32]{
      \includegraphics[width=0.33\textwidth]{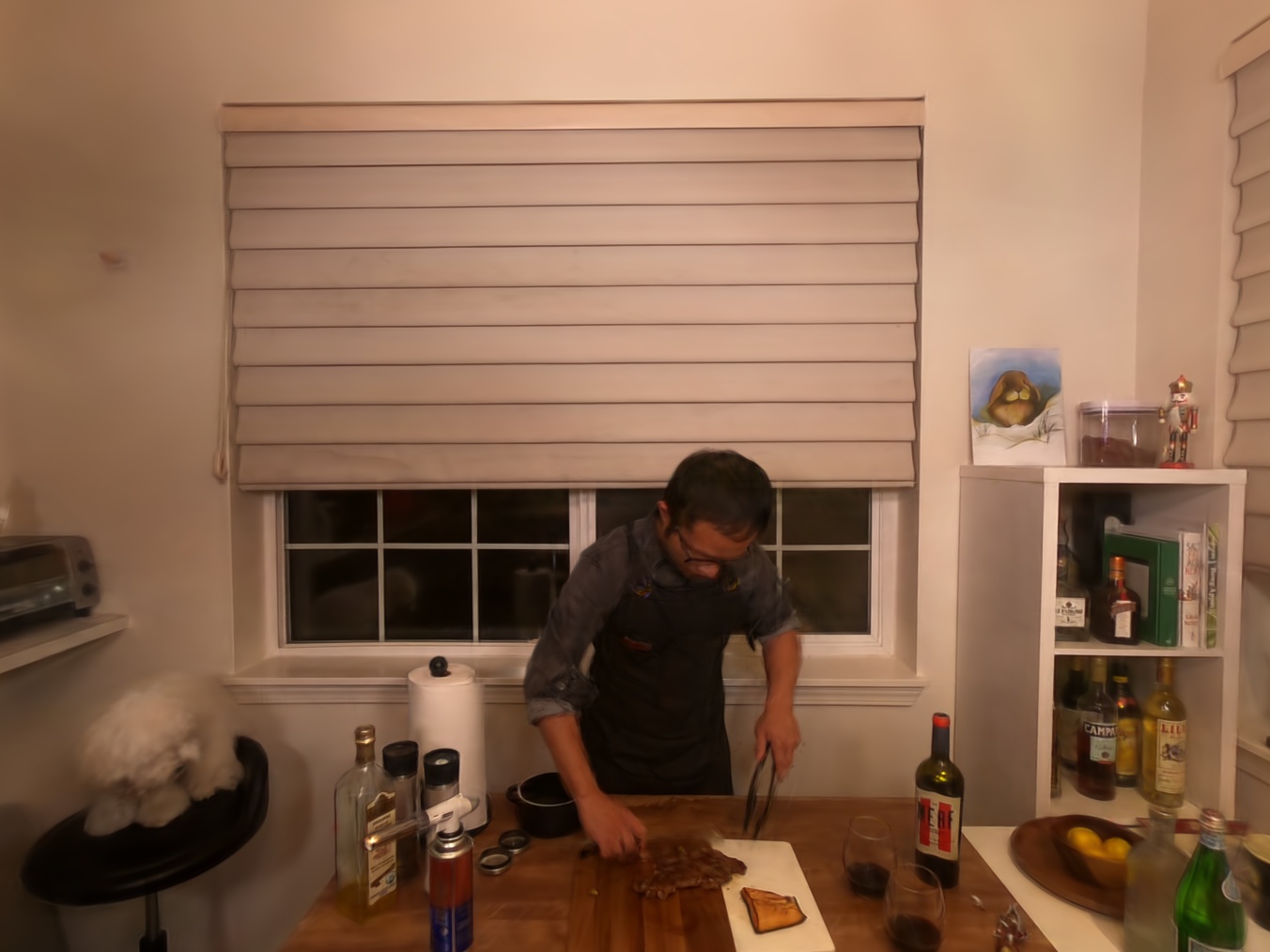}
   }
   \subfloat[GroudTruth]{
      \includegraphics[width=0.33\textwidth]{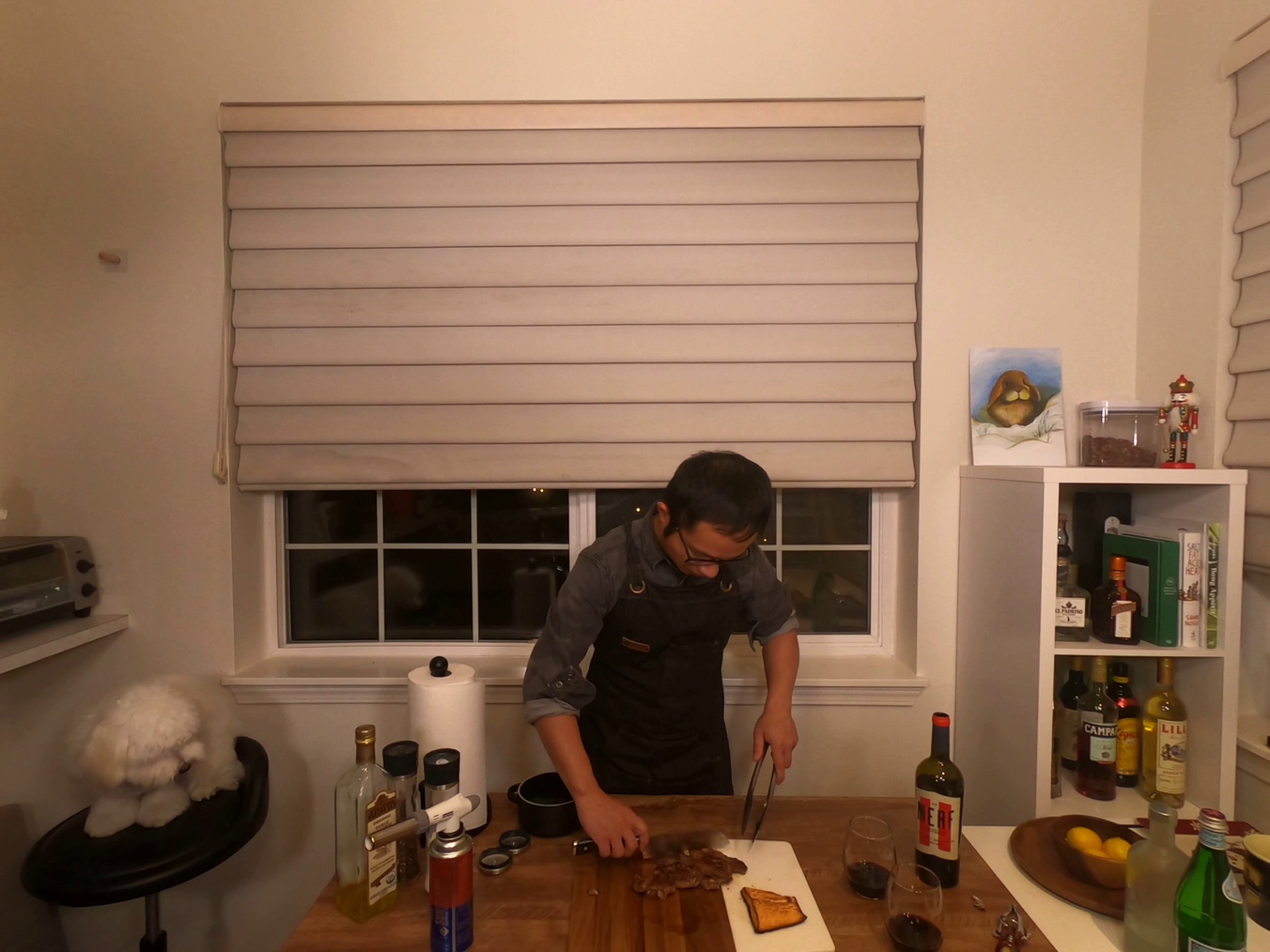}
   }
   \caption{Qualitative comparison under different compression qp on \textit{Cut Roasted Beef} scene of N3DV dataset.}
   \label{fig:qp-cut}
   \vspace{-0.2cm}
\end{figure*}

\begin{figure*}[!t]
   \centering
   \subfloat[QP=16]{
      \includegraphics[width=0.33\textwidth]{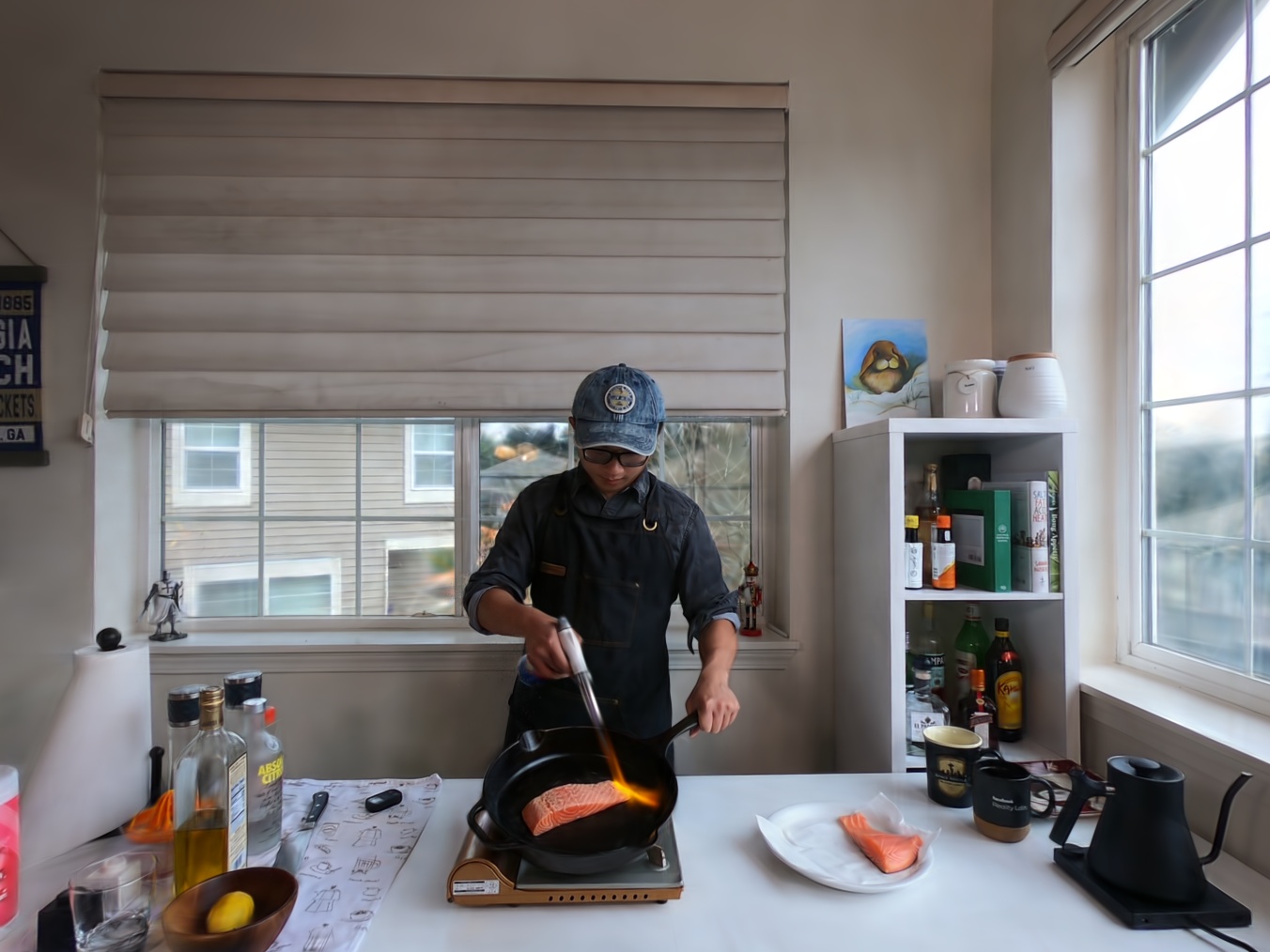}
   }
   \subfloat[QP=20]{
      \includegraphics[width=0.33\textwidth]{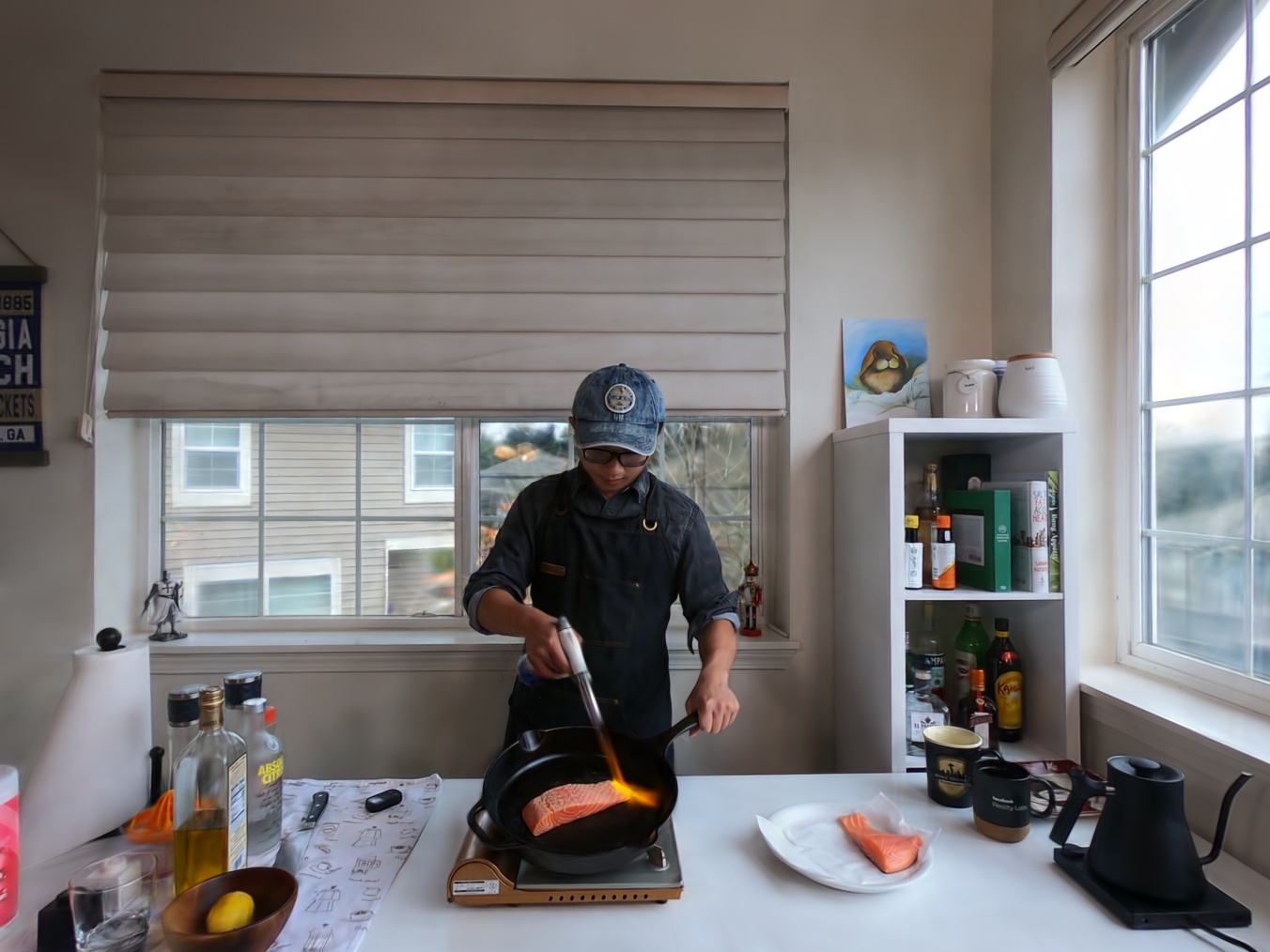}
   }
   \subfloat[QP=24]{
      \includegraphics[width=0.33\textwidth]{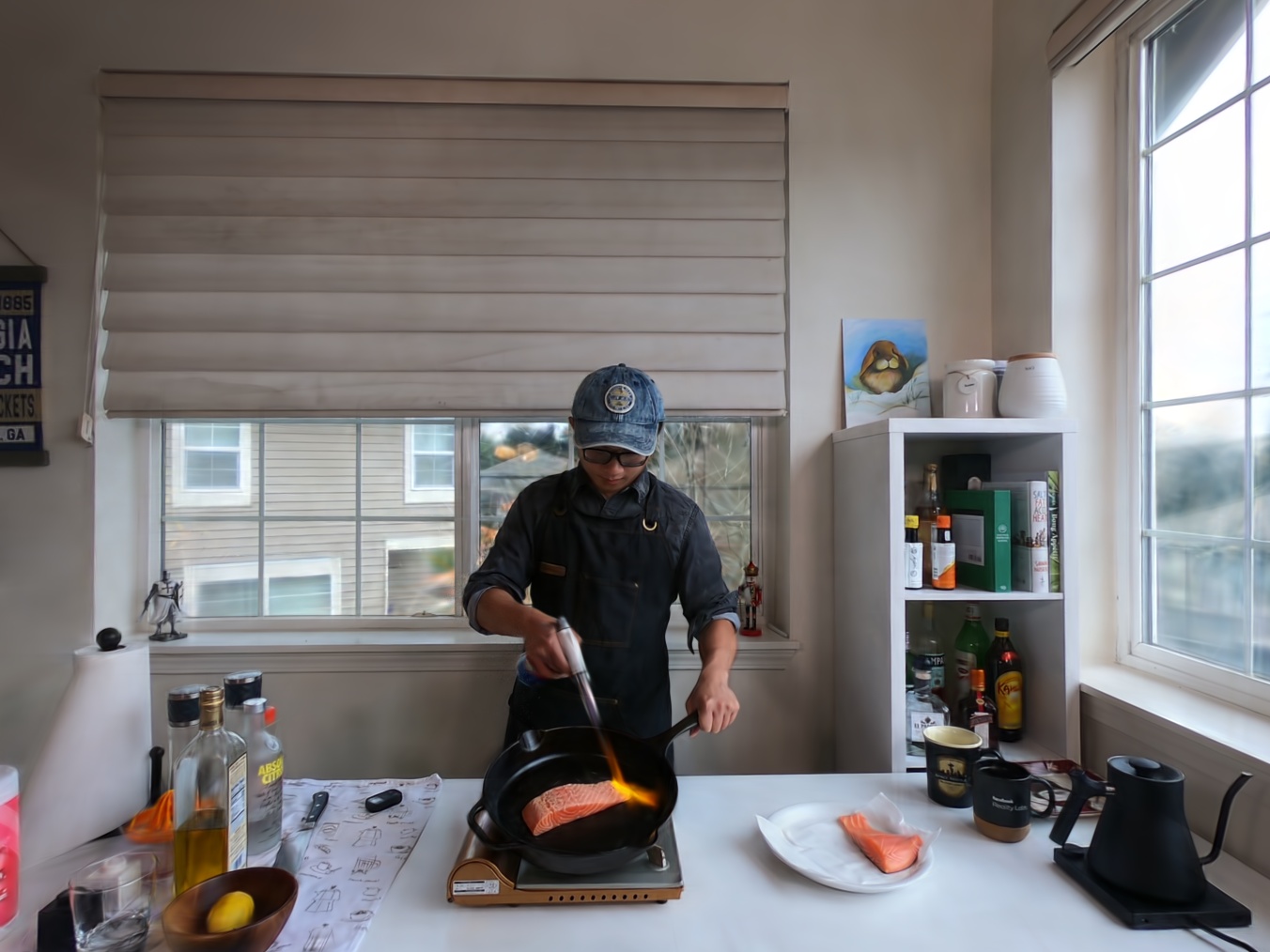}
   }
   \ 
   \subfloat[QP=28]{
      \includegraphics[width=0.33\textwidth]{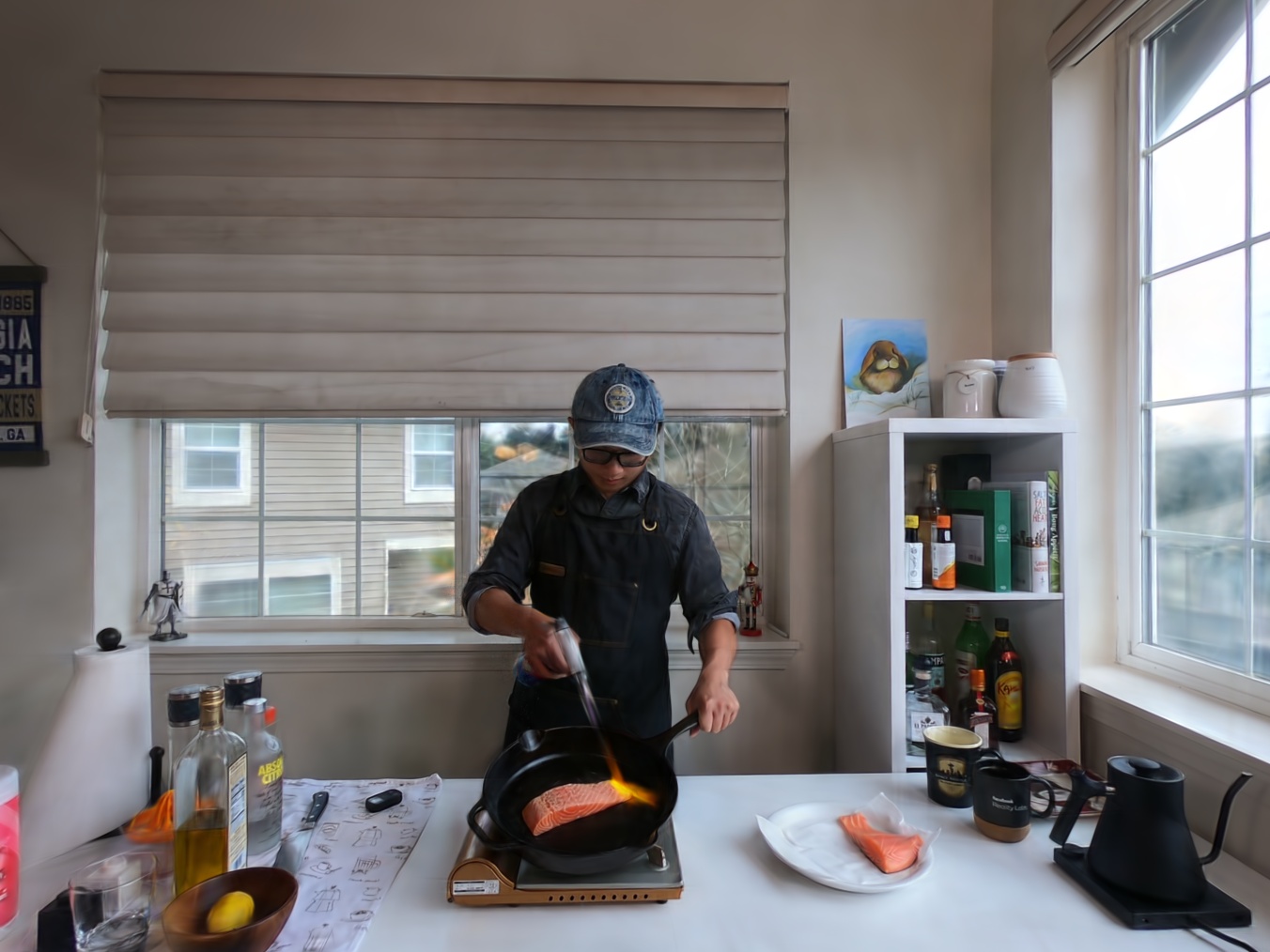}
   }
   \subfloat[QP=32]{
      \includegraphics[width=0.33\textwidth]{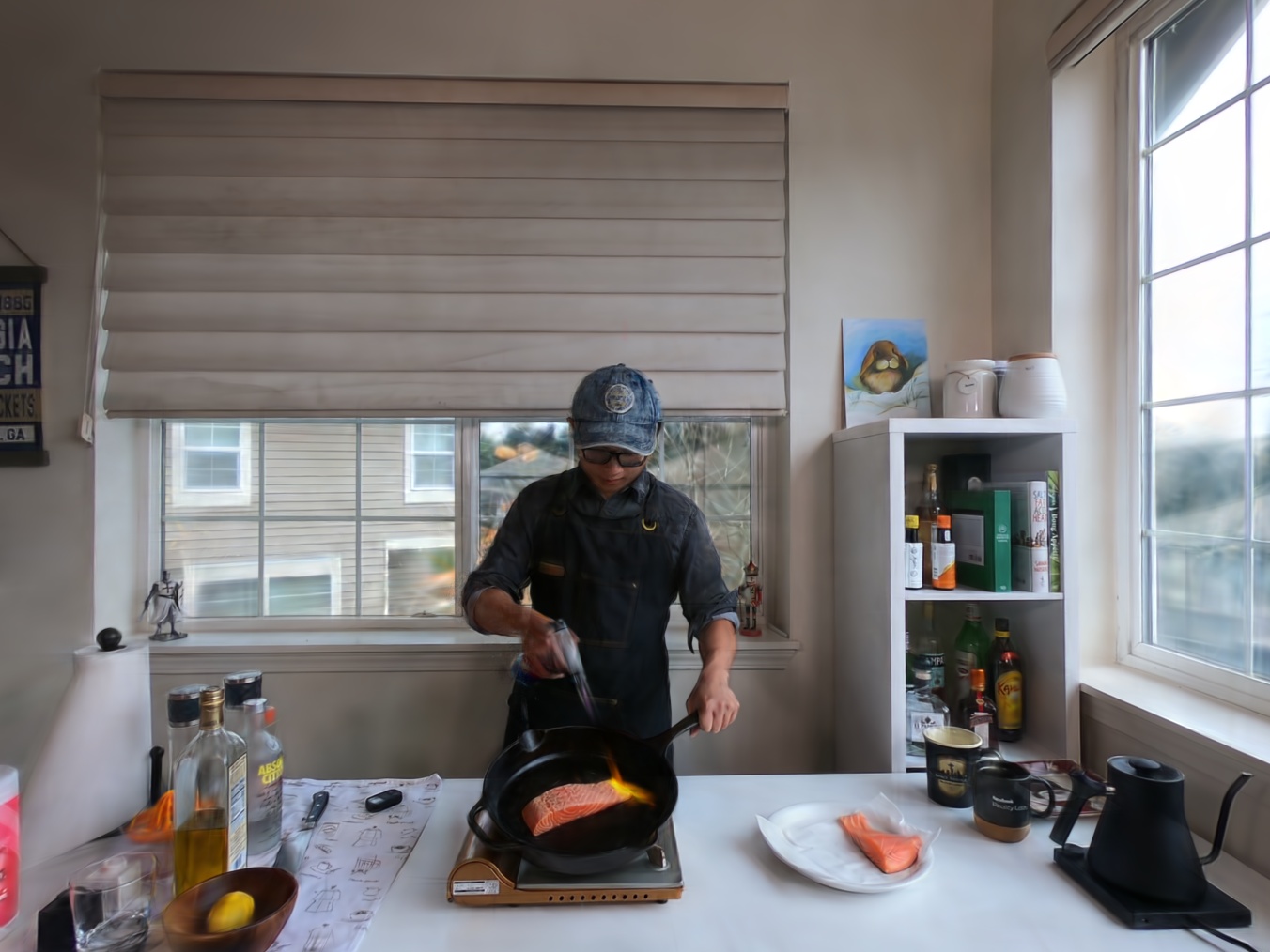}
   }
   \subfloat[GroudTruth]{
      \includegraphics[width=0.33\textwidth]{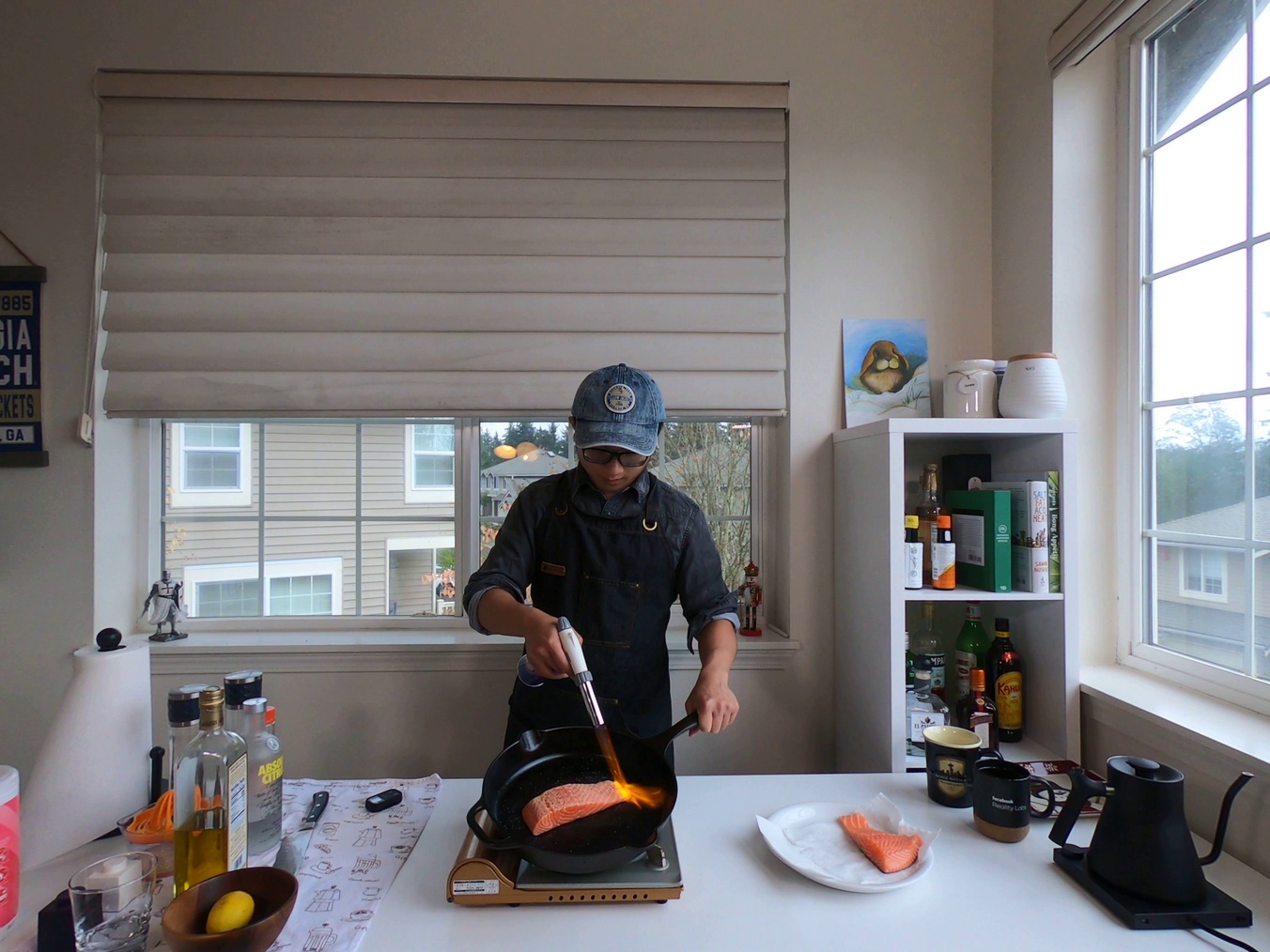}
   }
   \caption{Qualitative comparison under different compression qp on \textit{Flame Salmon} scene of N3DV dataset.}
   \label{fig:qp-salmon}
   \vspace{-0.2cm}
\end{figure*}

\begin{figure*}[!t]
   \centering
   \subfloat[QP=16]{
      \includegraphics[width=0.33\textwidth]{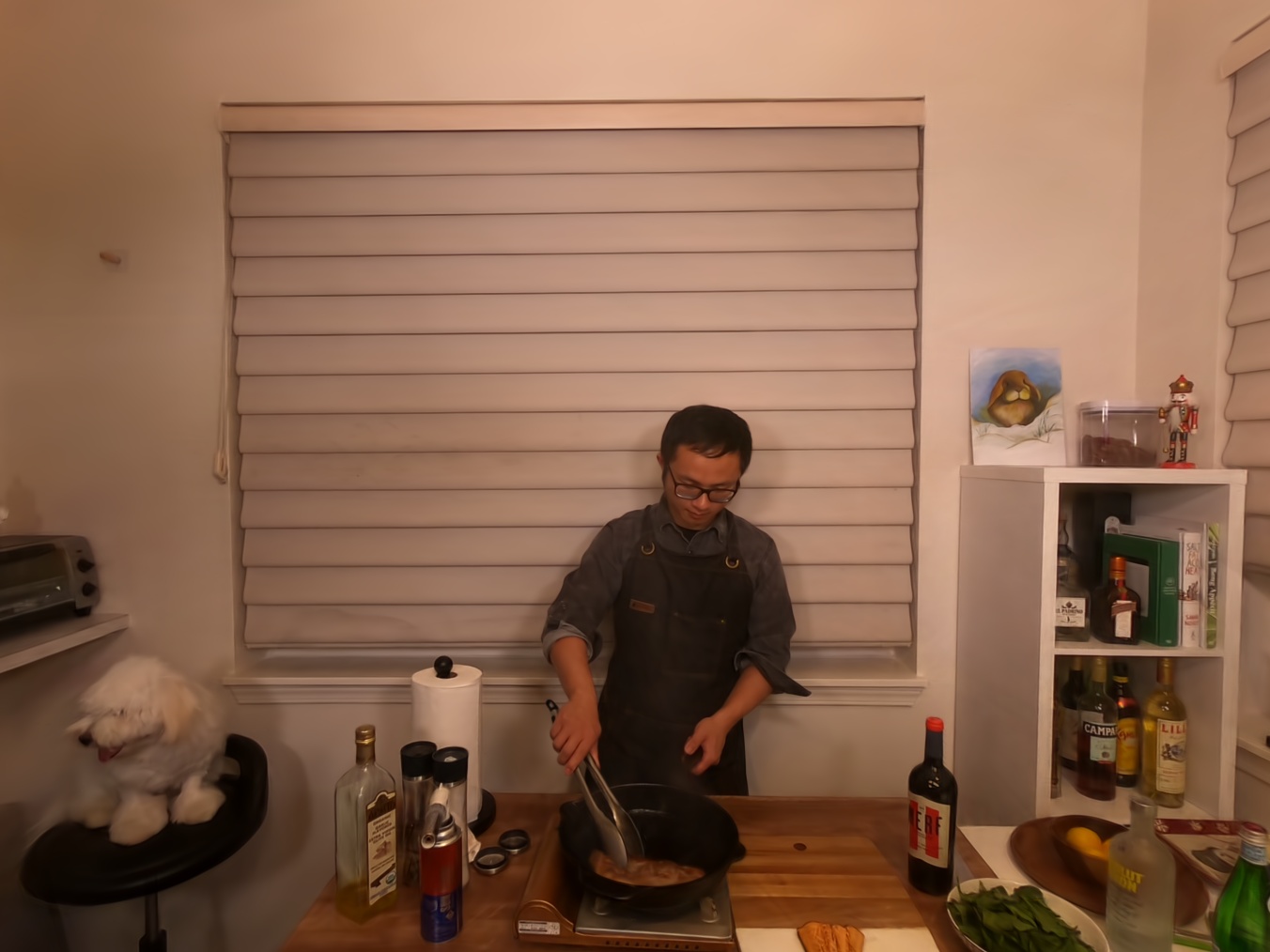}
   }
   \subfloat[QP=20]{
      \includegraphics[width=0.33\textwidth]{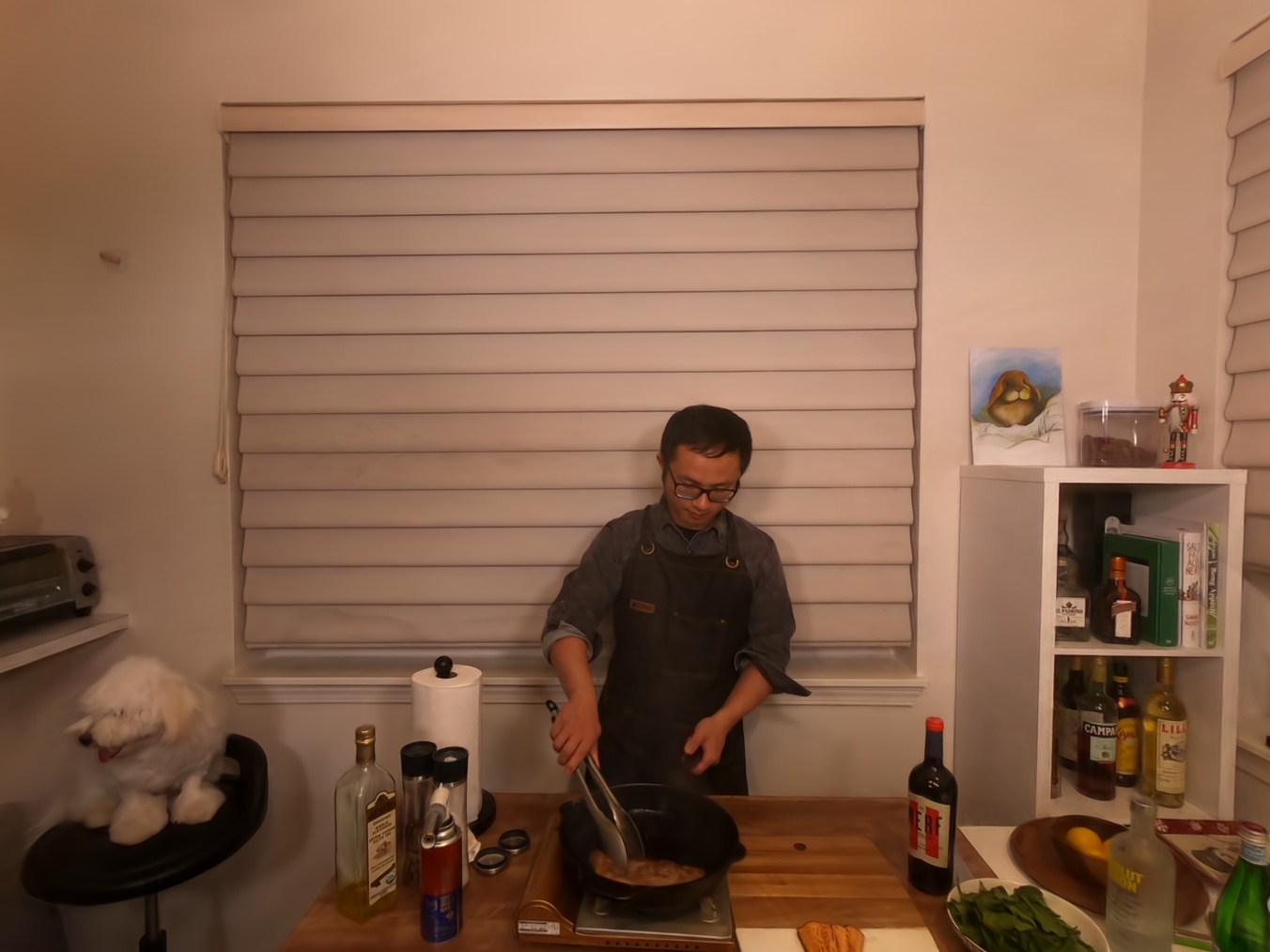}
   }
   \subfloat[QP=24]{
      \includegraphics[width=0.33\textwidth]{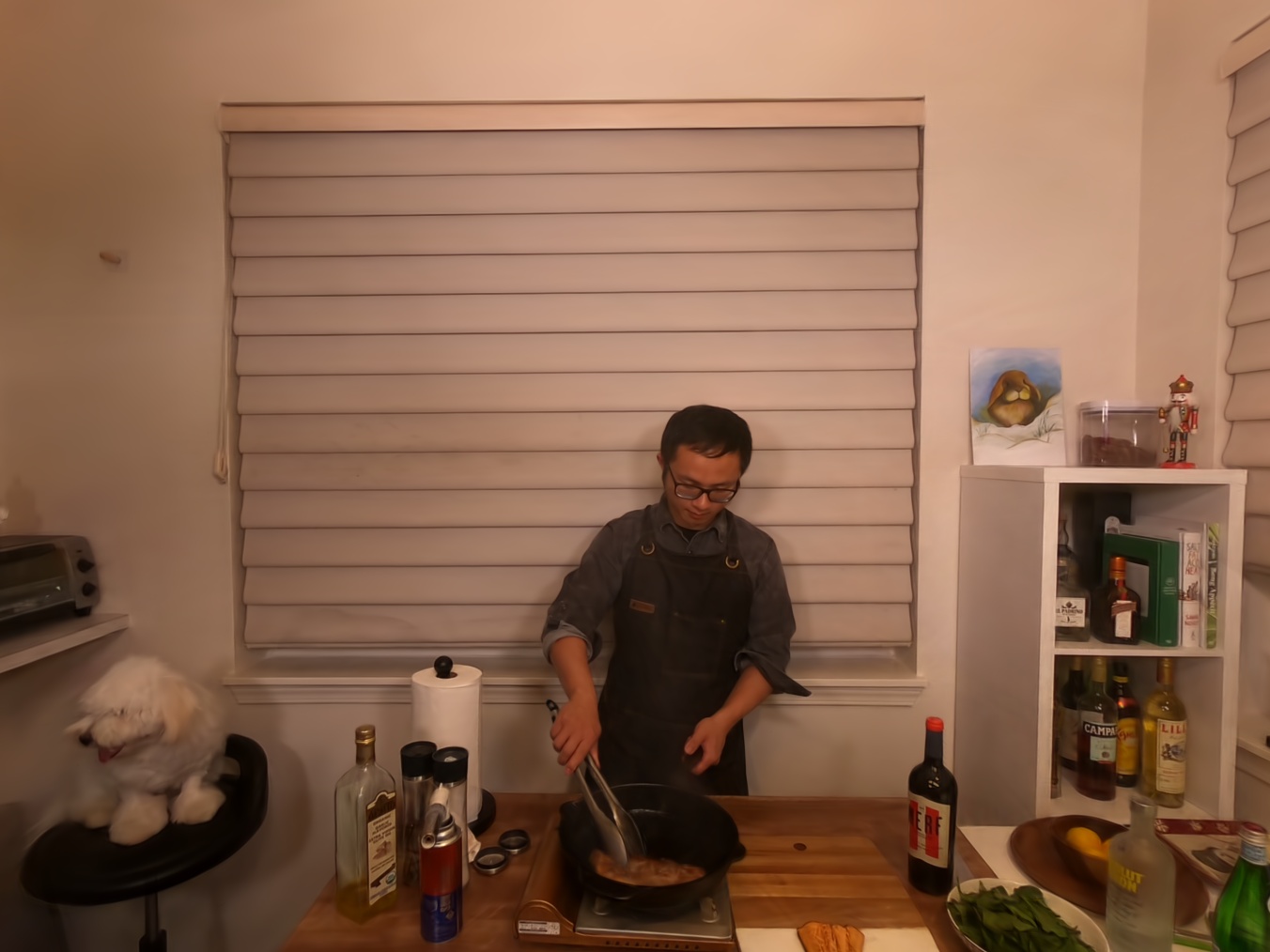}
   }
   \ 
   \subfloat[QP=28]{
      \includegraphics[width=0.33\textwidth]{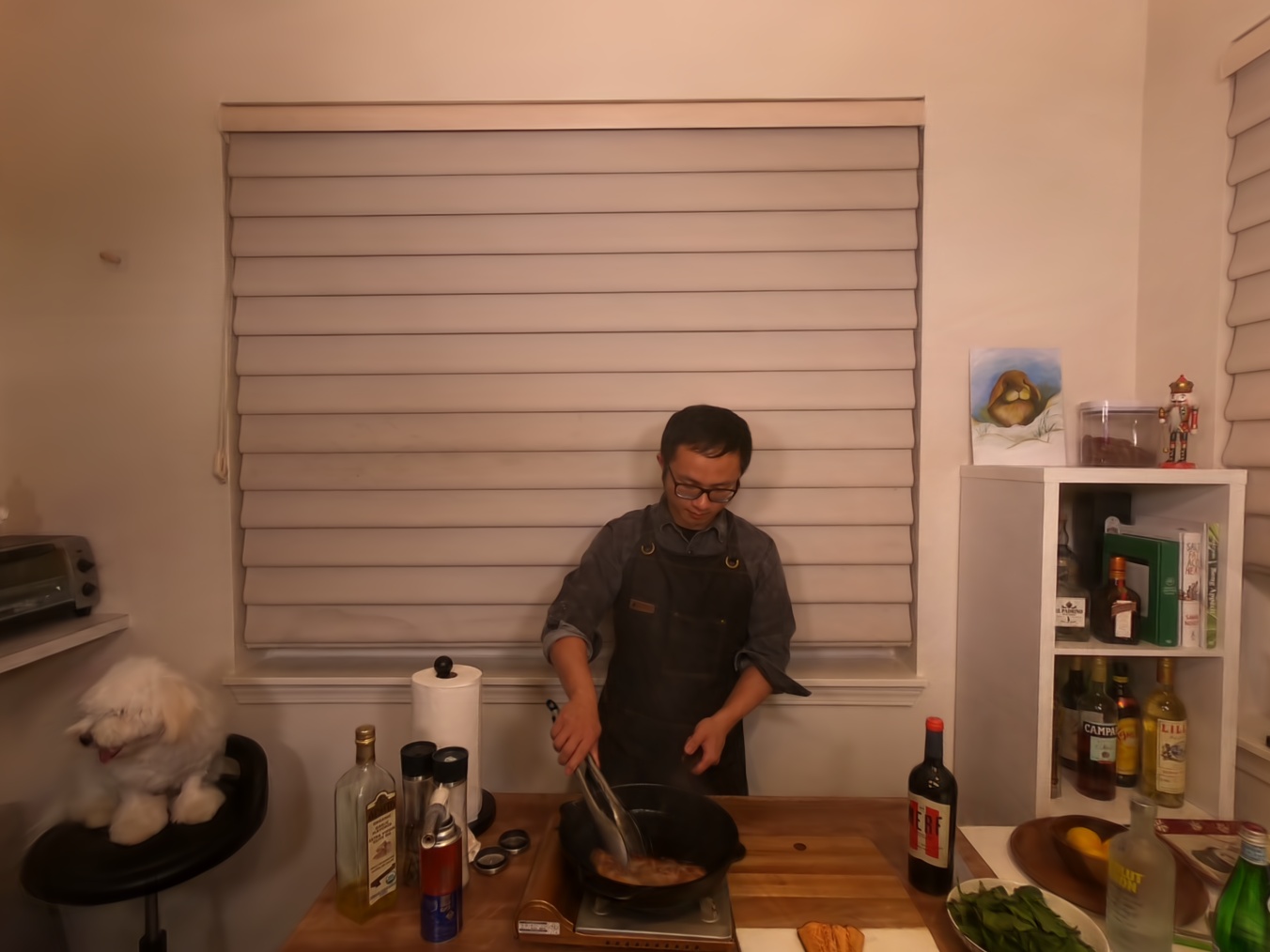}
   }
   \subfloat[QP=32]{
      \includegraphics[width=0.33\textwidth]{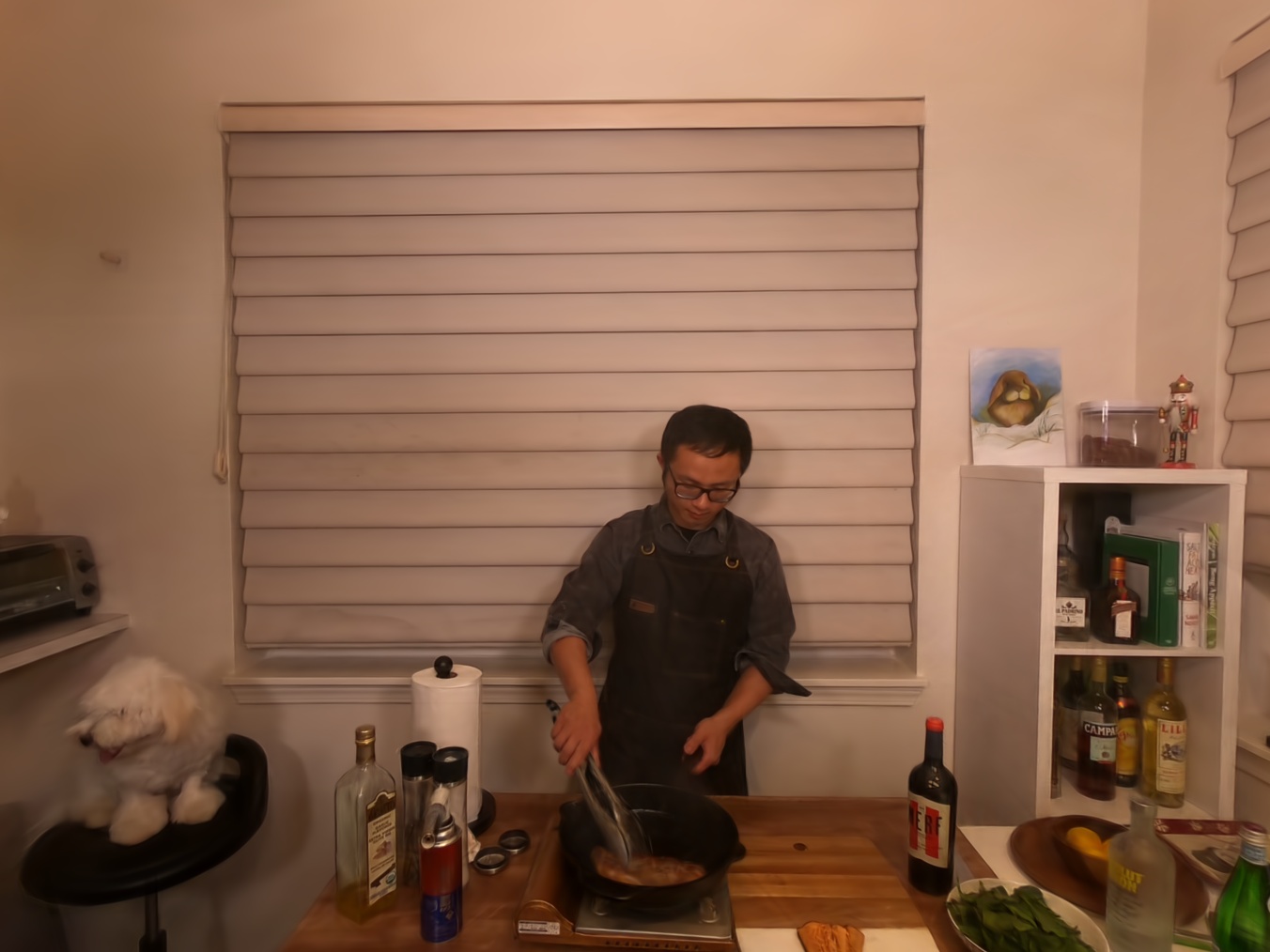}
   }
   \subfloat[GroudTruth]{
      \includegraphics[width=0.33\textwidth]{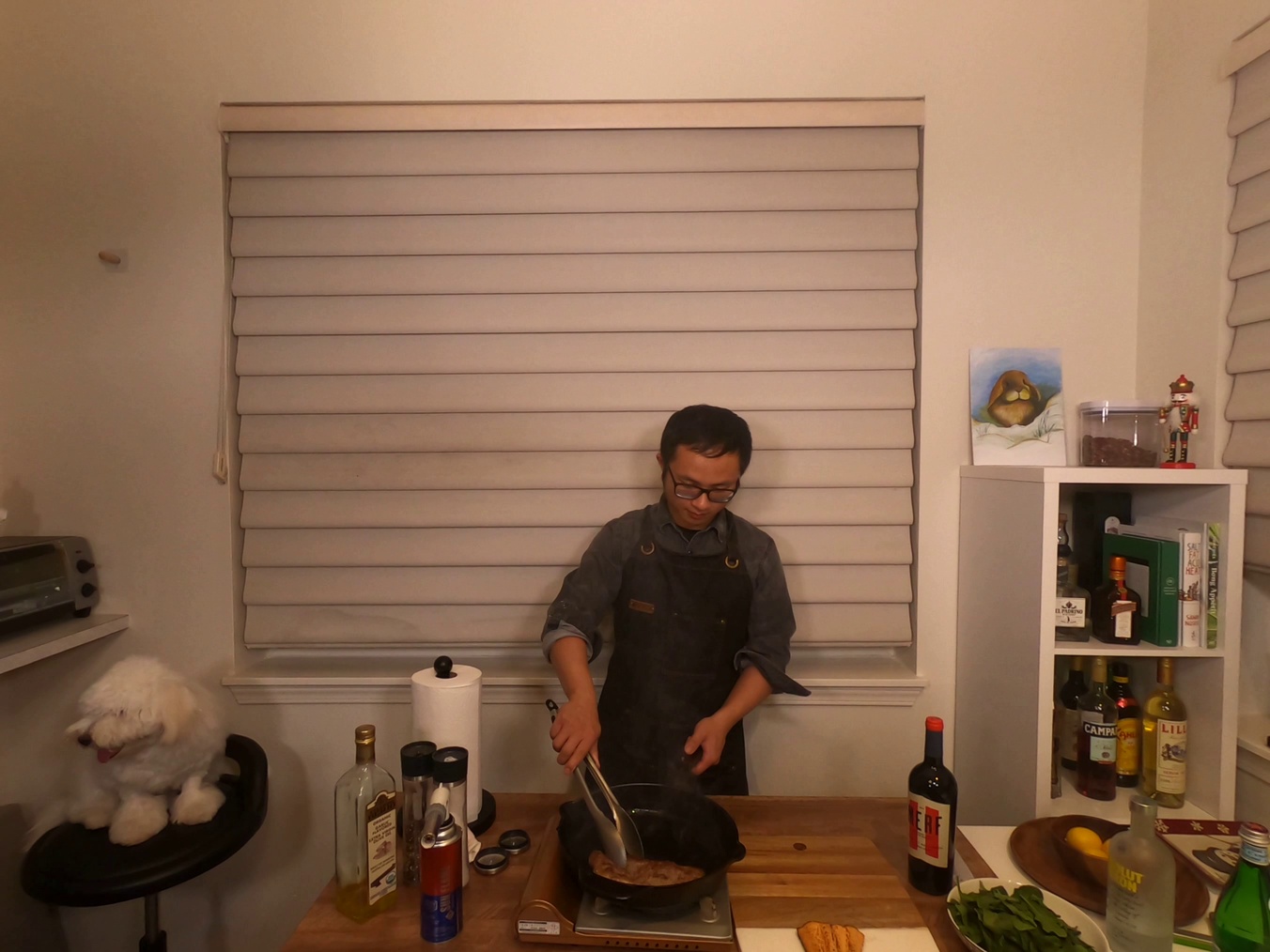}
   }
   \caption{Qualitative comparison under different compression qp on \textit{Sear Steak} scene of N3DV dataset.}
   \label{fig:qp-sear}
   \vspace{-0.2cm}
\end{figure*}

To supplement the quantitative analysis presented in this paper, we provide a comprehensive visual assessment of all scenes under varying compression parameters, as illustrated in Fig.~\ref{fig:qp-coffee} through Fig.~\ref{fig:qp-sear}. When $QP=32$, the N3DV dataset achieves an average PSNR of 30.76, while maintaining relatively high rendering quality. Static regions exhibit satisfactory quality within a QP range of 16 to 32, with only minor artifacts appearing in dynamic regions.

\typeout{get arXiv to do 4 passes: Label(s) may have changed. Rerun}
\end{document}